\documentclass[a4paper,11,twoside]{report}
\usepackage[left=2.5cm,right=2cm,top=2cm,bottom=2cm]{geometry}
\usepackage{graphicx}
\usepackage{subfigure}
\usepackage[section]{placeins} 
\usepackage{algorithm}
\usepackage{algorithmic}
\usepackage{listings}
\usepackage{comment}
\usepackage{amsmath}
\graphicspath{{/},{results/}}
\begin{document}

\begin{titlepage}

  \begin{center}

    {\Large Imperial College London}\\[0.5cm]
    {\Large Department of Computing}\vfill

    {\huge Reinforcement Learning in a Neurally Controlled Robot Using Dopamine Modulated STDP}\\[0.4cm]
    {\large by}\\[0.4cm]
    {\Large Richard Evans}\\[0.4cm]
    \vfill

    {\Large Submitted in partial fulfilment of the requirements for the MSc Degree in Advanced Computing of Imperial College London }\\[2cm]

    {\Large September 2012}
  \end{center}

\end{titlepage}

\pagenumbering{roman}
\tableofcontents

\begin{abstract}
Recent work has shown that dopamine-modulated STDP can solve many of the issues associated with reinforcement learning, such as the distal reward problem.  Spiking neural networks provide a useful technique in implementing reinforcement learning in an embodied context as they can deal with continuous parameter spaces and as such are better at generalizing the correct behaviour to perform in a given context.

In this project we implement a version of DA-modulated STDP in an embodied robot on a food foraging task.  Through simulated dopaminergic neurons we show how the robot is able to learn a sequence of behaviours in order to achieve a food reward.  In tests the robot was able to learn food-attraction behaviour, and subsequently unlearn this behaviour when the environment changed, in all 50 trials.  Moreover we show that the robot is able to operate in an environment whereby the optimal behaviour changes rapidly and so the agent must constantly relearn.  In a more complex environment, consisting of food-containers, the robot was able to learn food-container attraction in 95\% of trials, despite the large temporal distance between the correct behaviour and the reward.  This is achieved by shifting the dopamine response from the primary stimulus (food) to the secondary stimulus (food-container).

Our work provides insights into the reasons behind some observed biological phenomena, such as the bursting behaviour observed in dopaminergic neurons.  As well as demonstrating how spiking neural network controlled robots are able to solve a range of reinforcement learning tasks.

\end{abstract}
\chapter*{Acknowledgements}
I would like to thank my supervisor Prof. Murray Shanahan.  Whose support and guidance helped greatly in allowing me to conduct research into neural robotics.

\pagenumbering{arabic}

\chapter{Introduction}
\label{ch:intro}
\section{Motivation}
The ability of a robot to operate autonomously is a highly desirable characteristic and has applications in a wide range of areas, such as space exploration, self-driving cars, cleaning robot and assistive robotics.  Looking at the animal kingdom, a key property that a large variety of animals possess is that they are able to learn which behaviours to perform in order to receive a reward or avoid unpleasant situations, for example in foraging for food or avoiding prey.  This is also a useful property for robots to have, where we want the robots to achieve some goal but the exact sequence of behaviours needed to achieve this goal may be highly complex or change over time.  This learning paradigm is known as reinforcement learning and has been researched in the context of machine learning for many years.  Despite this, even simple animals are able to outperform robots in the vast majority of real world reinforcement learning problems.  

In recent times some of the underlying neural processes by which the brain can solve these reinforcement learning tasks has started to be revealed.  Changes in connection strength between neurons has long been thought to be the key process by which animals are able to learn.  A key process that is believed to control how the connection weights are modified is that of dopamine-modulated spike timing dependent plasticity (DA-modulated STDP)~\cite{Izhikevich_2007}.  Using spiking neural networks, that implement these models of learning and plasticity, to control robots has two main advantages.  Firstly, it is hoped that these neurally inspired control algorithms will, at least in some domains, be able to outperform their classical machine learning counterparts.  This is due to the vast arrays of problems that real brains can deal with and their implicit ability to deal with continuous environmental domains.  Secondly, by implementing the current models of neurological learning in an embodied context we can gain greater insight into the range of behaviour that can be explained by these models, and also where these models fail.

\section{Project Aims}

In this project we will attempt to solve a reinforcement learning task of food foraging and poison avoidance using a robot controlled with a spiking neural network.  By incorporating several aspects of the brain that have been observed we aim to show how a robot can learn attraction and avoidance behaviours in a dynamic environment. By designing an environment that requires a sequence of behaviours before food is reached we aim to show that the robot can learn by propagating the reward signal (dopamine) to reward-predicting stimulus.  In this way we aim to show how a robot is able to learn sequences of behaviour over larger time-scales than have previously been demonstrated.

This work builds on the models of Izhikevich~\cite{Izhikevich_2007} and others in providing the neurological models that will be used to control the robot.  The use of these models to control a robot in a foraging/avoidance task is based on the work of Chorley \& Seth~\cite{Chorley_2008}.  We aim to show that through several key extensions, such as incorporating the dopamine signal into the network directly, as well as using a more generic network architecture, our robot is able to deal with a much wider variety of problems, learn over longer time-scales and learn sequences of behaviour that previous implementations wouldn't have been able to learn.  These are vital properties for an agent interacting in the real world and a key step towards a fully autonomous agent that can deal with the wide variety of problems scenarios that the real world provides.

The mechanisms of plasticity and stability play a key role in many properties of the brain, such as learning, memory and attention.  Through implementing a neurally inspired system that incorporates many of the current neural models we aim to show how well these models are able to explain some of the learning behaviours that are seen in the animal world.

\section{Thesis Outline}
The remainder of this report is structured into five chapters.  In chapter 2 we review current research and relevant literature in the context of reinforcement learning in neurally controlled robots.  Chapter 3 presents the techniques, models and methodology that was used in this project.  The architecture of the robot, set-up of the environment and the spiking neural network control architecture is discussed. The results of evaluating our robot in a wide range of scenarios is discussed in chapter 4, along with a discussion of which properties of the robot allow it to achieve these results.  In chapter 5 we relate the results to the wider context of reinforcement learning and neurological models and discuss their implications.  Finally in chapter 6 we present possible future work, based on the results of this paper.

\chapter{Background}
\label{ch:background}

\section{Neurons}
\label{section:neurons}
The underlying biological structure of neurons, and models that have been used to simulate them, are discussed in this section.
\subsection{Biological Neurons}
\begin{figure}[htbp]
	\begin{center}
	  	\includegraphics[width=0.6\textwidth]{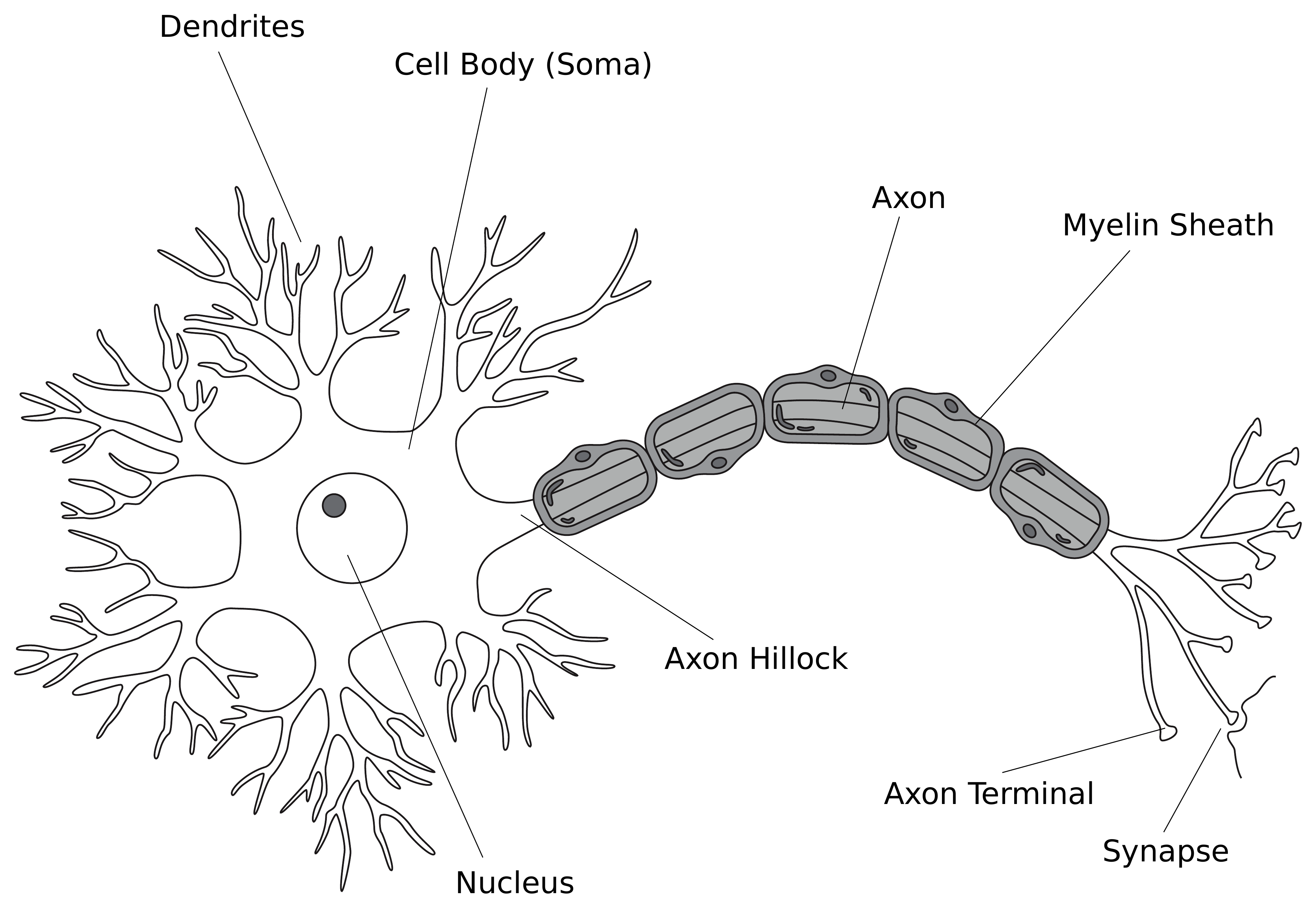}	
	\end{center}
	\caption{The structure of a neuron.}
	\label{fig:neuron}
\end{figure}

\begin{figure}[htbp]
	\begin{center}
	  	\includegraphics[width=0.6\textwidth]{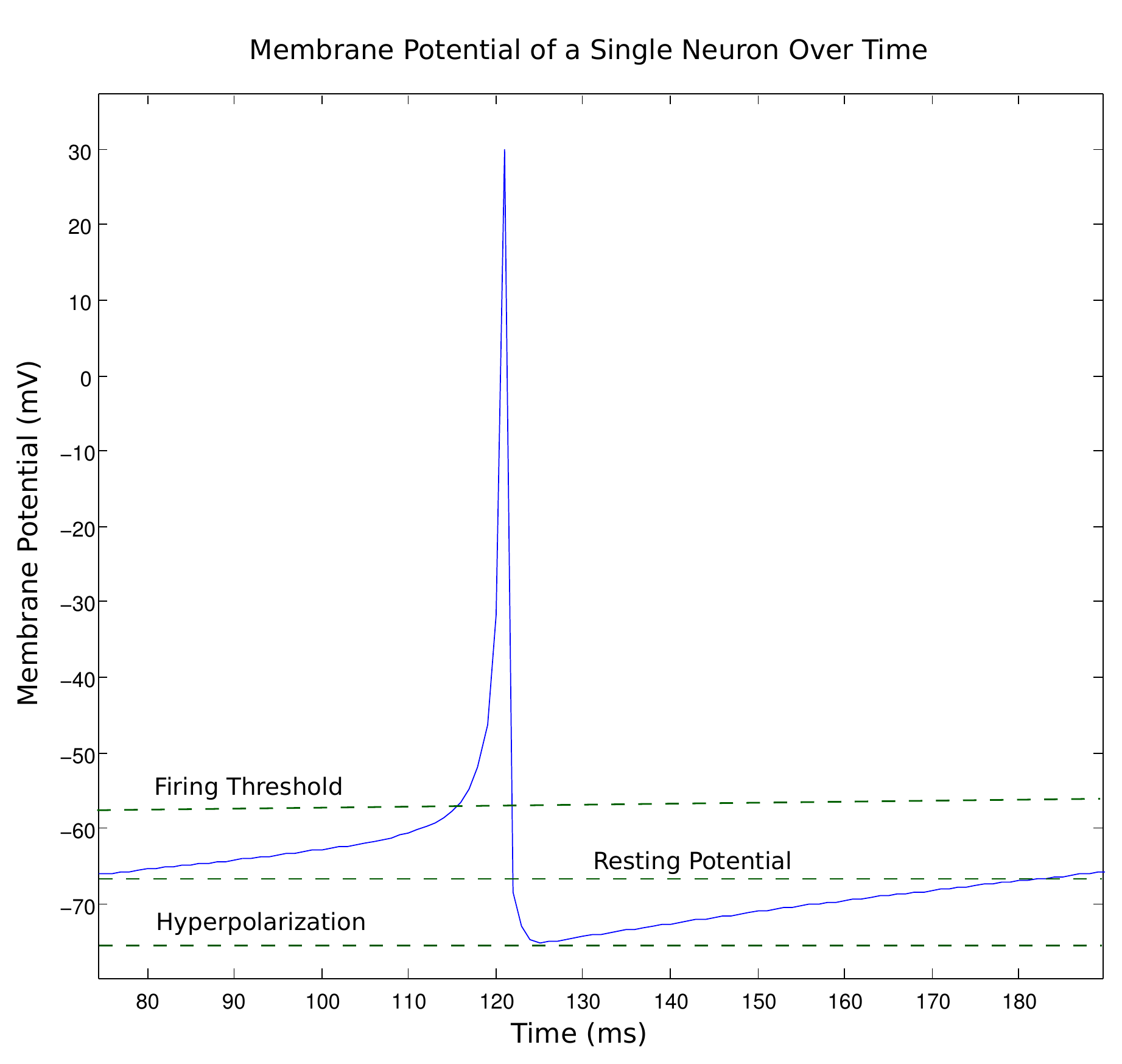}	
	\end{center}
	\caption{The membrane potential of a neuron receiving constant input current plotted over time.}
	\label{fig:neuron_spike}
\end{figure}

The brain consists of a vast connected network of neurons.  These neurons can be further subdivided into different types, however the general architecture is the same in almost all neurons.  Figure \ref{fig:neuron} shows the basic structure of a typical neuron.  
Within the neuron communication takes place via the electrical properties of the cell, figure \ref{fig:neuron_spike} shows the membrane potential of a neuron plotted over time whilst receiving a constant input current.  In its resting state the membrane potential of the interior of a neuron sits at approximately -70mV.  Each neuron receives inputs from several other neurons via its dendrites.  When one of these pre-synaptic neurons fires it has the effect of raising or lowering the membrane potential of the post-synaptic neuron.  Once the membrane potential at the axon hillock (see figure \ref{fig:neuron}) reaches a critical threshold, voltage gated ion channels open along the axon which causes the neuron to rapidly depolarize.  This is referred to as the neuron spiking or firing.  After this wave of depolarization reaches the axon terminal it is converted to a chemical signal that is passed across the synapse to the next neuron.  Once this depolarization reaches a critical threshold a second voltage gated ion channel is opened which causes the neuron to rapidly repolarize before returning to its resting potential.  This repolarization overshoots the resting potential causing hyperpolarization.  After firing the neuron enters a refractory period during which it cannot fire.

The two main types of neurons that will be used in this paper are excitatory and inhibitory neurons.  Excitatory neurons have the effect of raising the membrane potential of the post-synaptic neuron, as such increasing its likelihood of firing.  Whereas inhibitory neurons have the effect of decreasing the membrane potential of the post-synaptic neuron, decreasing its likelihood of firing.

The synapse between neurons has an associated weight or strength.  This is the degree to which the post-synaptic membrane potential is raised or lowered following a pre-synaptic spike. The mechanisms by which neurons communicate have several useful properties.  Firstly, communication within the cell is fast, via the use of voltage gated ion channels.  Secondly, by communicating in an all or nothing fashion to the post-synaptic cell, the network of neurons is able to represent highly complex non-linear functions.  Research has shown that spiking neural networks can represent any real-valued function to within a specified degree of accuracy~\cite{Iannella_2001}.


\subsection{Hodgkin-Huxley Model}
In 1952 Hodgkin and Huxley developed an accurate model of the membrane potential of a neuron~\cite{Hodgkin_1952} based on modelling the ion channels within the neuron.  The equation is defined as:
\begin{equation}
C\frac{dv}{dt}=-\sum_k I_k+I \\
\end{equation}
\begin{align*}
  \text{where} \\
  C &= \text{The capacitance of the neuron,} \\
  v &= \text{The membrane potential of the neuron,} \\
  I_k &= \text{The various ionic currents that pass through the cell,} \\
  I &= \text{The external current coming from pre-synaptic neurons,} \\
  t &= \text{Time.}
\end{align*}
The ionic currents are modelled using further differential equations, the details of which are not included here but a full definition can be found in Hodgkin \& Huxley~\cite{Hodgkin_1952}.  The Hodgkin-Huxley equation is very biologically accurate, however it incurs a heavy computational cost and as such is infeasible in most practical situations.

\subsection{Izhikevich Model}
\label{section:izhikevich}
Several attempts~\cite{Lapicque_1907,FitzHugh_1961,Hindmarsh_1984} have been made to formulate a model that provides a good compromise between biological accuracy and computational feasibility.  One of the best models in terms of computational efficiency and biological accuracy was formulated by Izhikevich~\cite{Izhikevich_2003}.  In the Izhikevich model the membrane potential is modelled as:
\begin{equation}
\frac{dv}{dt}=0.04v^2+5v+140-u+I
\end{equation}
Where $v$ is the membrane potential and $u$ is the recovery variable that determines the refractory period of the neuron and is modelled as:
\begin{equation}
\frac{du}{dt}=a(bv-u)
\end{equation}
Where $a$ and $b$ are parameters of the model.  When a spike occurs the membrane potential is reset and the recovery variable is incremented, formally:
\begin{equation}
  \text{if $v \geq 30$ then} \left\{
  \begin{array}{l l}
  	v\leftarrow c & \quad \\
  	u \leftarrow u+d & \quad \\
  \end{array} \right.
\end{equation}

\begin{figure}[htbp]
	\begin{center}
	  	\includegraphics[width=0.9\textwidth]{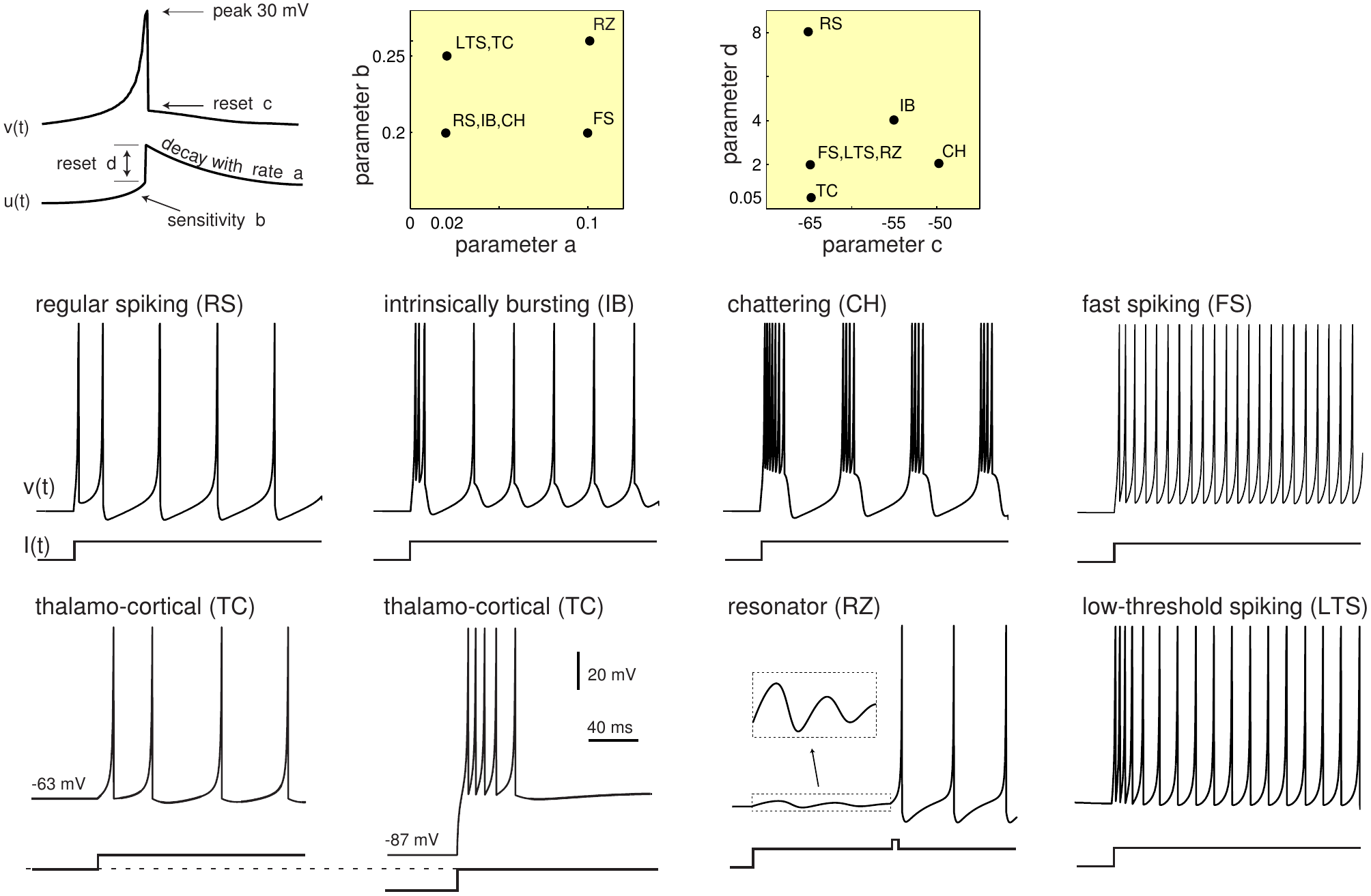}	
	\end{center}
	\caption[An overview of some types of neurons that can be modelled with the Izhikevich model.]{An overview of some types of neurons that can be modelled with the Izhikevich model\protect\footnotemark.}
	\label{fig:izhik_model}
\end{figure}

The four variables $a$,$b$,$c$ and $d$ have been shown to be able to create neurons that have a wide variety of behaviours.  Figure \ref{fig:izhik_model} summarizes some of the neuronal types that can be modelled by varying these parameters.
\footnotetext{Electronic version of the figure and reproduction permissions are freely available at www.izhikevich.com}

\section{Models of Neural Networks}
In this section the different approaches that have been used to model networks of connected neurons are discussed.
\subsection{Artificial Neural Networks}
\begin{figure}[htbp]
	\begin{center}
	  	\includegraphics[width=0.5\textwidth]{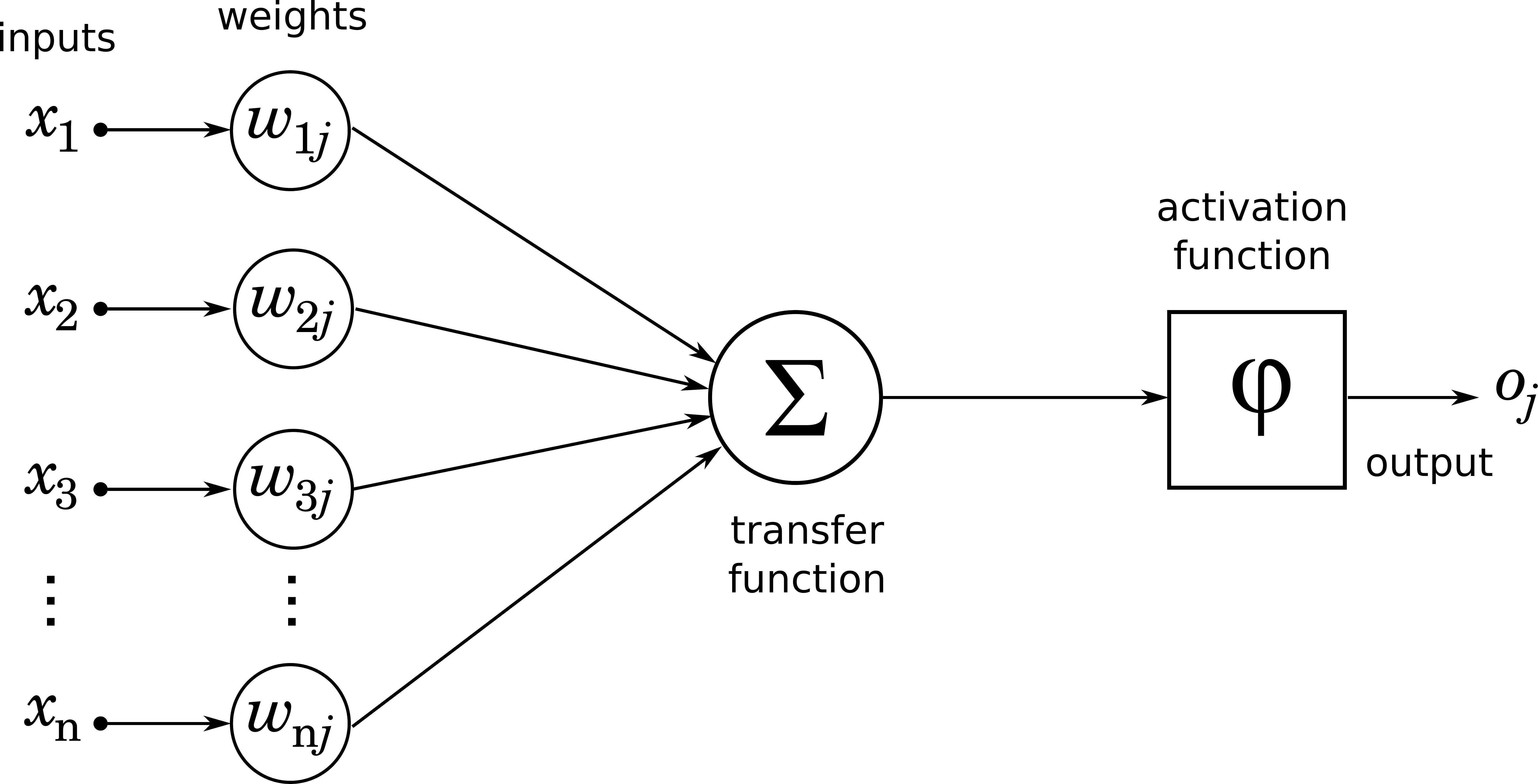}	
	\end{center}
	\caption{A perceptron that computes the weighted sum over the inputs $x$ with weights $w$, then passes this value through the activation function to produce a binary output.}
	\label{fig:perceptron}
\end{figure}

The first computational model of a neural network that was developed was the artificial neural network (ANN).  The ANN consists of a set of processing units, or neurons.  Each neuron computes the weighted sum of its inbound connections, which may be other neurons or external inputs.  This value is then passed through an activation function, the result of which is then passed to the next set of neurons or as an output of the system.  The most common activation function that is used is the sigmoid function as this provides a differentiable approximation to the all or nothing processing that is used in real neurons.  Figure \ref{fig:perceptron} shows the model of a single neuron in an ANN.

Initially these ANN's consisted of a single layer of neurons~\cite{Rosenblatt_1962,Widrow_1986}.  These ANN's, known as perceptrons, are limited in that they can only represent linear functions of their inputs.  For example in a classification task, the data must be linearly separable for a single layer network to be able to classify all data correctly.  To combat these problems a multilayer perceptron was introduced~\cite{Werbos_1974}, where outputs from previous layers serve as inputs to the next layer.  

The development of an efficient training algorithm for the multilayer perceptron, known as back-propagation, which updates the weights of the network to minimize the squared error on a set of training examples has allowed ANN's to be used in a wide variety of fields.

The biggest limiting factor in artificial neural networks, especially as a model of the brain, is that they can't process information encoded in the time domain.

\FloatBarrier
\subsection{Spiking Neural Networks}
\label{section:spiking_neural_networks}
A more realistic model of biological neural networks was developed with spiking neural networks (SNN).  These have the advantage of being able to process temporally distributed data, as well as having a large memory capacity. Unlike ANN's, which are commonly directed acyclic graphs, SNN's are more amenable to networks with cycles in them.  These cycles allow spiking neural networks to have a form of working memory.

The individual neurons in a spiking neural network can be modelled in a variety of ways, several of these models were outlined in section \ref{section:neurons}.  Some methods by which SNN's can be trained are outlined in section \ref{section:reinforcement_learning_in_the_brain}.

\section{Reinforcement Learning in Machine Learning}
\label{section:reinforcement_learning_in_machine_learning}
Reinforcement learning is the process by which an agent can automatically learn the correct behaviour in order to maximize its performance.  Reinforcement learning is an important learning paradigm for a robot that interacts with the real world, in this scenario the robot is often not explicitly told what the correct action is in a given environment state, it must be inferred from rewards or punishments that are received after an action is performed.  In the standard reinforcement learning set-up the agent is given a reward when it performs an action or is in a specific state, and it seeks to maximize this reward over time.  Sutton \& Barto give a good history of reinforcement learning algorithms~\cite{Sutton_1998}, the key approaches relevant to this paper are outlined below.

\subsection{Markov Algorithm's}

\begin{figure}[htbp]
	\begin{center}
	  	\includegraphics[width=0.5\textwidth]{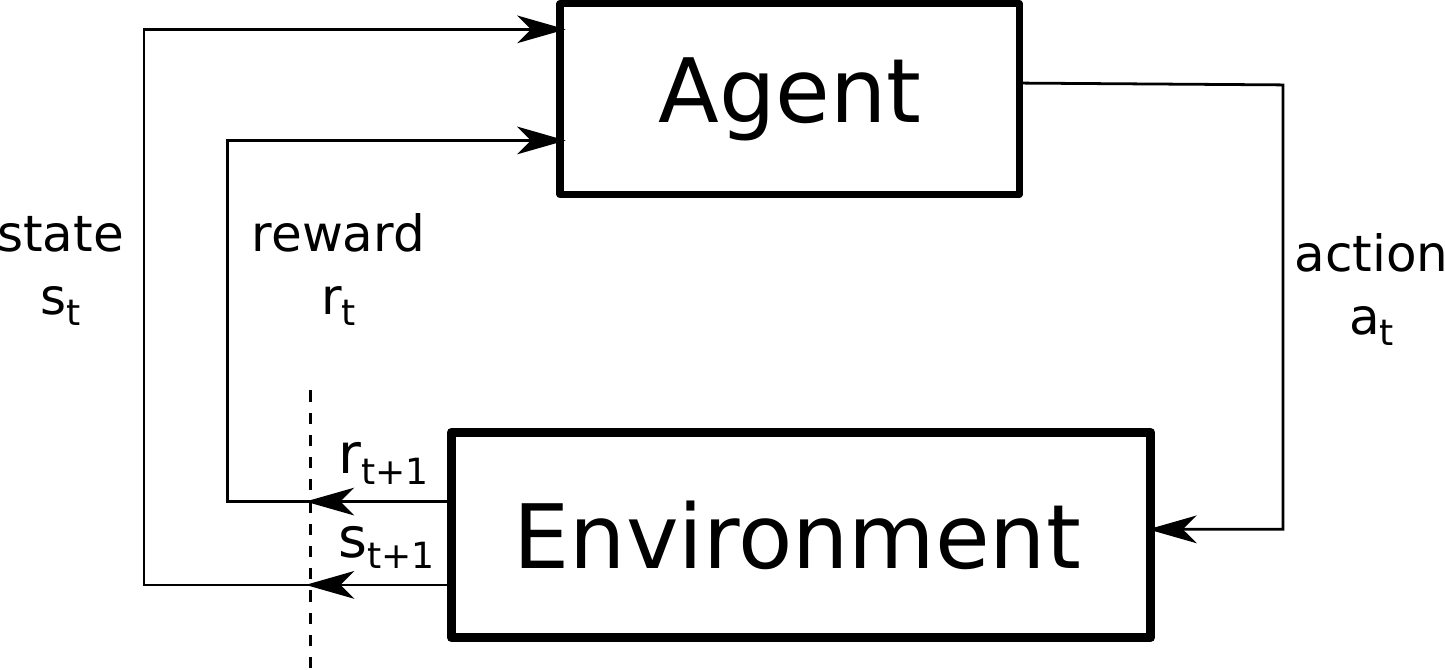}	
	\end{center}
	\caption{The agent-environment interaction.  Given a state $s_t$ and reward $r_t$ at time $t$ the agent must decide which action $a_t$ to perform.  This action will update the state $s_{t+1}$ and result in a further reward $r_{t+1}$. Figure from Sutton \& Barto~\cite{Sutton_1998}.}
	\label{fig:agent_environment}
\end{figure}

The majority of work on reinforcement learning have modelled the problem as a Markov decision process.  The agent can visit a finite number of states, in each state the agent can perform a finite number of actions that transform the environment into a new state.  The state that is reached after performing an action only depends on the previous state, and the action.  This agent-environment interaction is shown in figure \ref{fig:agent_environment}.

An agent's policy $\pi$ is defined as the probability of choosing an action in a given state: $\pi(s,a)=p(a|s,\pi)$.  An important function in reinforcement learning is the Q-function, this gives the expected future reward given a state and action, defined as:
\begin{equation}
Q^\pi(s,a)=E[\sum_{k=0}^\infty \gamma^kr_{t+k+1}|\pi,s_t=s,a_t=a]
\end{equation}
Where $\gamma$ determines the importance we place on rewards that are closer in time than ones that are more distant.  If we know the Q-function then the optimal policy becomes trivial, we pick the action with highest expected reward.  Most RL algorithms are concerned with finding the value of the Q-function.

An underlying feature of most RL algorithms is the concept of temporal difference (TD) learning.  Where the difference between the predicted reward and received reward is used to update the agents policy.  RL algorithms are divided into off-line and on-line learning.  Off-line algorithms assume that a series of states and actions generated as in figure \ref{fig:agent_environment} will eventually reach a terminating state.  Learning is only performed once this terminating state is reached.  On-line algorithms, however, aim to continuously update the agents estimate of the Q-function as the agent explores the environment.  On-line algorithms provide the advantage of not having to wait until the end of an episode to incorporate new information and, as is the case in this paper, an agent interacting in the real world will not always have a clearly defined termination condition that causes an episode to end.

Within the set of on-line RL algorithms there is a further distinction between on-policy and off-policy algorithms.  On-policy algorithms attempt to learn the optimal Q-function and policy, whilst simultaneously following their own estimate of the optimal policy.  Note that the agents estimate of the optimal policy is likely to be different to the actual optimal policy, as such this method can get stuck in local minima and not converge to the optimum policy.  Off-policy algorithms on the other hand attempt to learn the optimal policy whilst not necessarily following it, choosing actions with a probability related to their expected reward.  This has the advantage that the agent can further explore the state space, however this comes at the cost of a possible reduced reward compared to the on-policy methods, due to the exploration of sub-optimal states.  In this paper we will demonstrate a mechanism by which off-policy learning can be achieved using spiking neural networks.

The best known on-line algorithm is Sarsa, the outline of which is given in Algorithm \ref{alg:sarsa}.  The key step is that the observed reward $r$ and the predicted Q-value at the next time step are used to estimate the correct Q-value: $r + \gamma Q(s',a')$.  This is compared against the predicted Q-value, $Q(s,a)$, to give a prediction error which can be used to update the estimate of $Q(s,a)$.

\begin{algorithm}
\caption{Sarsa}
\label{alg:sarsa}
\newcommand{\INDSTATE}[1][1]{\STATE\hspace{#1\algorithmicindent}}
\begin{algorithmic}[frame]
\STATE Initialize $Q(s,a)$ arbitrarily
\STATE Repeat (for each episode)
\INDSTATE	Initialize $s$
\INDSTATE   Choose $a$ from $s$ using policy derived from $Q$
\INDSTATE	Repeat (for each step of episode):
\INDSTATE[2]	Take action $a$, observe $r$, $s'$
\INDSTATE[2]	Choose $a'$ from $s'$ using policy derived from $Q$
\INDSTATE[2]	$Q(s,a) \leftarrow Q(s,a) + \alpha [r + \gamma Q(s',a') - Q(s,a)]$
\INDSTATE[2]	$s \leftarrow s';a \leftarrow a';$
\INDSTATE   until $s$ is terminal
\end{algorithmic}
\end{algorithm}

The best known algorithm for off-line learning is Q-learning, the pseudocode for which is given in Algorithm \ref{alg:q_learning}.  The key difference is that the algorithm takes the maximum Q-value in the next state, over all actions.

\begin{algorithm}
\caption{Q-Learning}
\label{alg:q_learning}
\newcommand{\INDSTATE}[1][1]{\STATE\hspace{#1\algorithmicindent}}
\begin{algorithmic}[frame]
\STATE Initialize $Q(s,a)$ arbitrarily
\STATE Repeat (for each episode)
\INDSTATE	Initialize $s$
\INDSTATE   Choose $a$ from $s$ using policy derived from $Q$ with exploration
\INDSTATE	Repeat (for each step of episode):
\INDSTATE[2]	Take action $a$, observe $r$, $s'$
\INDSTATE[2]	$Q(s,a) \leftarrow Q(s,a) + \alpha [r + \gamma max_{a'}Q(s',a') - Q(s,a)]$
\INDSTATE[2]	$s \leftarrow s'$
\INDSTATE   until $s$ is terminal
\end{algorithmic}
\end{algorithm}

\subsection{Eligibility Traces}
\begin{figure}[htbp]
	\begin{center}
	  	\includegraphics[width=0.65\textwidth]{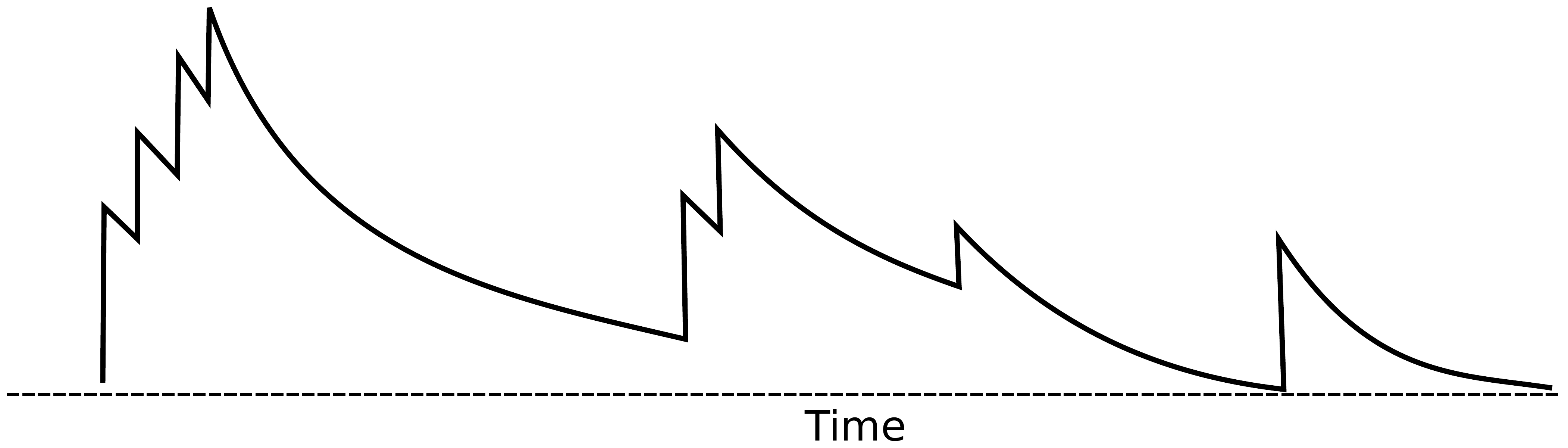}	
	\end{center}
	\caption{The eligibility trace for a state over time, as it is repeatedly visited.}
	\label{fig:eligibility_trace}
\end{figure}

Both the Q-Learning and Sarsa algorithm as described only have one state of lookahead.  They only look at the reward at the next time step when updating the Q-function.  Due to this we often have to revisit a state many times before the algorithm will converge.  We encounter further problems if we relax the Markov property, for example if the reward received for an action is not received until several states later.  In this case it can be shown that the algorithms above may fail to converge to the correct Q-function.  What we would like to be able to do is incorporate the reward from all future states in our update function.  This is accomplished by introducing the concept of an eligibility trace.  The eligibility trace for a state keeps a weighted track of visits to that state, in this way we can update previously visited states when we receive a new reward, the concept of the eligibility trace becomes important when discussing the dynamics of the reinforcement learning in the brain, and in our implementation of DA-modulated STDP.  Figure \ref{fig:eligibility_trace} demonstrates how the eligibility trace changes over time, as the state is repeatedly visited.

We can incorporate the eligibility trace into the Sarsa algorithm (referred to as Sarsa($\lambda$)), if we define $e(s,a)$ as the eligibility trace for the state $s$ and action $a$ then the Q-function update becomes:
\begin{equation}
Q(s,a) \leftarrow Q(s,a) + \alpha e(s,a) [r + \gamma Q(s',a') - Q(s,a)]
\end{equation}
Where $\alpha$ is the step size; the amount that we update the Q values by, at each time step.  $\gamma$ determines the importance we place on more recent rewards over future rewards.  $Q(s',a')$ is the Q value for the next state and action.  Note that this update is applied to all states and actions, unlike in the previous algorithm where it was restricted only to the current state-action pair.  The full Sarsa($\lambda$) procedure is outlined in Algorithm \ref{alg:sarsa_lambda}.

\begin{algorithm}
\caption{Sarsa($\lambda$)}
\label{alg:sarsa_lambda}
\newcommand{\INDSTATE}[1][1]{\STATE\hspace{#1\algorithmicindent}}
\begin{algorithmic}
\STATE Initialize $Q(s,a)$ arbitrarily and $e(s,a)=0$, for all $s$,$a$
\STATE Repeat (for each episode)
\INDSTATE	Initialize $s$, $a$
\INDSTATE	Repeat (for each step of episode):
\INDSTATE[2]	Take action $a$, observe $r$, $s'$
\INDSTATE[2]	Choose $a'$ from $s'$ using policy derived from Q
\INDSTATE[2]	$\delta \leftarrow r + \gamma Q(s',a') - Q(s,a)$
\INDSTATE[2]	$e(s,a) \leftarrow e(s,a)+1$
\INDSTATE[2]	For all $s$,$a$:
\INDSTATE[3]		$Q(s,a) \leftarrow Q(s,a) + \alpha \delta e(s,a)$
\INDSTATE[3]		$e(s,a) \leftarrow \gamma \lambda e(s,a)$
\INDSTATE[2]	$s \leftarrow s';a \leftarrow a'$
\INDSTATE   until $s$ is terminal
\end{algorithmic}
\end{algorithm}

\subsection{Continuous Parameter Spaces}
So far our discussion has been limited to reinforcement learning in discrete state and action spaces.  The simplest way of applying RL to a continuous state space is simply to divide the continuous space up such that it becomes a discrete one.  However there are two main issues with this, firstly we may end up with a very large parameter space, especially if we wish to have a fine grained representation of the continuous parameter space.  This will mean that the RL algorithm will take a long time to converge on the optimal solution as the probability of visiting each state will be low.  The second issue is that we lose the ability to generalize, it often the case that knowing the expected future reward for a particular state will tell us something about the expected reward for nearby states.  

In the continuous case the Q-function becomes a continuous valued function over $s$ and $a$.  If we treat each update from the discrete RL algorithm as an example input-output of this function. Then our problem becomes one of generalizing from specific inputs.  This is a standard machine learning problem, which can be solved through, among other things, neural networks, decision trees, genetic algorithms and gradient descent.

Another method of performing reinforcement learning in a continuous parameter space is to use spiking neural networks, a full description of spiking neural networks was given in section \ref{section:spiking_neural_networks}.  Spiking neural networks offer the advantage that, if constructed correctly, they can generalize to learn the correct action in a previously unseen state.  Reinforcement learning was first investigated, and algorithms developed, before it was fully understood how the brain performed similar tasks.  As will be shown in the following section the underlying processes used by the brain has many correlates with these reinforcement learning algorithms.  In this paper we implement a spiking neural network version of reinforcement learning, how the underlying processes in our implementation compare to standard machine learning algorithms for reinforcement learning are discussed in section \ref{section:conclusion_reinforcement_learning}.

\section{Reinforcement Learning in the Brain}
\label{section:reinforcement_learning_in_the_brain}
In recent times it has been shown that the brain implements a reinforcement learning system similar to the TD-Learning proposed by Sutton \& Barto~\cite{Sutton_1998}.  The mechanisms by which this happens are outlined below.

\subsection{BCM Theory}
\label{section:bcm_theory}
In 1981 Bienenstock, Cooper and Munro developed a model by which synapses can increase and decrease over time~\cite{Bienenstock_1982}.  This slow increasing and decreasing of synaptic weights is known as long term potentiation (LTP) and long term depression (LTD) respectively.  In BCM theory the amount that a synaptic weight changes is dependent on the product of the presynaptic firing rate and a non-linear function $\phi$ of the postsynaptic activity.  For low post-synaptic firing rates $\phi$ is negative and for high post-synaptic firing rates $\phi$ is positive.  The effect of this is that if the pre-synaptic neuron is causing a lot of firing in the post-synaptic neuron then the synapse will be potentiated.

Some evidence for the BCM model has been seen in the brain, however it cannot account for synaptic potentiation based on spike timings, which has also been observed in the brain.

\subsection{Spike-timing Dependent Plasticity}
\label{section:stdp}
The main mechanism through which the connections in the brain are modified over time is believed to be spike-timing dependent plasticity~\cite{Markram_1997,Song_2001}.  The core concept is that neuron connection strength is increased when pre-synaptic activity is followed by post-synaptic activity, and weakened when post-synaptic activity is followed by pre-synaptic activity. The synaptic weight update rule under STDP is defined as:

\begin{equation}
  \Delta w = \left\{
  \begin{array}{l l}
  	A^+e^{-\Delta t/\tau^+} & \quad \textrm{if $\Delta t \geq 0$}\\
  	-A^-e^{\Delta t/\tau^-} & \quad \textrm{if $\Delta t < 0$}\\
  \end{array} \right.
\end{equation}

\begin{figure}[htbp]
	\begin{center}
	  	\includegraphics[width=0.3\textwidth]{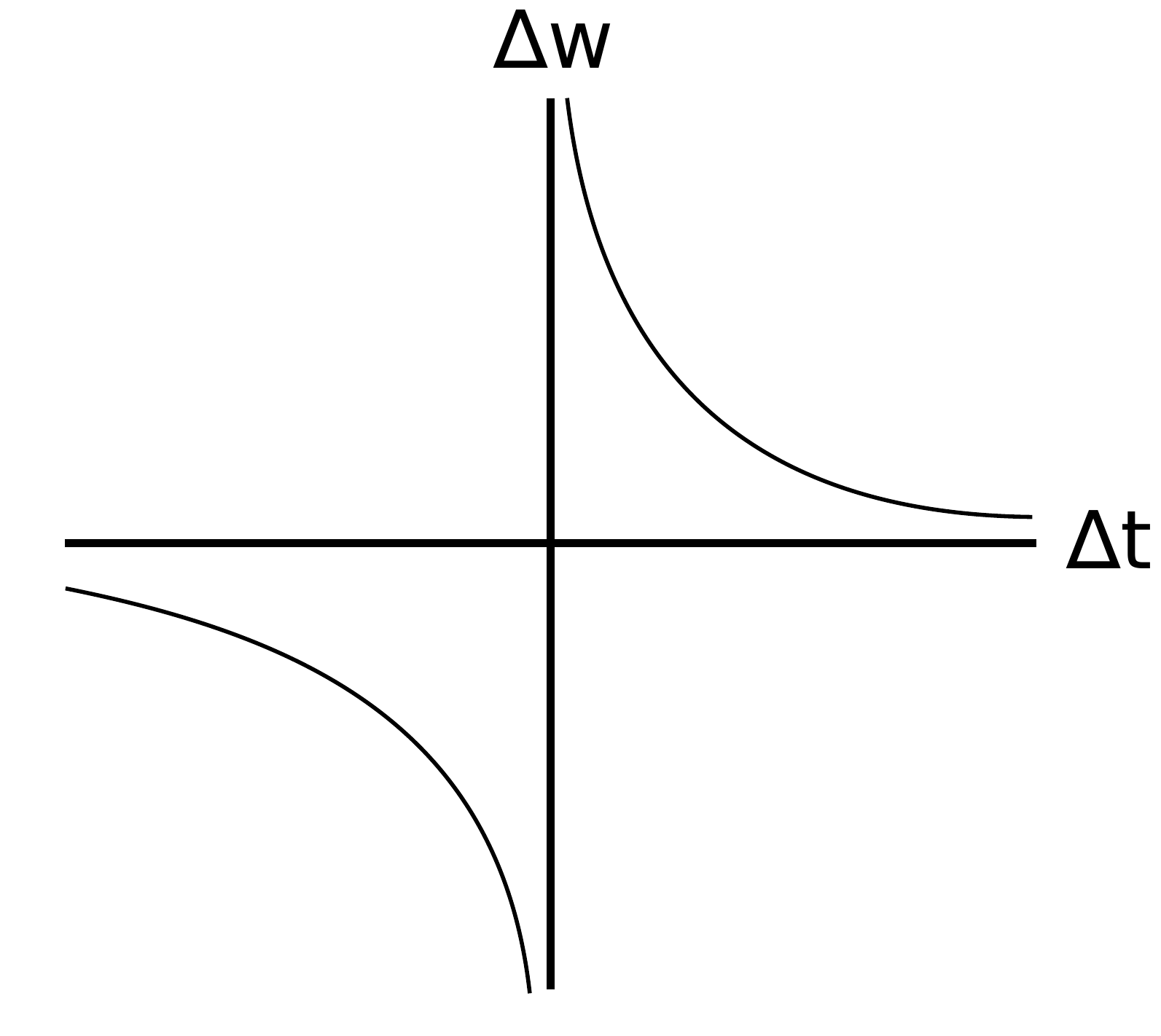}	
	\end{center}
	\caption{Graph showing how the weight update $\Delta w$ relates to the $\Delta t = t_{\mathit{post}} - t_{\mathit{pre}}$ parameter.}
	\label{fig:stdp_graph}
\end{figure}

Where $\Delta w$ is the weight update, $\Delta t$ is the time difference between the pre and post synaptic neurons firing and $A^+$,$A^-$,$\tau^+$ and $\tau^-$ are constants that define how STDP is applied over time.  These constants are normally chosen so that long term depression is favoured over long term potentiation to prevent uncontrollable growth of $w$.  Figure \ref{fig:stdp_graph} shows how $\Delta w$ changes with respect to $\Delta t$, this shows the window over which STDP can act.  Large values of $\Delta t$ will have negligible effect on $\Delta w$.

There are two main update schemes that are followed when calculating the change in weight $\Delta w$.  In the all-to-all scheme all pre-synaptic and post-synaptic spikes within the STDP window are considered when updating the weight.  In the nearest-neighbour scheme only the most recent pre or post synaptic spike is considered.  In practice these methods are often equivalent, as the probability of multiple firings for a specific neuron within the STDP window is low.  STDP can be thought of as an associative learning mechanism.  If two inputs to the neural network are often presented together, and in sequence, then the connections from the first input to the second input will become strengthened.

Experiments have provided evidence for both STDP and the BCM model in the brain.  Therefore any model of synaptic weight modification will need to account for both these processes.  Izhikevich~\cite{Izhikevich_2003b} has shown that under the nearest-neighbour mechanism it is possible for the observed properties of the BCM model to be generated through STDP.  As such it may be that the underlying mechanism for synaptic potentiation/depression in the brain is indeed STDP.
\FloatBarrier
\subsection{Dopamine Modulated STDP}
\label{section:dop_stdp}

\begin{figure}[htbp]
	\begin{center}
	  	\includegraphics[width=0.55\textwidth]{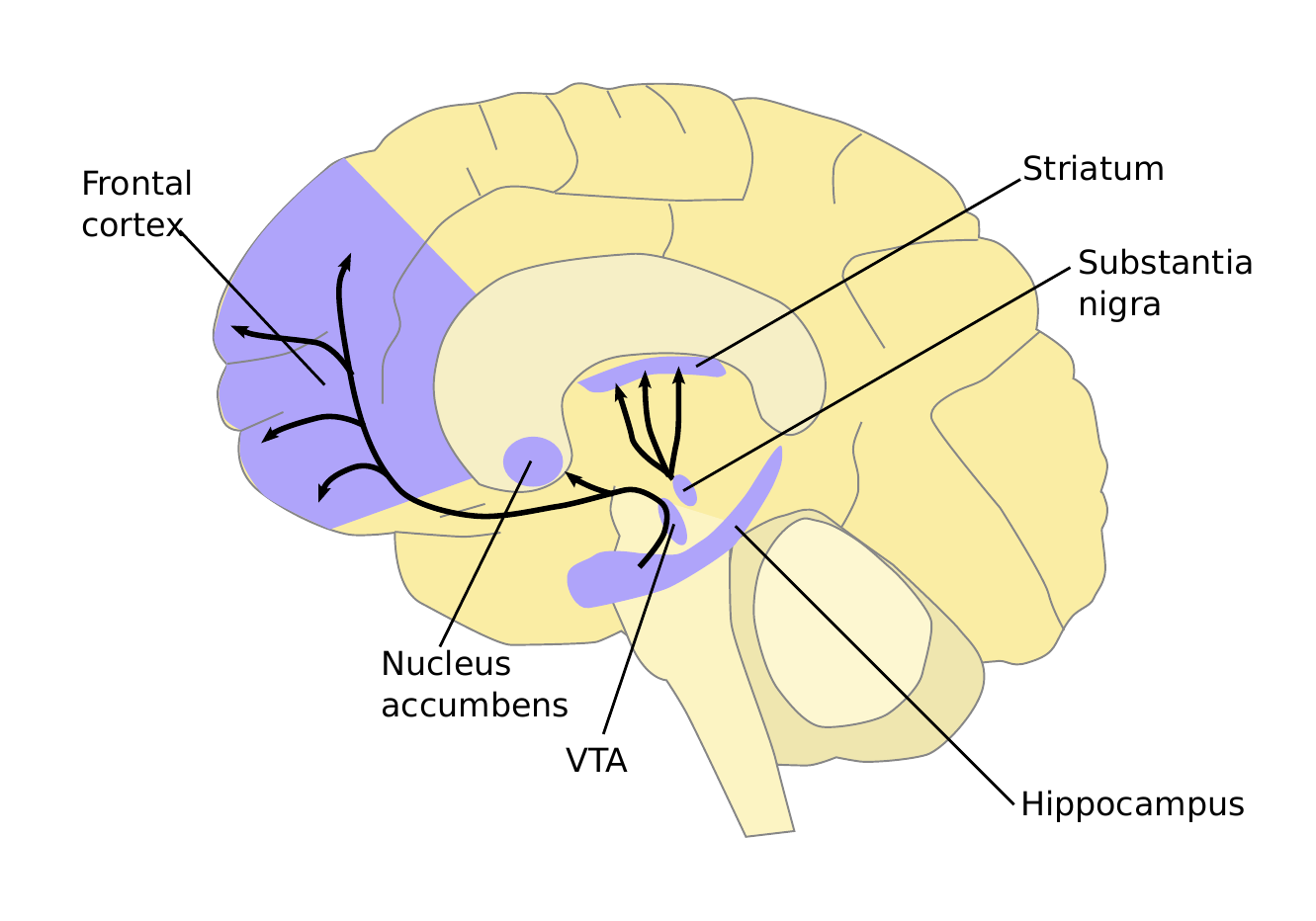}	
	\end{center}
	\caption{The main dopamine pathways in the human brain.}
	\label{fig:dopamine_pathways}
\end{figure}

STDP alone cannot account for the ability of an animal to perform reinforcement learning as it has no concept of reward.  It has been observed that dopamine can modulate synaptic plasticity when it is received within a 15-25s window~\cite{Otmakhova_1996}.  Dopamine in the brain is modulated by way of dopaminergic neurons, these being neurons that use dopamine as their neurotransmitter. The majority (90\%) of the dopaminergic neurons are contained in the ventral part of the mesencephalon~\cite{Chinta_2005}.  Within this region the most important area is the ventral tegmental area (VTA) which is vital in the reward circuitry of the brain.  Figure \ref{fig:dopamine_pathways} shows the main dopamine pathways in the brain.  The VTA has connections to the nucleus accumbens, which plays an important role in the perception of pleasure, as well as to the hippocampus which has been associated with planning complex behaviour, decision making and moderating social behaviour~\cite{Miller_2002}. Whenever the VTA is stimulated it produces a burst of spiking activity which in turn raises the level of dopamine in the nucleus accumbens and the hippocampus.

\begin{figure}[htbp]
	\begin{center}
	  	\includegraphics[width=0.4\textwidth]{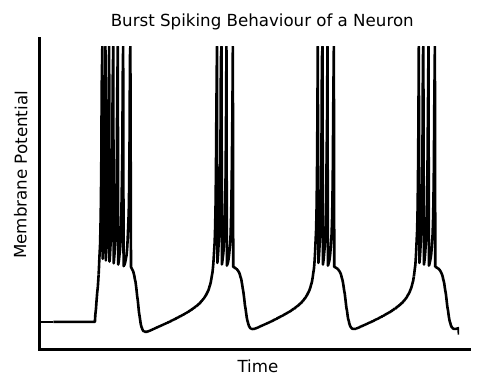}	
	\end{center}
	\caption{An example of a neuron exhibiting bursting behaviour, the neuron fires in bursts followed by periods of inactivity.}
	\label{fig:bursting}
\end{figure}

Dopaminergic neurons are characterized as having two different firing patterns.  In the absence of any stimulus they exhibit a slow (1-8Hz) firing rate~\cite{Ping_1996}, this is known as background firing.  When stimulated the dopaminergic neurons exhibit burst firing~\cite{Overton_1997}.  Burst firing is where neurons fire in very rapid bursts, followed by a period of inactivity.  This behaviour is illustrated in figure \ref{fig:bursting}.  One important, especially in the context of this paper, aspect of burst firing is that for many central synapses bursts facilitate neural transmitter release whereas single spikes do not~\cite{Lisman_1997}.  This means that stimulus induced firing of dopaminergic neurons will result in a spike in the level of dopamine, whereas background firing will not not have a significant effect on the level of dopamine.  This is a useful property for Pavlovian learning and its implications are explored further in section \ref{section:moving_dopamine}.

Matsumoto \& Hikosaka~\cite{Matsumoto2009} have recently shown, via experiments on monkeys, that the dopaminergic neurons do not function as a homogeneous group as previously thought.  Some neurons exhibit a reward predicting behaviour, with increased firing when a reward is received, and a drop in activity for negative stimulus.  However, other dopamine neurons have been found to exhibit an increased response for both a negative and a positive stimulus.  The spatial location of these two groups is distinct and it is likely that they play different roles in learning in the different ways that they are connected to the rest of the brain, though this mechanism is not yet understood fully.

In order for dopamine to be able to modulate synaptic plasticity over relatively large time-scales on the order of seconds it has been shown that synapses must have a form of synaptic tag that remembers the activity of the two neurons over a short time~\cite{Frey_1997,Papper_2011}.  The precise chemical process that underlies this synaptic tagging is not yet known.  Research has shown~\cite{Redondo_2011} that this synaptic tag, and long term potentiation are dissociable.  That is, the induction of synaptic potentiation creates the potential for a lasting change in synaptic efficacy, but does not commit to this change.  This is exactly what is needed for reinforcement learning, in standard machine learning algorithms for reinforcement learning the eligibility trace itself does not modify the value function (estimate of reward for a given environment state), but it is modified through a combination of the eligibility trace and received reward.

\begin{figure}[htbp]
	\begin{center}
	  	\includegraphics[width=0.6\textwidth]{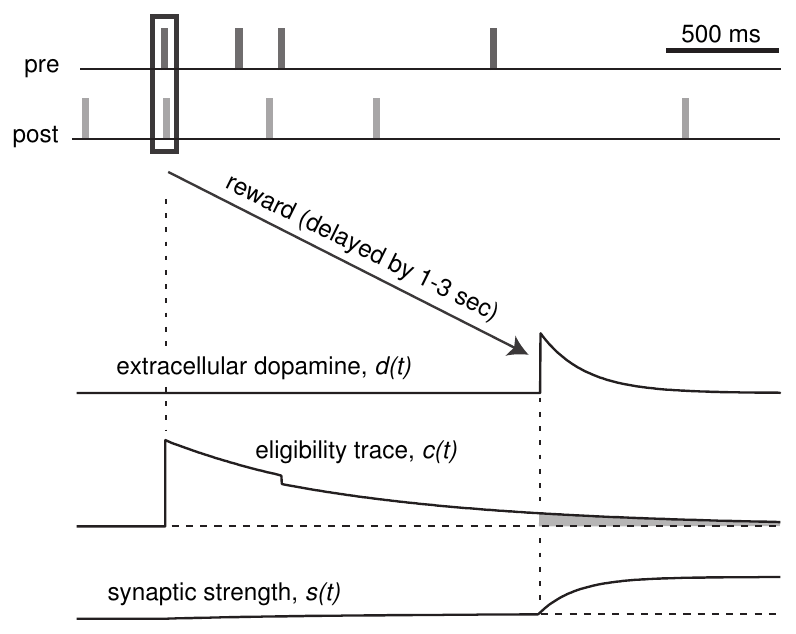}	
	\end{center}
	\caption{Figure from Izhikevich~\cite{Izhikevich_2007} showing how an eligibility trace can cope with a delayed reward.  When the pre then post synaptic neurons coincidentally fire, a delayed reward is received in the form of a spike in dopamine $d$.  This coincident firing increases the eligibility trace $c$, which slowly decays.  It is still positive when the reward is received, and as such the synaptic strength $s$ is increased.}
	\label{fig:da_mod_stdp}
\end{figure}

Izhikevich~\cite{Izhikevich_2007} has shown that dopamine modulated STDP can perform many aspects of reinforcement learning and can accurately model many of the features of learning observed in the brain.  One of the key problems in reinforcement learning is that the receipt of a reward is often delayed after the action, or neural pattern, that caused the reward.  This is known as the distal reward problem~\cite{hull_1943}.  STDP as described above works over the millisecond time-scale and so another mechanism is needed to account for this distal reward problem.  This is solved in DA modulated STDP by the inclusion of a variable that acts as a synaptic tag and has a direct correlation with the eligibility trace used in machine learning.  This synaptic tag is incremented in proportion to the amount of STDP that would normally be applied at a synapse and then decays over time, it is defined as:
\begin{equation}
\dot{c}=-c/\tau_c + STDP(\tau)\delta(t-t_{pre/post})
\end{equation}
Where $\tau_c$ defines the decay rate of the eligibility trace, and hence the time-scale over which DA modulation can work.  $\delta(t)$ is the Dirac delta function and has the effect of step increasing $c$ when STDP takes place.  The value of $c$ over time looks very similar to the eligibility trace of machine learning as shown in figure \ref{fig:eligibility_trace}.  This variable is then used to modify the synaptic weight using:
\begin{equation}
\dot{s}=cd
\end{equation}
Where $s$ is the synapse strength and $d$ is the current level of dopamine.  One important feature of this model is that it can deal with random firings in between the stimulus and the reward, this is due to the low probability of randomly occurring coincident firings between the stimulus and reward. As such the eligibility trace maintains the record of the pre-post firing that we wish to strengthen.  Figure \ref{fig:da_mod_stdp} shows an example of how the network is able to respond to a delayed reward.

\begin{figure}[htbp]
	\begin{center}
	  	\includegraphics[width=0.7\textwidth]{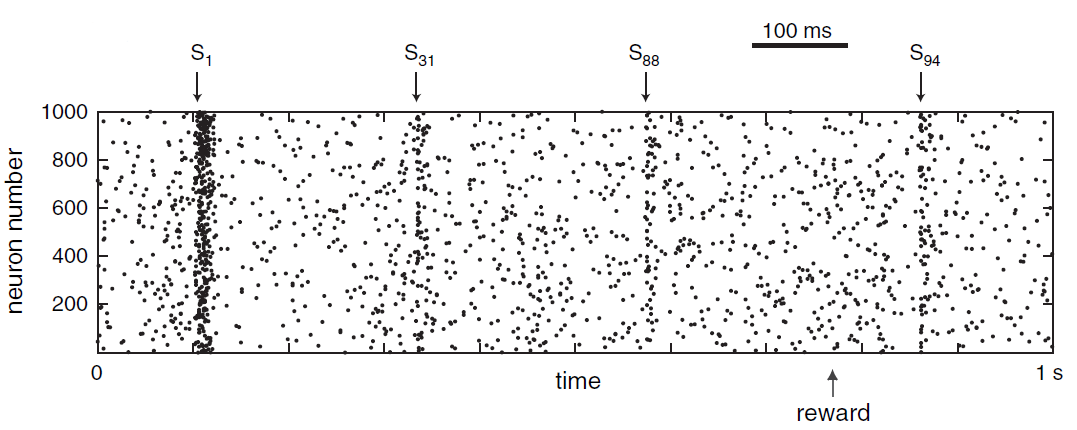}	
	\end{center}
	\caption{The response of a conditioned stimulus $S_1$ after being subject to DA modulated STDP, as compared with several unconditioned stimulus. Figure reproduced from Izhikevich~\cite{Izhikevich_2007}.}
	\label{fig:conditioned_stimulus}
\end{figure}

Using this model it has been shown that a network of neurons is able to correctly identify a conditioned stimulus embedded in a stream of equally salient stimuli, even when the reward is delayed enough such that several random stimuli occur in between the CS and the reward.  After training, the learnt response shows up as an increased firing when the CS is presented compared with the other stimuli, as in figure \ref{fig:conditioned_stimulus}.

In Pavlovian conditioning~\cite{pavlov_1927} repeated pairing of a conditioned stimulus with a reward-producing unconditioned stimulus can move the response from the US to the CS.  By the introduction of a set of dopamine releasing neurons similar to those in the ventral tegmental area in the brain, it was shown~\cite{Izhikevich_2007} that the DA modulated STDP model can exhibit this response shifting behaviour.  The US was initially connected with maximum weights to the dopamine neurons, simulating an US.  When a CS was repeatedly presented preceding this US there was an increase in the connections between the CS and the dopamine neurons, the dopamine response to the US stimulus was also reduced.  This effect is similar to the effect observed in vivo in monkeys and rats~\cite{Pan_2005}.  A modified version of this DA-modulated STDP process is used in this paper to control a robot. The ability to move the dopamine response is an important property when multiple, sequential, behaviours need to be learnt before a reward is received.

The above model demonstrates the ability to model several features needed for reinforcement learning.  However, the dopaminergic response to a stimulus does not correlate directly to the the difference between the expected reward and received reward that is crucial to classical machine learning RL algorithms.  This should manifest itself as a negative (drop in neural activity) response when a reward is expected, but not received.  This dip in neural activity has been observed in vivo~\cite{Ljungberg_1991}.  As the model developed by Izhikevich~\cite{Izhikevich_2007} did not include any sort of working memory it did not produce this dip in activity.

\subsection{Time-Based DA-modulated STDP}

Chorley \& Seth~\cite{Chorley_2011} have incorporated the ``dual-path'' model~\cite{Brown_1999,Tan_2008} into the DA modulated STDP model proposed by Izhikevich~\cite{Izhikevich_2007}.  In the dual path model there are two pathways between a stimulus and dopaminergic neurons, an excitatory pathway and a time-dependent inhibitory pathway.  The consequence of this is discussed in more depth, in the context of the new hybrid model, below.  The new model accounts for several key features of DA responses observed in the brain, as well as being closer to the standard machine learning algorithms.  These features, as defined by Chorley \& Seth~\cite{Chorley_2011}, are:
\begin{itemize}
\item DA neurons display phasic activation in response to unexpected rewards~\cite{Schultz_1990,Schultz_1998}.
\item These DA neurons display phasic responses to reliably reward-predicting
stimuli, yet do not respond to stimuli which are themselves predicted by earlier stimuli~\cite{Ljungberg_1992}.
\item Reward-related DA responses reappear if a previously
predictable reward occurs unexpectedly~\cite{Ljungberg_1992}.
\item DA neurons display a brief dip in activity at precisely the
time of an expected reward, if that reward is omitted~\cite{Ljungberg_1991}.
\end{itemize}

\begin{figure}[htbp]
	\begin{center}
	  	\includegraphics[width=0.7\textwidth]{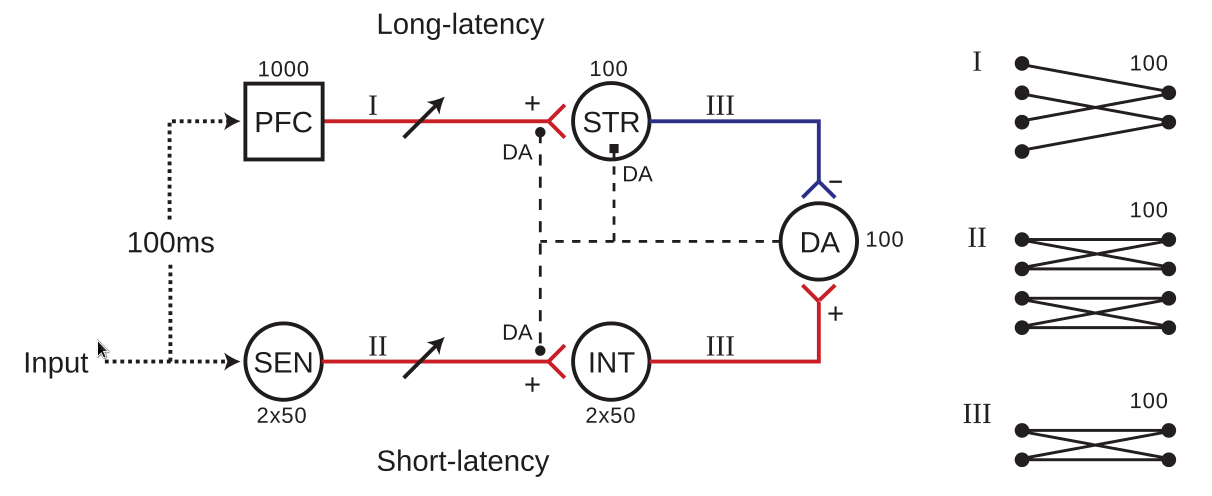}	
	\end{center}
	\caption{The network architecture used by Chorley \& Seth~\cite{Chorley_2011}, red lines represent excitatory connections and blue represent inhibitory connections.  When neurons in the DA module fire then dopamine is released which causes STDP of the PFC$\rightarrow$STR and SEN$\rightarrow$INT pathways.  The mean firing rate of the STR module is also modulated by the amount of dopamine.}
	\label{fig:da_dual_path_wiring}
\end{figure}

An overview of the network architect that was used is given in figure \ref{fig:da_dual_path_wiring}.  The key feature that allows the network to display a dip in DA firing when a reward is expected, but not received, is that that the PFC has a basic working memory.  This is implemented as a sequence of distinct neural firing patterns that occur in sequence after a stimulus is presented.  This means that when an untrained stimulus (CS) is followed by a reward producing stimulus (US), the corresponding phasic release of DA will strengthen connections between the current PFC time-dependent firing pattern and the STR.  Therefore the next time the CS is presented, and after the same time period, there will be an increase in firing in the STR and hence an increased inhibition of the DA cells.

\subsection{Plasticity and Stability}
\label{section:plasticity_and_stability}
One of the problems with Hebbian learning through STDP is that it can cause runaway feedback loops.  For example, if two neurons fire in sequence then their synaptic weight will be increased.  This means that in future they are more likely to fire in sequence, and as such they are more likely to have their synaptic weight further increased.  Without any kind of stability mechanism this process will quickly cause the weights to go towards infinity.  One simple method that is often used~\cite{Izhikevich_2007,Chorley_2011} is to restrict the range of values that the synaptic weights can take, for example limiting an excitatory synapse to the range 0-4mV.  This prevents the synaptic weight from becoming negative, and hence becoming an inhibitory synapse, whilst also preventing the synaptic weight increasing forever.

This synaptic weight capping does not, however, prevent synapses from becoming quickly potentiated to their maximum allowable value.  In the brain it has been shown that highly active neurons have their excitability decreased over time, and highly inactive neurons have their excitability level increased~\cite{Turrigiano_1999}.  This is achieved by processes that modify the tonic level of Ca$^{2+}$ within the neurons.  This raises or lowers the membrane potential of the neuron, and hence increases or decreases the neurons excitability.  It has also been shown that within populations of neurons a global mechanism for maintaining homeostatic plasticity is used~\cite{Turrigiano_1998}.  This mechanism has the effect that if the firing rate of a population of neurons is increased then the strength of all synapses within the population is decreased, and equivalently if the firing rate decreases then the strength of all synapses is increased.

A consequence of these homeostatic mechanisms is that they can foster competition between competing behaviours.  For example, if a group of neurons has the possibility of representing two distinct behaviours and the neurons corresponding to behaviour 1 are highly active, whereas the neurons for behaviour 2 aren't as active.  Then, due to the high firing rate, all the synaptic weights will be reduced, which will have the result of further reducing the activity of behaviour 2.  Since connected neurons with high firing rates will tend to increase their synaptic weights faster through STDP, the neurons corresponding to behaviour 1 will have their synaptic weights increased faster than behaviour 2.  This process will continue until only the neurons corresponding to behaviour 1 are firing.

\section{Neural encoding}

In the brain there needs to be a mechanism to convert external stimulus into a neural firing pattern, and to convert a neural firing pattern into an output of the network.  There are three main neural encoding mechanisms that have been observed in the brain, these being rate coding, temporal coding and population coding.

In rate coding the strength of the stimulus or output is encoded in the firing rate of a population of neurons.  A stronger stimulus emits a faster firing rate, and equivalently a faster firing rate will result in a stronger output of the network, such as joint movement.  This encoding mechanism was first observed by Adrian \& Zotterman in 1926~\cite{Adrian_1926}.  Rate coding is useful when an increase in the value of a stimulus should result in a correlated increase in the response to the stimulus.

Population coding involves the stimulation of different neurons for different values of the external stimulus.  For example, the direction of movement of the eye could be represented by a network of 180 neurons with each neuron coding for a direction range of 2 degrees.  In practice there is usually some overlap between the neural firing with a Gaussian distribution in firings being common, this allows a wider range of values to be represented as well as being more robust to noise.  Population coding is used in various parts of the brain, for example in coding the direction of movement of an observed object~\cite{Maunsell_1983}.  Population coding has the advantage of being able to react to changes in the external stimulus faster than rate coding.

In temporal coding the external stimulus is represented by the precise spike timings in a population of neurons, in this way a greater range of parameters can be represented by the same group of neurons than could be represented with rate coding~\cite{Zador_1995}.

In the context of learning taxis behaviour for a simple robot rate coding of sensor inputs/motor outputs is an appropriate mechanism as it allows the strength of response to an input stimulus to be correlated with the strength of the input stimulus to be learnt more easily.

\section{Taxis}
\label{section:taxis}
In the animal world, one of the simplest forms of behaviour is that of taxis.  Where taxis is defined as motion towards or away from a directed stimulus.  Examples include phototaxis, movement directed by light, and chemotaxis, movement directed by a chemical gradient.  This behaviour has advantages such as food foraging and poison avoidance.

Through a series of thought experiments Braitenberg~\cite{Braitenberg_1984} showed how taxis behaviour, as well as more complex behaviours, could be implemented in a simple robot.  The robots were endowed with two sensors and two motors, attraction behaviour can then be achieved through connecting the left sensor the right motor, and the right sensor to the left motor.  Avoidance can be achieved through connecting the left sensor to the left motor and the right sensor to the right motor.  A version of these simple Braitenberg vehicles will be used in this paper to demonstrate reinforcement learning.

\section{Spiking Neural Network Controlled Robots}

A variety of research has been conducted into using SNN's to control robots.  Initially this was done by manually setting the weights of the network~\cite{Indiveri_1999,Lewis_2000}.  For example Lewis et al.~\cite{Lewis_2000} used a very small network of 4 neurons to control an artificial leg.  The complexity of SNN's means that manually setting the synaptic weights is only feasible for very small networks.  Various mechanisms for training the weights of a SNN have been used, these are outlined below.

\subsection{Robot Training using Genetic Algorithm's}
One such solution to the problem of determining the synaptic weights is to use genetic algorithms to determine the weights. Genetic algorithms learn the best solution to a problem by using a population of possible solutions.  These possible solutions are then evaluated to give some measure of the fitness of each solution in solving the given problem.  Highly fit solutions are then combined to create new individuals which are replaced with unfit solutions and the process is repeated.  

Hagras et al.~\cite{Hagras_2004} have used genetic algorithms to train a SNN controlled robot such that it exhibits wall following behaviour.  Evolving SNN's for robot control work well when the environment is mostly static and the optimal solution does not change over time.  In this case the training can be done off-line in a simulated environment with a large number of robots.  However in dynamic environments or where we are restricted to online learning with a single robot then GA's do not perform as well.

\subsection{Robot Training using STDP}
Another solution to determining the synaptic weights is to modify the weights over time using STDP as described in section \ref{section:stdp}.  Bouganis \& Shanahan~\cite{Bouganis_2010} effectively used STDP to train a robotic arm to associate the direction of movement of the end effector with the corresponding motor commands.  This was achieved by a training stage during which the motors were randomly stimulated along with the actual direction of movement of the end effector.  STDP was able to modify the connection weights in the network such that if you subsequently stimulated the neurons corresponding to a desired end effector direction of movement, at a given position, then the arm would move in that direction.  This is an effective method for learning associations, however it is limited in that it cannot cope with goal oriented tasks where an agent has to maximize some reward.  It also requires that the action and response happen simultaneously, in many real world situations this is not the case.  For example if we want the robot to reach out and touch an object there will be a delay between the initial movement, and the act of touching the object.

\subsection{Robot Training using Reward Modulated STDP}
Several experiments have been conducted into training an SNN controlled robot using reward-modulated STDP.  A common learning task used to demonstrate reinforcement learning is that of the Morris water maze.  This learning paradigm consists of placing the agent (originally a rat) in a non-transparent liquid, in the environment there is a hidden submerged platform.  The agent must find this platform in order to escape the unpleasant experience.  When the experiment is run with a rat the platform is initially found by chance.  But in subsequent runs the rat learns to navigate directly towards the platform.

Vasilaki et al.~\cite{Vasilaki_2009} used a spiking neural network to control a robot in the Morris water maze task. The robot was able to learn to successfully learn to navigate directly towards the hidden platform when placed close to the platform.  The level of reward was simulated as an external parameter that was artificially set when the platform was found.  The robot only received a reward when it reached the platform, as such the robot was only able to learn the correct behaviour within a small radius around the hidden platform due to the eligibility trace decaying over larger distances.  In this paper the level of reward will be embedded within the network and as such allow the robot to learn over a much larger time frame.

\begin{figure}[htbp]
	\begin{center}
	  	\includegraphics[width=0.7\textwidth]{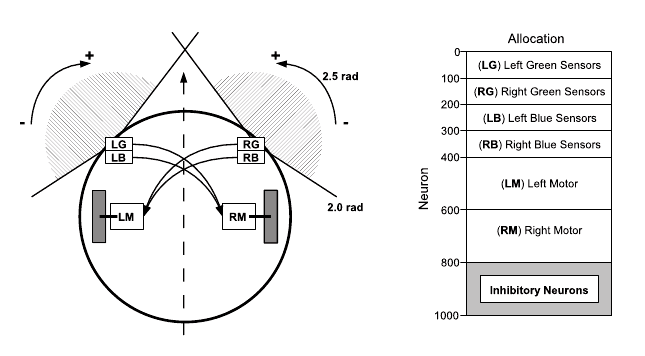}	
	\end{center}
	\caption{The experimental set-up used by Chorley \& Seth~\cite{Chorley_2008}.}
	\label{fig:chorley_robot}
\end{figure}

In the context of learning taxis behaviour using reward modulated STDP Chorley \& Seth~\cite{Chorley_2008} implemented a simple Braitenberg vehicle~\cite{Braitenberg_1984} that was able to learn to distinguish between two stimuli, one of which would elicit a reward.  The experimental set-up of the robot is shown in figure \ref{fig:chorley_robot}.  It is important to note that in the experimental set-up the robot is already wired for taxis attraction behaviour, with each sensor only being connected to the opposite motor. The learning phase consists of learning the strength of the taxis response to two different stimuli.  The robot was able to learn to elicit a stronger taxis response to the reward-giving stimuli, and it was also able to change its behaviour when the reward was switched from one stimuli to another.  Once again the level of reward was simulated as an external parameter rather than being embedded in the network itself.

In this paper we will extend the work of Chorley in several key ways, these being:
\begin{itemize}
\item In our implementation the robot will not be explicitly wired for attraction behaviour, the robot will have the ability to learn either attraction or avoidance behaviour.
\item By incorporating the level of dopamine into the network directly, we will aim to show how sequences of behaviour can be learnt.
\item Through the use of a form of hunger our implementation will be able to deal with situations in which there is no fixed optimal behaviour for an environment, and a dynamic set of behaviours must be learnt.
\end{itemize}

\chapter{Methodology}
\label{ch:methodology}
In this paper a simple SNN controlled Braitenberg vehicle was simulated in environments consisting of food, poison and containers.  The robot was subject to DA-modulated STDP and had to learn the correct behaviour to collect food items.  The precise implementation details are outlined in this chapter.

Due to the fact that robot and environment are simulated, all distances and sizes are relative and do not have a real world unit associated with them, by convention we will take the default unit to be centimetres.

\section{Environment}
The learning task that was used to test reinforcement learning using spiking neural networks was that of food collection and poison avoidance in a simulated environment by a simple robot.  There are four types of objects that can exist in the environment, these being:
\begin{itemize}
\item Food - Food items are discs with a radius of 2.4cm, when a food item is collected it is replaced by another food item in a random location.
\item Poison - Poison is identical to food, except that it induces a negative dopamine response (see section \ref{section:meth_plasticity}).
\item Food Container - A food container is a disc with radius of 14cm which contains a single food item randomly located within it.  Food items cannot be sensed from outside a food container, other containers cannot be sensed from within a container.
\item Empty Container - This is identical to a food container except that is does not contain any food items, as well as being sensed differently by the robot (see section \ref{section:robot_and_sensors}).
\end{itemize}

The environment has no walls, and has the characteristic of a torus.  That is, if the robot moves over the right edge of the environment it will reappear at the left of the environment and the same for the top and bottom 'edges' of the environment.  In this way the robot does not have to deal with collision avoidance.

\begin{figure}[htbp]
     \begin{center}
        \subfigure[Food-Only Environment]{
            \label{fig:foodonly_env}
            \includegraphics[width=0.58\textwidth]{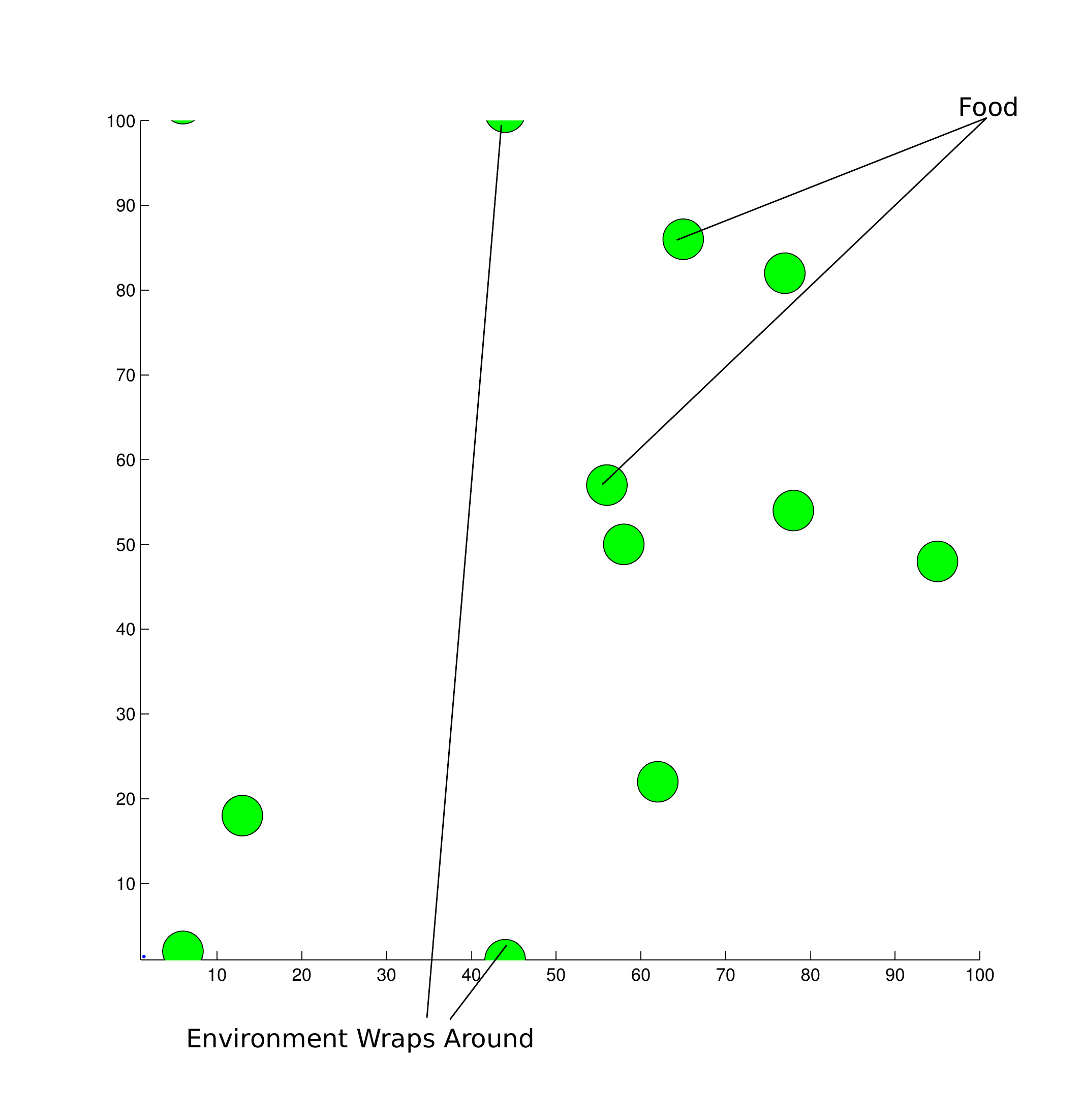}
        }
        \subfigure[Food-Container Environment]{
           \label{fig:foodcontainer_env}
           \includegraphics[width=0.58\textwidth]{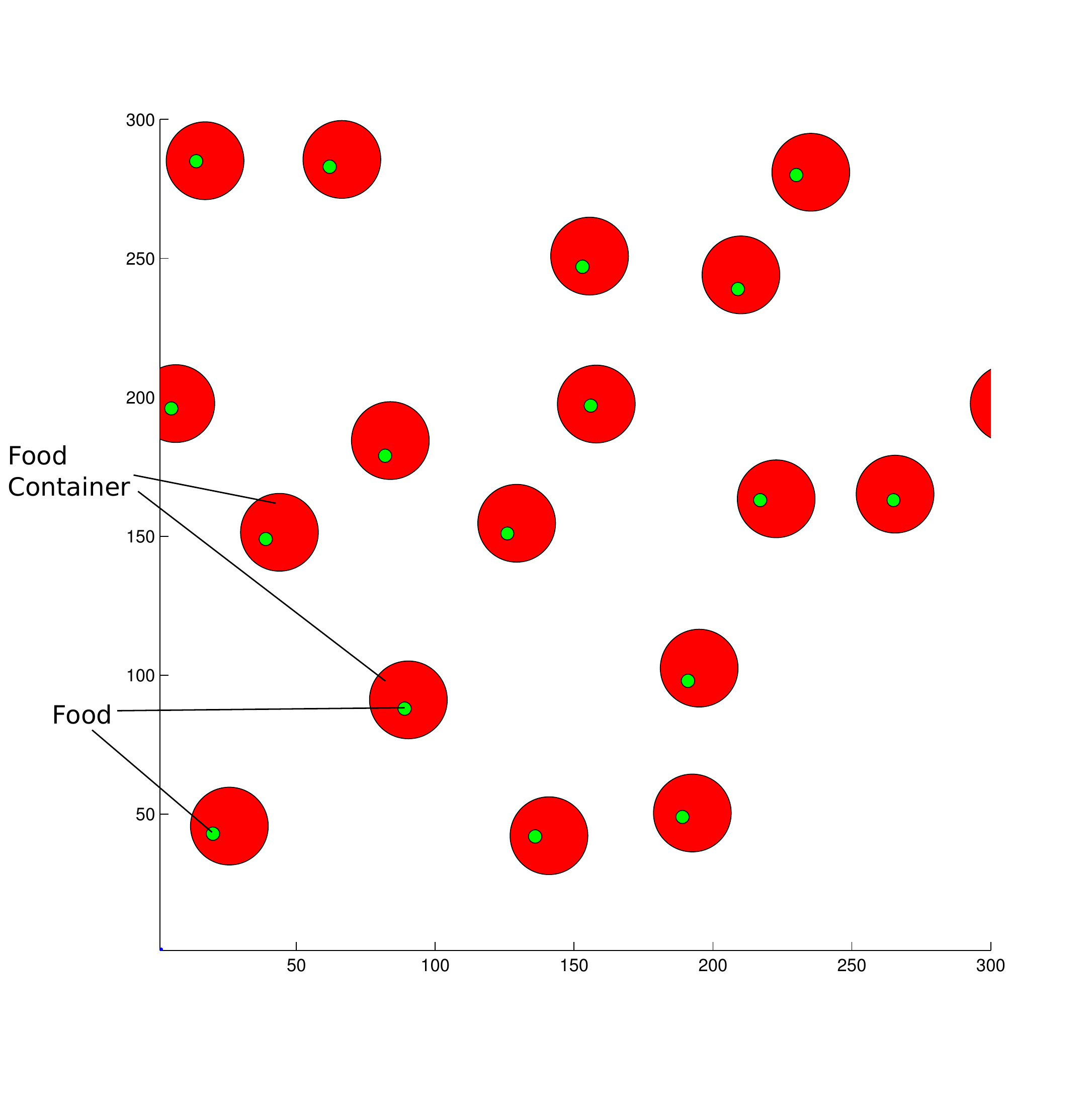}
        }
    \end{center}
    \caption{\subref{fig:foodonly_env} shows a 100cmx100cm environment containing randomly placed food.  \subref{fig:foodcontainer_env} shows a 300cmx300cm environment with food items that are positioned within containers.  Food cannot be sensed from outside the container, other containers cannot be sensed when inside a container.}
   \label{fig:environments}
\end{figure}

There are two main variations in the environment.  In the simplest case (figure \ref{fig:foodonly_env}) the environment consists of randomly located food objects.  Whenever the robot reaches an item of food the piece of food is moved to another random location within the environment.  This environment set-up will be referred to as the food-only environment. In the second type of environment food is contained within containers (figure \ref{fig:foodcontainer_env}).  The food cannot be sensed until the robot is within a container, other containers cannot be sensed when the robot is within a container.  This forces the robot to have to learn to turn towards food containers, something that does not directly give it a reward.  When food is collected the food and container are moved to a random location in the environment.  This environment will be referred to as the food-container environment.  An extension to this environment includes empty containers which do not contain food items.  

\section{Robot and Sensors}
\label{section:robot_and_sensors}
\begin{figure}[htbp]
	\begin{center}
	  	\includegraphics[width=0.4\textwidth]{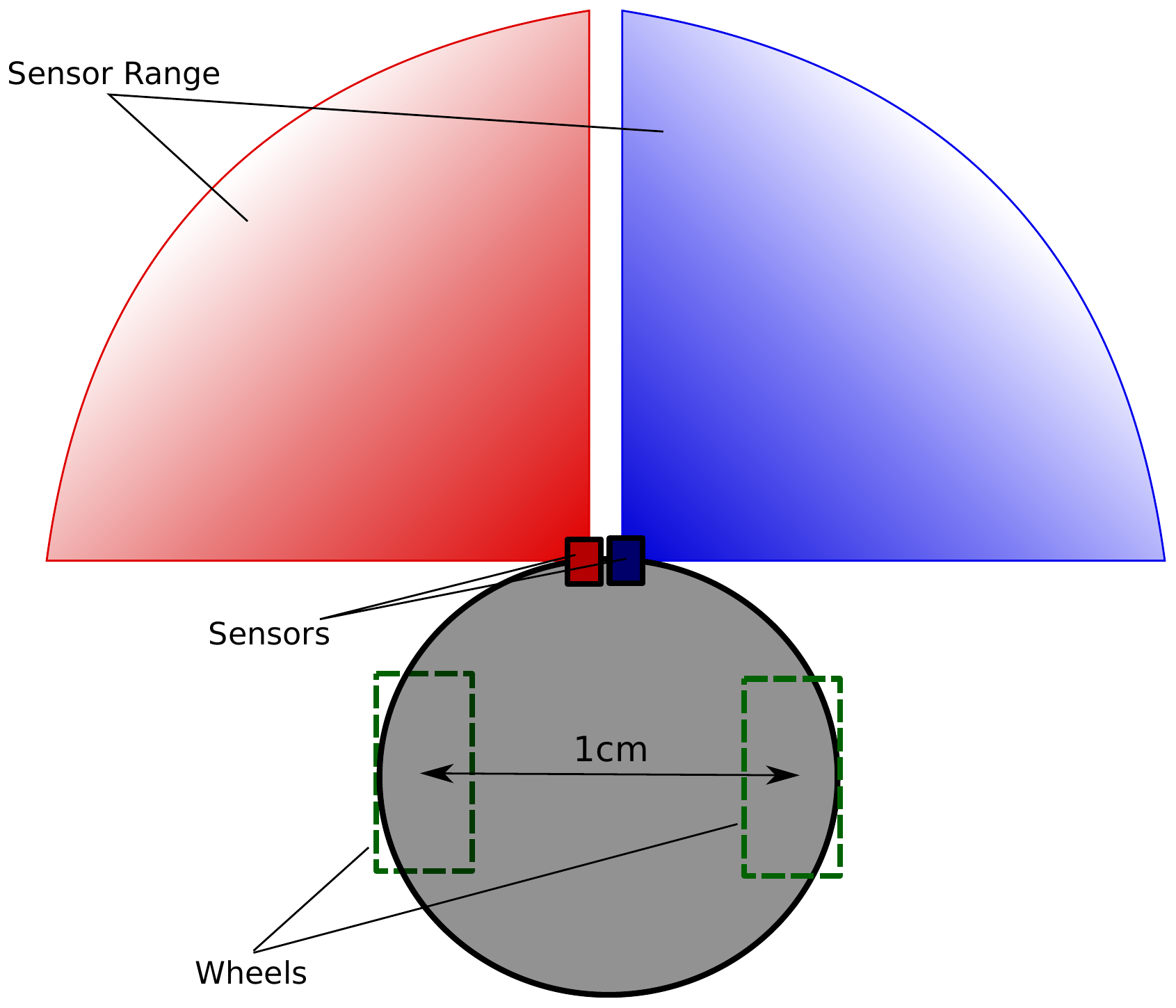}	
	\end{center}
	\caption{Structure of the robot.}
	\label{fig:robot}
\end{figure}

The robot consists of a circular body with two wheels positioned 1cm apart on either side of the robot.  To ensure that the robot is always moving, the wheel motor velocities are restricted to the range 25cm/s - 31.2cm/s. This gives the robot a minimum turning circle of radius 4.5cm.  The robot is equipped with two types of sensor, range-sensors and touch-sensors.  The range-sensors come in pairs, one left and one right, each covers a range of $\pi/2$ from directly in front of the robot to the left and right as in figure \ref{fig:robot}.

The touch-sensors elicit an instantaneous response that lasts for 1ms whenever the robot comes into contact with the object for which they are sensitive.  The response of the range-sensors is linear with the distance to the object.  So an object directly in front of the robot will elicit a sensor response of one, whereas an object at the edge of the sensor's range will produce a response of zero in the sensor.  If there is more than one object in a sensor's range then only the closest object is sensed.  This prevents the robot being distracted by objects further away before responding to closer ones.  

\begin{figure}[htbp]
	\begin{center}
	  	\includegraphics[width=0.4\textwidth]{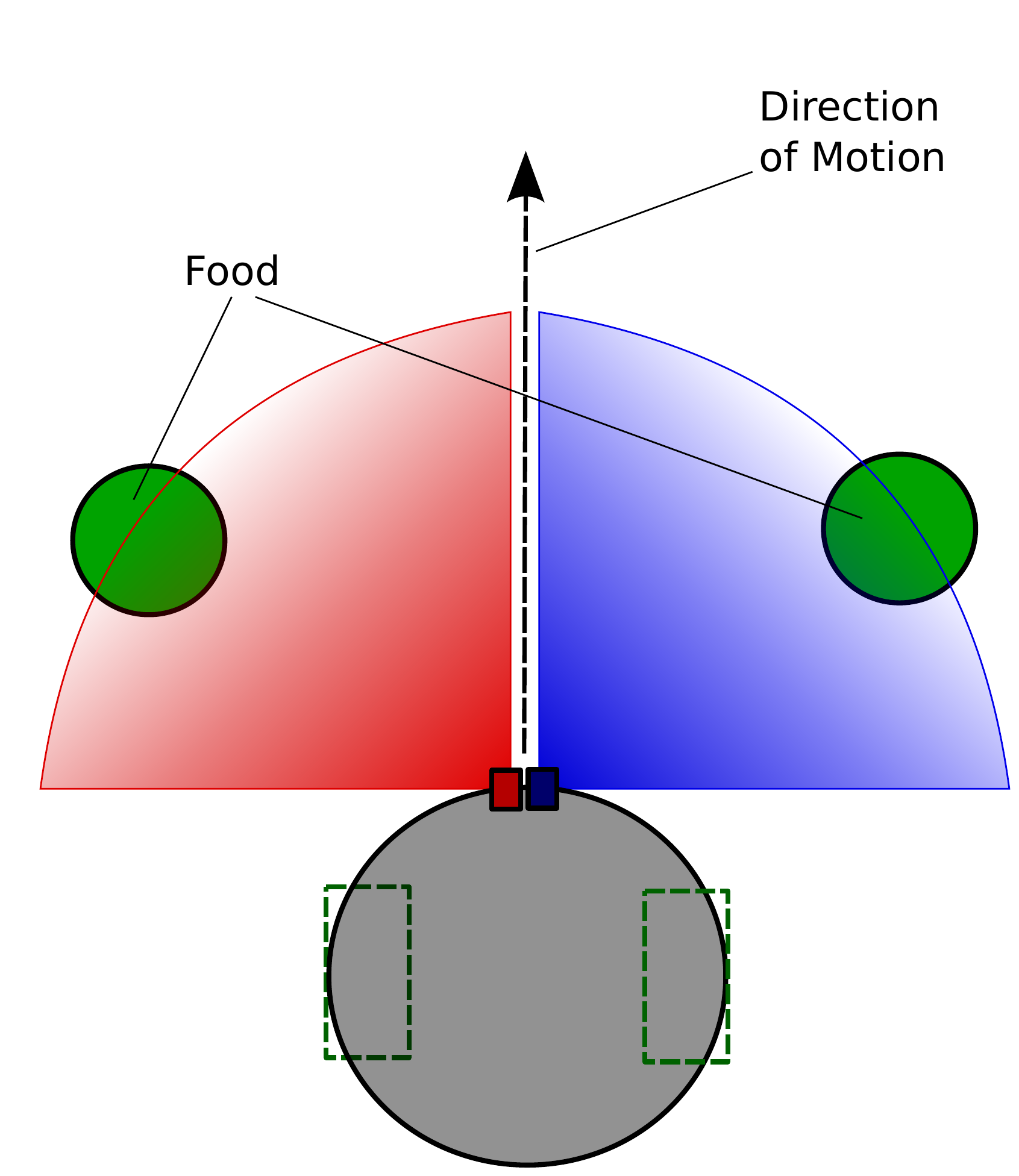}	
	\end{center}
	\caption{Without a winner-takes-all mechanism both sensors will respond equally and the robot will increase both left and right motor speeds, ultimately driving straight between the food items.}
	\label{fig:robot_wta}
\end{figure}

A winner-takes-all mechanism was implemented between the two range-sensors.  Such that if both sensors detect an object at any given time, then only the sensor with the strongest value elicits a response.  This winner-takes-all mechanism is useful in situations where two food items are located at equal distances to the left and right of the robot, as illustrated in figure \ref{fig:robot_wta}.  Without a winner-takes-all mechanism both sensors will sense food items and as such the robot will increase its left motor velocity to turn right, it will also increase its right motor velocity to turn left.  The effect of which will be that the robot drives straight between the two food items.  It has been shown that a winner-takes-all mechanism is biologically plausible~\cite{Gupta_2009}.  For simplicity this winner-takes-all mechanism was implemented in the sensors, an alternative implementation would be to implement it directly in the wiring of the robot's neural network.

Table \ref{tab:sensors} gives an overview of all the sensors that the robot can have.  In the food-only environment the robot has two food range-sensors and a food touch-sensor.  In some environments the food items are replaced with poison, to demonstrate the ability of the robot to modify its behaviour.  In this case the same food range and touch sensors are used for sensing poison.

In the environment with food containers the robot has an additional container range-sensor, this sensor reacts to both food-containers and empty-containers.  It also has a food-container touch-sensor, this sensor only elicits a response when a food-container is entered (no response is elicited whilst the robot is inside the container).  Equivalently it has an empty-container sensor, which only reacts to entering containers which don't contain food items.  This set-up is useful in demonstrating that the robot can differentiate between the reward predicting stimulus of the food-container touch-sensor, and the non-reward predicting stimulus of the empty-container touch-sensor.

\begin{table}[htbp]
\centering
\begin{tabular}{|p{4cm}|l|l|p{3cm}|p{4cm}|}
\hline
Object Sensed & Type & Range & Available in Food-Only Environment & Available in Food-Container Environment \\ \hline
Food/Poison & Range & 30cm & Yes & Yes \\ \hline
Containers & Range & 60cm & No & Yes \\ \hline
Food & Touch & 0cm & Yes & Yes \\ \hline
Food-Containers & Touch & 0cm & No & Yes \\ \hline
Empty-Containers & Touch & 0cm & No & Yes \\ \hline
\end{tabular}
\caption{The different types of sensor that the robot can have, and in which environment the robot is equipped with these sensors.}
\label{tab:sensors}
\end{table}

\section{Neural Model}
The Izhikevich model was used to simulate the individual neurons.  An overview of this model is given in section \ref{section:izhikevich}.  The dynamics of the neuron are modelled by two differential equations:
\begin{equation}
\frac{dv}{dt}=0.04v^2+5v+140-u+I
\end{equation}
Where $I$ is the inbound current from pre-synaptic spikes (weighted by the strength of the synapse), $v$ is the membrane potential and $u$ is the recovery variable that determines the refractory period of the neuron and is modelled as:
\begin{equation}
\frac{du}{dt}=a(bv-u)
\end{equation}
The parameters of the model were set for excitatory neurons as:
\begin{equation}
a=0.2, b=0.02, c=-65+15r^2, d=8-6r^2
\end{equation}
Where $r$ is a random variable in the range [0,1].  This produces neurons with a regular spiking pattern (see figure \ref{fig:izhik_model}).  For inhibitory neurons the parameters were set to:
\begin{equation}
a=0.1, b=0.2, c=-65+15r^2, d=8-6r^2
\end{equation}
Which results in neurons which exhibit fast spiking behaviour.  These neural spiking patterns are consistent with observed behaviour in the brain, and previous neural models~\cite{Izhikevich_2003}.

\subsection{Numerical Approximation}
The differential equations given by the Izhikevich model were simulated using the Euler method of numerical approximation.  Given a differential equation such that $y'(t)=f(t,y(t))$, if we know the value of $y(t)$ for some time $t$ then, by the Euler method, we can approximate the value of $y(t+\delta t)$ for some small time step $\delta t$ as:
\begin{equation}
y(t+\delta t)=y(t)+\delta t.f(t,y(t))
\end{equation}
This formula can be applied repeatedly to approximate $y(t)$ over time.  In our simulations we used $\delta t=$0.5ms for approximating $v(t)$ and $\delta t=$1ms for approximating $u(t)$.  This is consistent with Izhikevich's original paper~\cite{Izhikevich_2003}.

\section{Phasic Activity}
\label{section:meth_phasic_activity}

The properties of individual neurons mean that if a group of neurons is interconnected and then stimulated with a low baseline current then the group of neurons will tend to fire in phase; synchronous firing followed by synchronous inactivity.  The frequency with which the neurons oscillate depends on various parameters of the neurons and network, phasic activity has been observed in the brain over a wide range of frequencies ranging from 0.02Hz to \textgreater 100Hz~\cite{Timofeev_2006}.  In our model the inputs and outputs to the network are updated every 70ms (corresponding to an update frequency of 14Hz), in the following sections we will describe the exact method by which the sensors and motors are encoded/decoded.  For simplicity this 14Hz update frequency was explicitly implemented outside of the neural network.  

One possible way in which this oscillatory property could be embedded in the network directly would be to have a separate group of interconnected inhibitory neurons with a baseline level of firing such that the firing rate of the group oscillates at 14Hz.  If this inhibitory population is connected to the relevant sensor and motor neurons then it would force them to fire with 14Hz frequency as well, and as such simulate the update model that was used in this paper.

The dynamics of STDP mean that, without phasic activity, two populations of connected neurons will have their synaptic weights depressed over time.  There is just as much probability of a pre-synaptic neuron firing before or after a post-synaptic neuron.  As the parameters of STDP were set to favour long term depression, the synapses would be depressed.  Through the use of phasic activity the network exhibits clear sense-respond behaviour.  For example, if the left sensor is stimulated and, for the rest of the active phase of the neurons, the right motor neurons are active, then we can say the right motor action was a response to the left sensor stimulus.  The eligibility trace for left sensor to right motor synapses would be high at this point, as the left sensor neurons fired before the motor neurons.  As such, if a reward was subsequently received then the robot would learn attraction behaviour.

\section{Sensorimotor Encoding}
\subsection{Sensor to Neuron Encoding}
\begin{figure}[htbp]
	\begin{center}
	  	\includegraphics[width=0.99\textwidth]{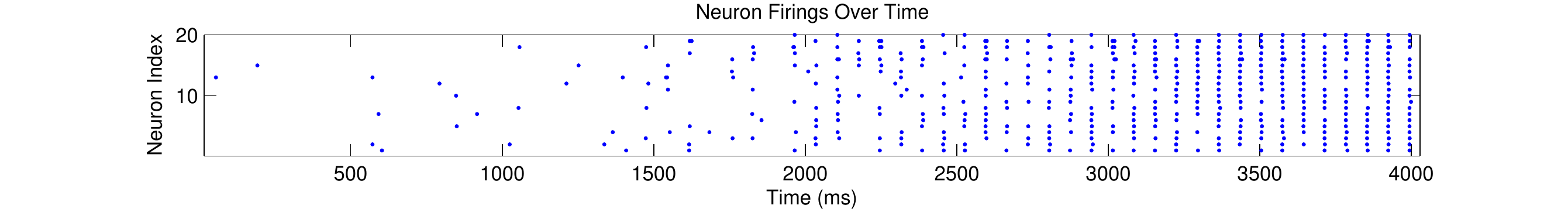}	
	\end{center}
	\caption{The firings in a group of 20 sensor neurons as the sensor value is linearly increased from zero to one.  The neurons are stimulated every 70ms.}
	\label{fig:sensor_encoding_firings}
\end{figure}

Both the food and container range-sensors return results in the range [0,1]. Every 70ms the corresponding sensor neurons are stimulated for 1ms with a current taken from a Poisson distribution with mean of $sensor\_value*30$mA.  The effect this stimulus current has on the firing rate of a population of sensor neurons is shown in figure \ref{fig:sensor_encoding_firings}.

The food touch-sensor and the container touch-sensor elicit an instantaneous response in the corresponding neurons.  This is achieved by stimulating them for 1ms with a current taken from a Poisson distribution with a mean of 12mA.

\subsection{Motor Velocity Calculation}
The simulation is updated every 70ms, the total firings during this 70ms for the left and right motor neurons ($f_\mathit{left}$ and $f_\mathit{right}$ respectively) are first calculated.  The motor velocities are then calculated, for example for the left motor, using the following formula:
\begin{equation}
  v_\mathit{left} = \left\{
  \begin{array}{l l}
  	v_\mathit{max} & \quad \textrm{if $f_\mathit{left} > f_\mathit{right}$}\\
  	v_\mathit{min} & \quad \textrm{if $f_\mathit{left} < f_\mathit{right}$}\\
  	\frac{v_\mathit{max}+v_\mathit{min}}{2} & \quad \textrm{otherwise.}\\
  \end{array} \right.
\end{equation}
Where $v_\mathit{left}$ is the output velocity of the left motor and $v_\mathit{max}$ and $v_\mathit{min}$ are the maximum and minimum allowable velocities of the motors, which are set to 31.2cm/s and 25cm/s respectively.  The equation for the right motor is equivalent, with $\mathit{left}$ and $\mathit{right}$ terms switched.  Note that this formula implements a winner-takes-all mechanism between the motors.  If one group of motor neurons has a greater firing rate then the corresponding motor will be set to maximum velocity, and the other motor set to its minimum velocity.  This helps with robot exploration as it reduces the probability of the robot driving in a straight line.  Also this winner-takes-all mechanism helps with learning attraction/avoidance behaviours as it amplifies the result of any slight difference in synaptic weights between the sensors and motors.

The robot was restricted to run at half-speed whilst in food-containers or empty-containers, such that its maximum and minimum velocities were set to 15.6cm/s and 12.5cm/s.  This was to increase the time between the reward predicting stimulus and the reward, as discussed in section \ref{section:moving_dopamine}.

\section{Network Architecture}
\label{section:network_architecture}
\begin{figure}[htbp]
	\begin{center}
	  	\includegraphics[width=0.8\textwidth]{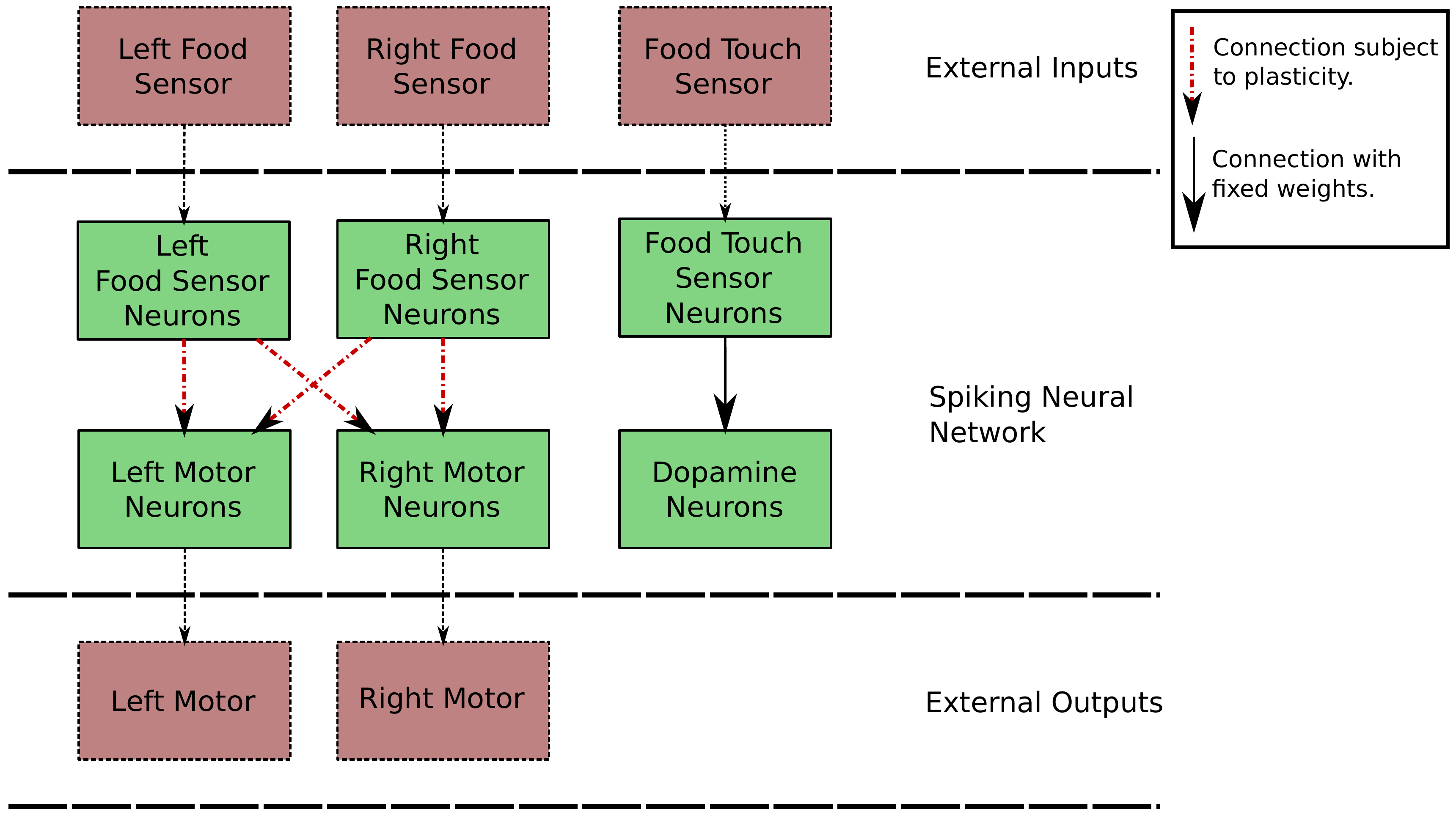}	
	\end{center}
	\caption{The architecture of the robot's neural network for the food-only environment.}
	\label{fig:simple_neural_network}
\end{figure}

\begin{figure}[htbp]
	\begin{center}
	  	\includegraphics[width=0.99\textwidth]{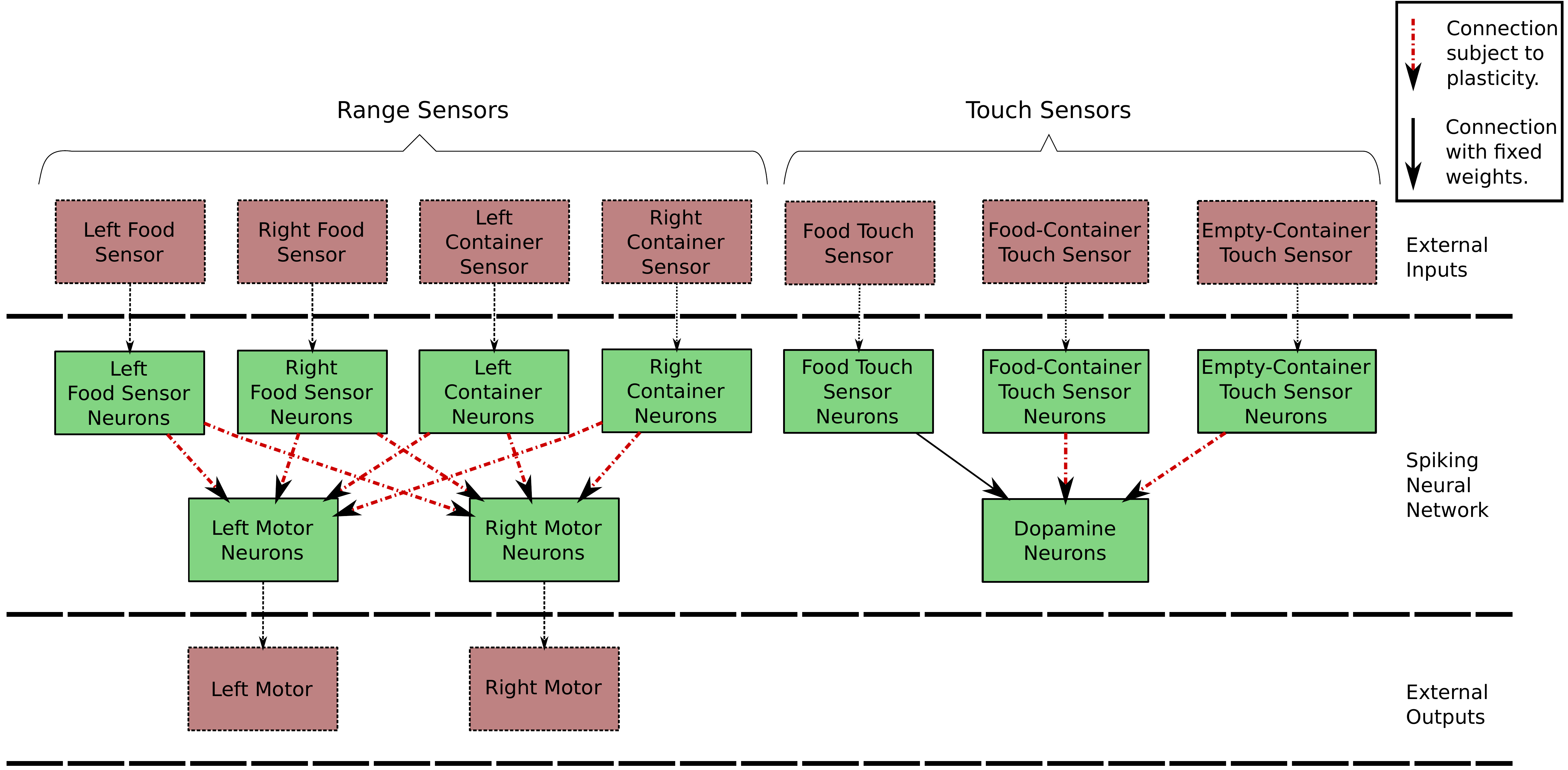}	
	\end{center}
	\caption{The architecture of the robot's neural network for the food-container environment.}
	\label{fig:complex_neural_network}
\end{figure}

The robot was controlled via a spiking neural network consisting of connected groups of neurons.  Figure \ref{fig:simple_neural_network} shows the connections in the neural network for the robot in the food-only environment.  This is a subset of the connections for the robot in the food-container environment which is shown in figure \ref{fig:complex_neural_network}.  All range-sensor neurons are connected to motor neurons with 85\% probability, that is, there is an 85\% chance that any individual range-sensor neuron is connected to any individual motor neuron.  All touch-sensor neurons are connected to dopaminergic neurons with 10\% probability.  The synaptic weights between food touch-sensor neurons and dopaminergic neurons is fixed and set to 3mV.  This high synaptic weight forces a strong response in the dopaminergic neurons whenever food is eaten.  Unless otherwise specified all other synapses are plastic and have their initial weights set to zero.

In addition to the neurons specified in these architectural diagrams there is also a group of inhibitory neurons, this group is connected to all other neurons with 10\% probability and its synaptic connections are initialized to random values in the range [0,-3] mV.  This has the effect of slightly dampening the activity of the network.

All of the neuron groups, with the exception of the dopaminergic neurons, consist of 20 neurons.  The dopaminergic group consists of 40 neurons.  The reason that the dopaminergic neuron group is larger is that learning to elicit a dopamine response when a reward predicting stimulus (food-container touch-sensor) is received relies heavily on random firings in the dopaminergic neuron group.  A larger number of neurons was found to increase the learning speed. The conductance delay for all synapses was set to 1ms, meaning that it would take 1ms for a pre-synaptic spike to propagate into a change in membrane potential of a post-synaptic neuron.

The spiking neural network for the food-only robot consists of 160 neurons, the distribution of these neurons across the different neural groups is shown in table \ref{tab:foodonly_neurons}.  The network for the food-container robot consists of 240 neurons in total, the distribution of the neurons for this robot is shown in figure \ref{tab:foodcontainer_neurons}.
\begin{table}[htbp]
\centering
\begin{tabular}{|l|l|}
\hline
Neuron Range & Neuron Group \\ \hline
1-20 & Left Food Sensor \\ \hline
21-40 & Right Food Sensor \\ \hline
41-60 & Left Motor \\ \hline
61-80 & Right Motor \\ \hline
81-100 & Food Touch-Sensor \\ \hline
101-140 & Dopamine Neurons \\ \hline
141-160 & Inhibitory Neurons \\ \hline
\end{tabular}
\caption{The organization of the neurons for the robot in the food-only environment.}
\label{tab:foodonly_neurons}
\end{table}

\begin{table}[htbp]
\centering
\begin{tabular}{|l|l|}
\hline
Neuron Range & Neuron Group \\ \hline
1-20 & Left Food Sensor \\ \hline
21-40 & Right Food Sensor \\ \hline
41-60 & Left Container Sensor \\ \hline
61-80 & Right Container Sensor \\ \hline
81-100 & Left Motor \\ \hline
101-120 & Right Motor \\ \hline
121-140 & Food Touch-Sensor \\ \hline
141-160 & Food-Container Touch-Sensor \\ \hline
161-180 & Empty-Container Touch-Sensor \\ \hline
181-220 & Dopamine Neurons \\ \hline
221-240 & Inhibitory Neurons \\ \hline
\end{tabular}
\caption{The organization of the neurons for the robot in the food-container environment.}
\label{tab:foodcontainer_neurons}
\end{table}

\section{Plasticity}
\label{section:meth_plasticity}

Synaptic plasticity was implemented using dopamine modulated STDP, an outline of which is given in section \ref{section:dop_stdp}.  The value of the function $\mathit{STDP}(t)$ at time $t$ is calculated as:
 \begin{equation}
  \mathit{STDP}(t) = \left\{
  \begin{array}{l l}
  	A^+e^{-\Delta t/\tau^+} & \quad \textrm{if $\Delta t \geq 0$}\\
  	-A^-e^{\Delta t/\tau^-} & \quad \textrm{if $\Delta t < 0$}\\
  \end{array} \right.
\end{equation}
Where $\Delta t$ is the time difference between the pre and post synaptic neurons firing.  The nearest-neighbour update scheme was used, such that only the most recent pre/post synaptic spike was considered when calculating $\mathit{STDP(t)}$ for a specific synapse.  The optimal parameters for the model were found by running the robot in the food-only environment for 1000 seconds, 15 times.  This was repeated for a large range of parameters and the value that achieved the best score (as measures by number of food items collected) was used.  These parameters were:
\begin{itemize}
	\item $A^+ = 0.1$
	\item $A^- = 0.15$
	\item $\tau^+ = 0.02$s
	\item $\tau^- = 0.11$s
\end{itemize}

\begin{figure}[htbp]
	\begin{center}
	  	\includegraphics[width=0.6\textwidth]{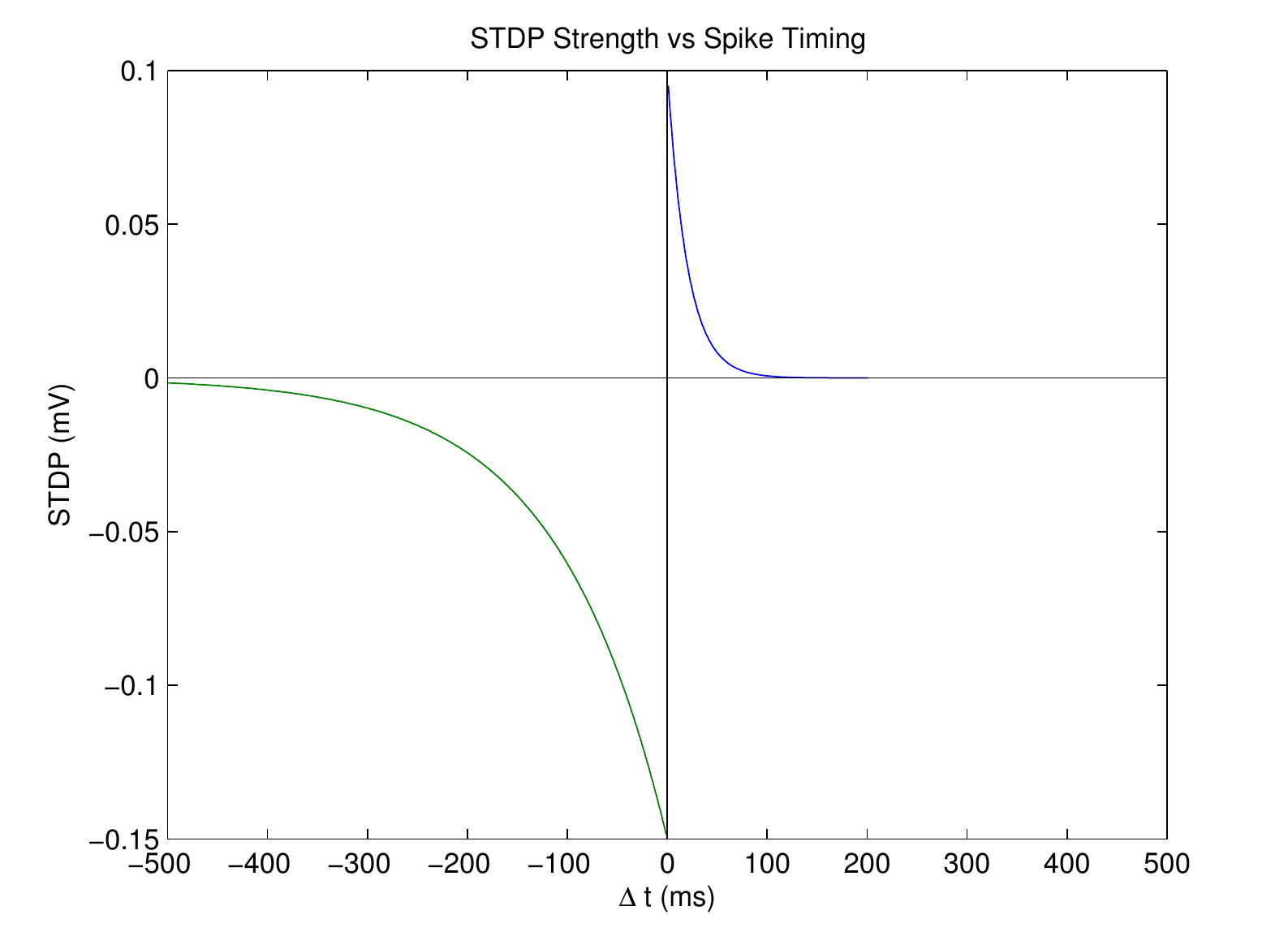}	
	\end{center}
	\caption{The Strength of STDP is shown against the difference in spike timings, where $\Delta t=t_{\mathit{post}}-t_{\mathit{pre}}$.}
	\label{fig:stdp_window}
\end{figure}

The strength of STDP is plotted against the difference in pre and post synaptic firing, $\Delta t$, in figure \ref{fig:stdp_window}.  There is a stronger response when the post-synaptic neuron fires before the pre-synaptic neuron ($\Delta t$ is negative) and as such long term depression is preferred over long term potentiation.  This means that uncorrelated firing between two populations of neurons will tend to decay the synaptic weights to zero and only correlated firings will have the effect of increasing the associated synaptic weights over time.  This is useful in helping the robot learn the correct behaviour and ignore any noise in the neuron firings.

\begin{figure}[htbp]
	\begin{center}
	  	\includegraphics[width=0.75\textwidth]{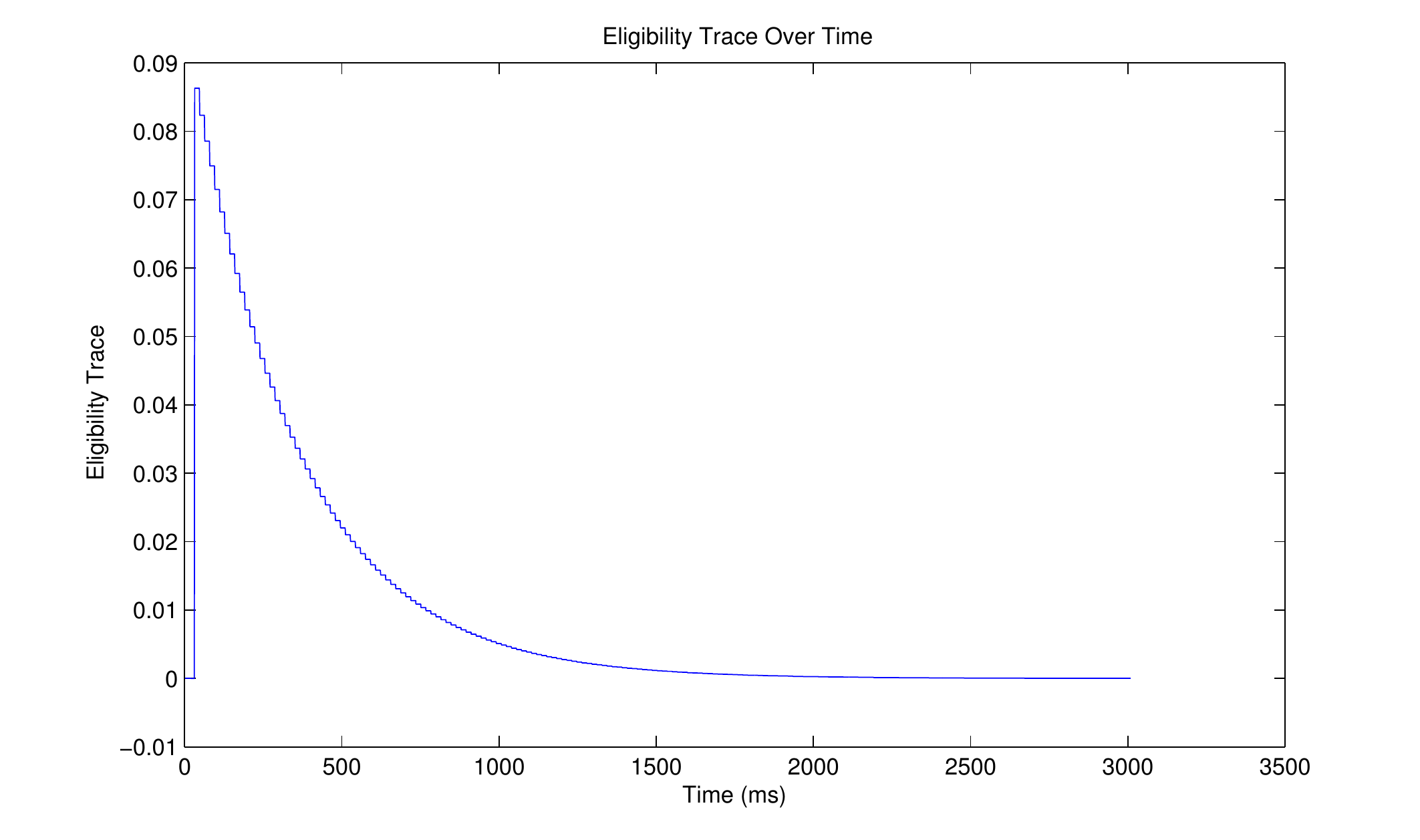}	
	\end{center}
	\caption{The eligibility trace for a synapse.  The pre-synaptic neuron fires after 30ms, the post-synaptic neuron fires after 32ms.  No more firing occurs and the eligibility trace decays.}
	\label{fig:example_eligibility_trace}
\end{figure}

Each synapse maintains an eligibility trace, $c$.  The value of which is updated according to:
\begin{equation}
\dot{c(t)}=-c(t)/\tau_c + \mathit{STDP}(t)\delta(t-t_\mathit{pre/post})
\end{equation}
Where $\tau_c$ is the decay rate of the eligibility trace and is set to 0.476s, $\delta$ is the Dirac delta function and ensures that the STDP only effects the eligibility trace when either the pre or post synaptic neuron fires.  Figure \ref{fig:example_eligibility_trace} demonstrates the eligibility trace of a single synapse after simultaneous pre-synaptic then post-synaptic firing.  The eligibility trace becomes negligible after a maximum of about one second after coincident firing.  This means that any action the robot is currently performing will be forgotten after this time.  See section \ref{section:moving_dopamine} for details of how this memory is extended further by moving the dopamine response.

Every millisecond the synaptic weights are then updated according to:
\begin{equation}
\dot{s}=cd
\end{equation}
Where $s$ is the synapse strength and $d$ is the current level of dopamine.  The level of dopamine is increased by 0.0035$\mu$M for every dopaminergic neuron that fires.  This has the effect of increasing the level of dopamine by approximately 0.12$\mu$M every time a food item is reached and the majority of the dopaminergic neurons fire.  During experiments which contain poison, the poison has the effect of reducing the level of dopamine by 0.0035$\mu$M for every dopaminergic neuron that fires.  The consequence of this is that synapses that were highly active in the lead up to the poison, and as such have a high eligibility trace, will have their synaptic weight reduced.  A delay of 5ms between a dopaminergic neuron firing and the corresponding increase in dopamine was used.  This gives enough time for the eligibility trace of synapses projecting to dopaminergic neurons to be updated, before the corresponding spike in dopamine.

The baseline level of dopamine in the system was set to -0.0004$\mu$M, meaning that in the absence of any firing by the dopaminergic neurons the level of dopamine would decay to this negative value.  This has the effect of working like hunger, if the robot does not collect any food for a long time then strong synaptic weights will become weaker and very weak synaptic weights will become stronger.  This allows the robot to change its behaviour when its current strategy isn't being effective in collecting food.

A constant current of 3.65mV was supplied to the dopaminergic neurons, this induced a background firing rate of 0.4Hz.  For values higher than this the robot was not able to learn correctly, this is talked about in more depth in section \ref{section:moving_dopamine}.

As described in section \ref{section:dop_stdp} dopaminergic neurons have two patterns of activity, background firing and stimulus induced burst firing.  The result of which is that that background firing does not have any significant effect on the level of dopamine.  To simulate this the level of dopamine was only increased if more than five dopaminergic neurons fired in a 1ms window.  The effect this has is discussed more in section \ref{section:moving_dopamine}.

\section{Moving Dopamine Response from Unconditioned-Stimulus to Conditioned Stimulus}
\label{section:moving_dopamine}

In many scenarios an agent will need to perform a sequence of behaviours in order to get a reward.  If dopamine is only released when the reward is achieved then plasticity will only occur at this time.  The efficacy of DA-modulated STDP drops off exponentially and is negligible after about 1000ms, therefore if the reward predicting behaviour takes place more than 1000ms before the reward is received then this behaviour will not be learnt.  For example, in the food-container environment, with the robot run at half-speed, the average time between entering a food-container and the food being collected is about 800ms.  Therefore food-container attraction behaviour would take a long time to learn, if at all, if dopamine was only released when the food was collected.  If the robot needed to learn a third behaviour, such as picking up a key to open the container, then it is likely that the temporal difference between this behaviour and the actual reward would be too great for it to ever be learnt.

A simple method to deal with this problem is for the agent to learn to elicit a dopamine response when it receives a reward predicting stimulus. For example, the food-container touch-sensor is often stimulated prior to a food item being collected and hence a spike in dopamine being received.  If the robot learnt to produce a dopamine spike when the food-container touch-sensor is stimulated then the robot should be able to learn food-container attraction behaviour much faster, as there wouldn't be a long delay between performing the behaviour and receiving a dopamine response.  It can be seen how this process could be repeated such that a long chain of behaviours could be learnt.

\begin{figure}[htbp]
	\begin{center}
	  	\includegraphics[width=0.8\textwidth]{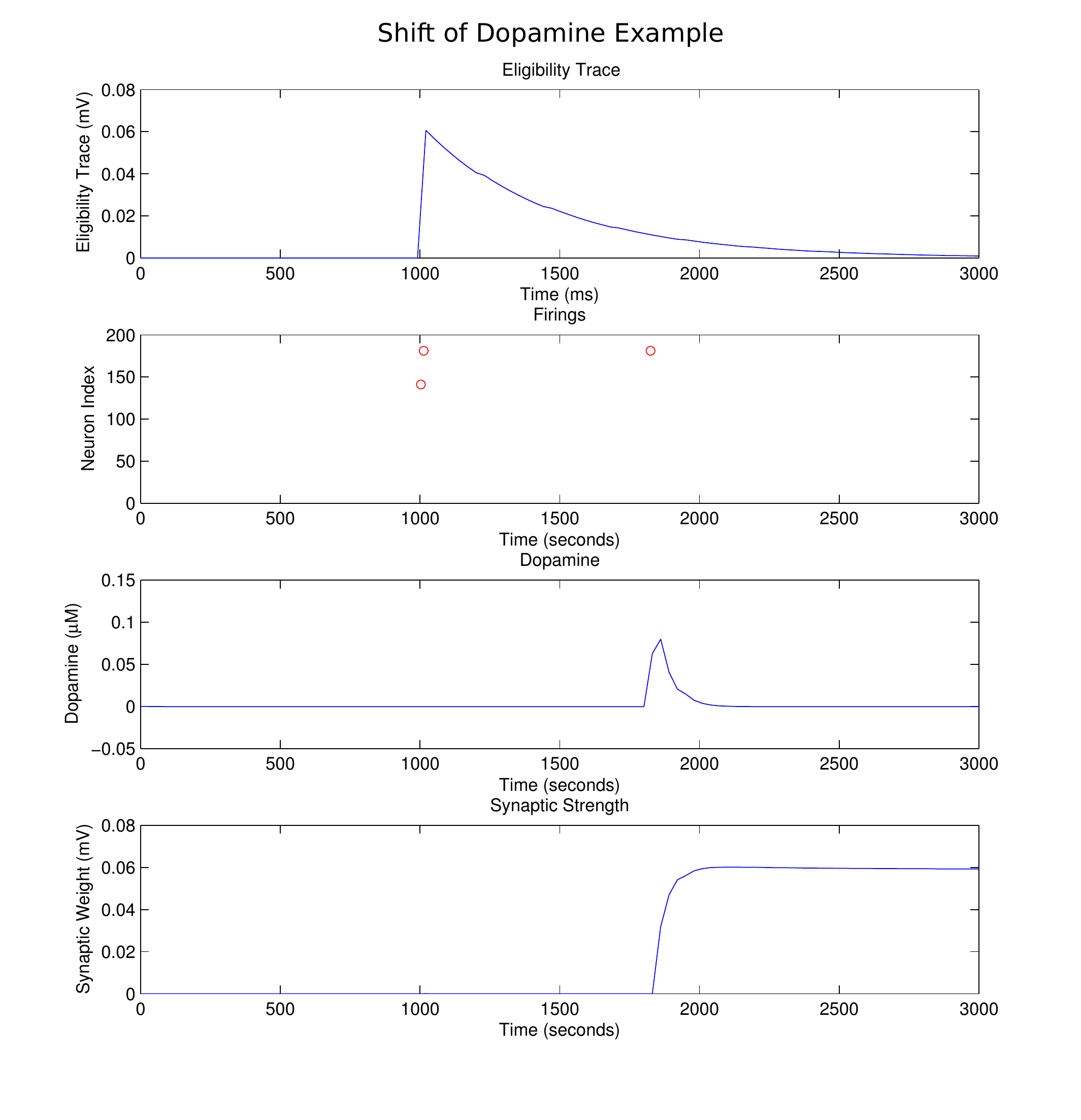}	
	\end{center}
	\caption{The activity of two neurons is plotted, one (index 141) from the food-container sensor group, and one (index 181) from the dopaminergic neuron group.  At 1000ms the robot enters the food container and the food-container touch-sensor neuron spikes, by chance the dopaminergic neuron fires soon after which raises the eligibility trace.  At 1800ms the robot collects the food item and the level of dopamine spikes. This dopamine spike, combined with the positive eligibility trace, increase the weight of the synapse from the food-container touch-sensor neuron and the dopaminergic neuron.}
	\label{fig:dop_move_example}
\end{figure}

This moving of dopamine response is achieved by connecting the container touch-sensor neurons to the dopaminergic neurons and allowing these synapses to be plastic.  Figure \ref{fig:dop_move_example} shows an example of how the robot is able to strengthen food-container touch-sensor to dopaminergic neuron connections.  The firings of just two neurons are plotted, one from the food-container touch-sensor group (index 141) and one from the dopaminergic group (index 181).  At 1000ms the robot enters a food container and the food-container touch-sensor neuron fires.  Purely by chance the dopaminergic neuron fires 20ms later due to background firing.  This causes the eligibility trace between these two neurons to jump up.  800ms later the robot reaches the food item and all the dopaminergic neurons fire, causing a spike in the level of dopamine.  This spike in dopamine, combined with the still positive eligibility trace, increase the weight of the synapse.  Since the synaptic weight is higher this increases the probability that the dopaminergic neuron will fire after the food-container touch-sensor neuron in future, and have its weight further increased.

The ability of the robot to learn this dopamine response relies on a combination of the background firing rate of the dopaminergic neurons and on the strength of long term potentiation as compared with long term depression.  If the background firing is too low then the chance of a dopaminergic neuron firing soon after a food-container touch-sensor neuron will be small and as such the robot will take a long time to learn, if at all.  If the background firing rate is too high, and the level of long term depression is also high then random uncorrelated firings between the food-container sensor neurons and the dopaminergic neurons will, on average, cause the synapses' eligibility traces to become negative and cancel out any potentiation of the synapses.

If we consider what would happen if we did not threshold the number of neurons required for the level of dopamine to be raised (or equivalently if dopaminergic neurons in the brain produced bursting activity during background firing) then we can see that non-reward predicting stimulus can also have their synaptic weights, for synapses projecting to dopaminergic neurons, potentiated.  This is due to the feedback loop that can be created.  For example, if by chance an empty-container touch-sensor neuron fires before a dopaminergic neuron.  The eligibility trace will increase, and the robot will immediately receive an increase in dopamine (due to the dopaminergic neuron firing), this will increase the probability that the next time the empty-container sensor neuron is triggered, the dopaminergic neuron fires afterwards, resulting in its synaptic weight increasing even further.

If the parameters of the model were altered such that plasticity happened at a much slower rate then this feedback loop could be avoided, the small increase in synaptic weight in the previous example would not alter the probability of the neurons firing in a correlated manner much.  Therefore it would be likely that the dopaminergic neuron would at some point fire before the sensor neuron and cause the synaptic weight to decrease again.  However the advantage of the above method of thresholding the dopaminergic firing rate needed to increase the level of dopamine is that the robot is more robust in its learning, as well as being able to learn much faster.

\section{Network Stability}
\label{section:meth_stability}
STDP is an inherently unstable mechanism, once a synaptic weight has been increased then the probability of the pre-synaptic neuron firing before the post-synaptic neuron becomes increased and as such the probability of the synaptic weight potentiating further is increased.  This feedback mechanism can lead to synapses quickly becoming potentiated to their maximum value.  To combat this two mechanisms were used.  The synaptic weights of excitatory synapses was restricted to the range [0,4] mV.  

In addition, a local dampening mechanism was used based on the observation that, in the brain, a highly active group of neurons have their excitability decreased over time, which was discussed in section \ref{section:plasticity_and_stability}.  The synapses were divided into three distinct groups, the synapses belonging to these three groups being:
\begin{itemize}
\item Food Taxis Group - Food range-sensor neuron to motor neuron synapses.
\item Container Taxis Group - Container range-sensor neuron to motor neuron synapses.
\item Container Dopamine Group - Container touch-sensor neuron to dopaminergic neuron synapses.
\end{itemize}

\begin{figure}[htbp]
	\begin{center}
	  	\includegraphics[width=0.5\textwidth]{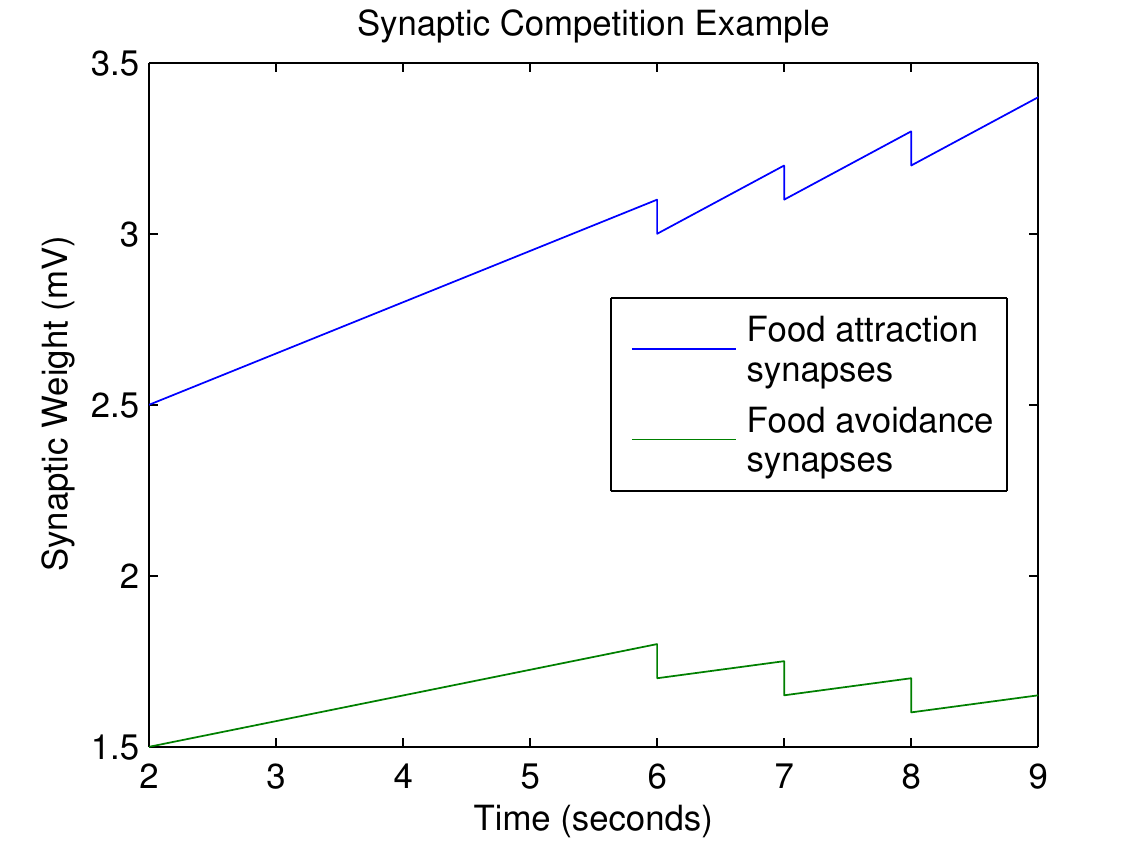}	
	\end{center}
	\caption{An example of how synaptic weight dampening can lead to competition between behaviours.  Both the synaptic weights for food attraction and for food avoidance behaviour are being potentiated, but the food attraction synaptic weights are potentiating faster.  At 6 seconds the average synaptic weight over both groups reaches the maximal allowable value, and all weights are reduced by 0.1mV.  The food attraction synaptic weights then increase at a faster rate, and as such the next time the maximal average synaptic weight is reached the food attraction behaviour has increased by more than 0.1mV and the food avoidance behaviour has increased by less than 0.1.  In this way the strength of the food avoidance behaviour becomes reduced over time, whilst the strength of the food attraction behaviour is increased.}
	\label{fig:synaptic_competition_example}
\end{figure}

For each of these groups, if the mean synaptic weight becomes potentiated by more than 2mV (half the maximum value) then all synapses within the group have their weights reduced by 0.1mV. This prevents a feedback loop such that all the synaptic weights become potentiated to their maximum value.  The other benefit of this approach is that it can lead to competition between behaviours within the synaptic group, figure \ref{fig:synaptic_competition_example} demonstrates how this competition can occur between the food attraction and food avoidance behaviours. 

\section{Taxis}
The robot's architecture is constructed such that it has the potential for taxis behaviour, for example food attraction or poison avoidance.  This is achieved through having two sensors at the front (left and right) of the robot, which are wired to two motors.  The wiring of the robot does not predispose it to prefer attraction or avoidance behaviour, there is an equal number of connections from each sensor to each motor.  The acquisition of attraction/avoidance behaviour is left up to DA-modulated STDP to strengthen the synaptic connections corresponding to the behaviour to be learnt.

\begin{figure}[htbp]
     \begin{center}
        \subfigure[Left Sensor Active, Right Motor Active]{
            \label{fig:robot_taxis1}
            \includegraphics[width=0.35\textwidth]{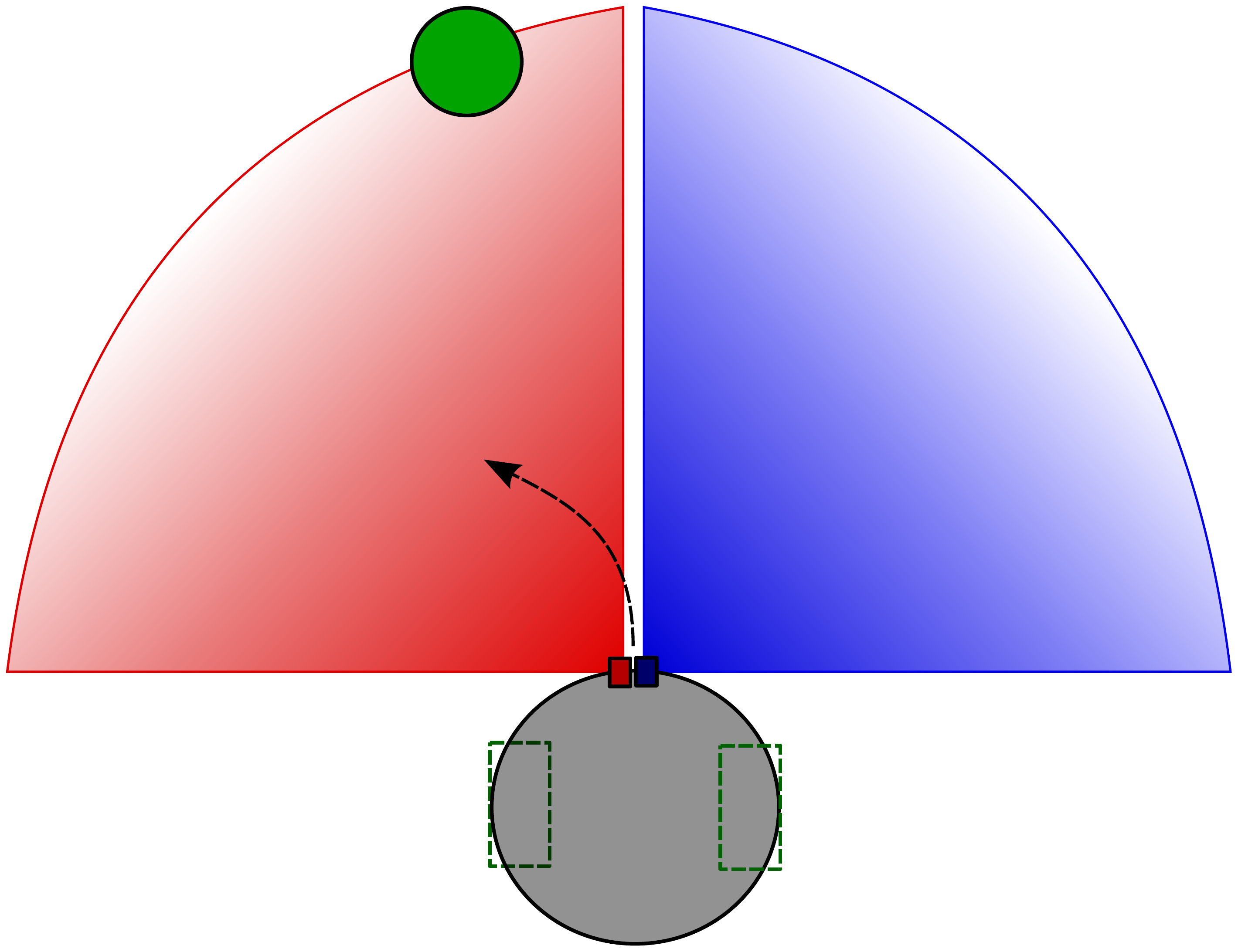}
        }
        \subfigure[Right Sensor Active, Left Motor Active]{
            \label{fig:robot_taxis2}
            \includegraphics[width=0.35\textwidth]{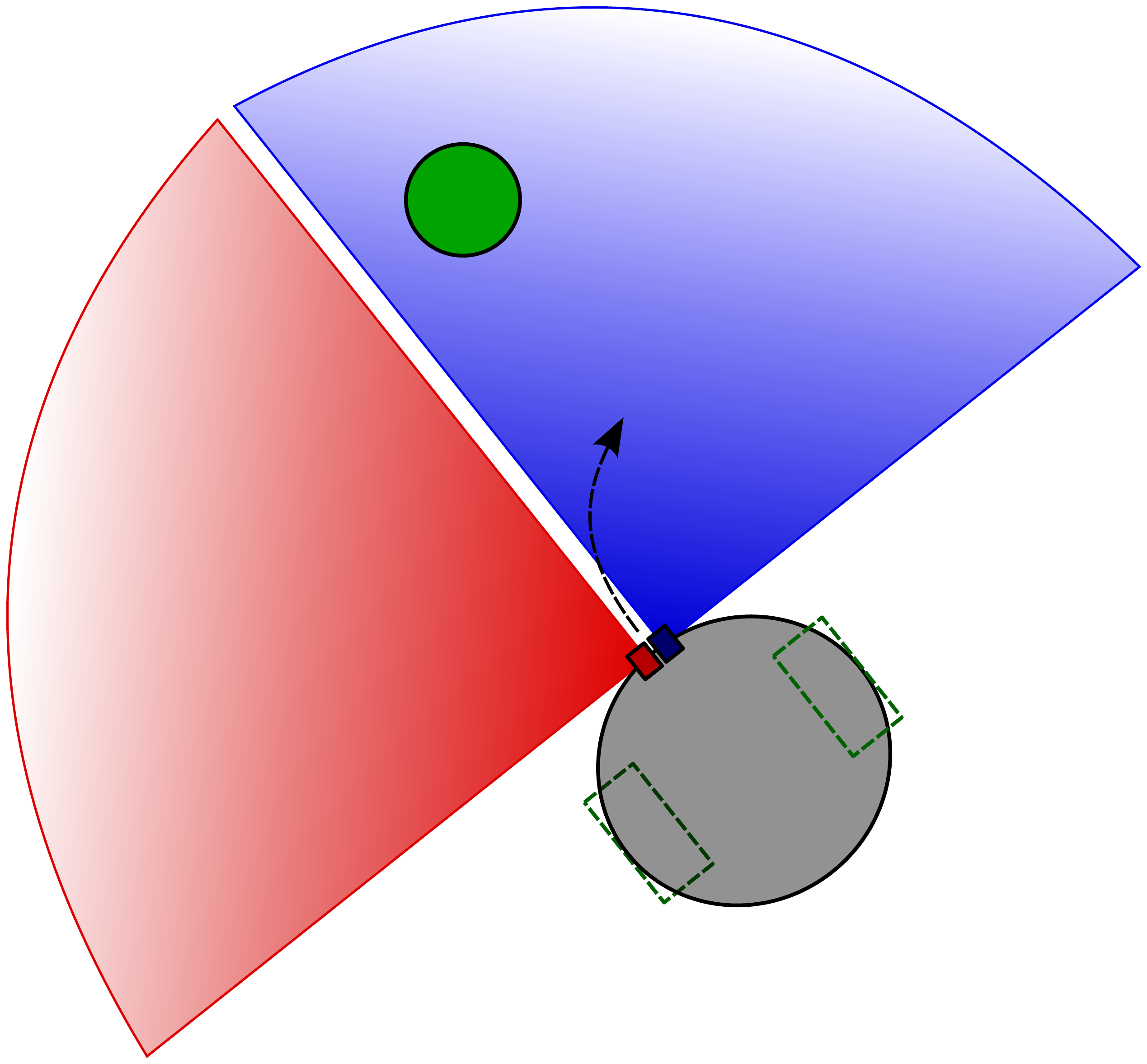}
        }
        \subfigure[Left Sensor Active, Right Motor Active]{
            \label{fig:robot_taxis3}
            \includegraphics[width=0.35\textwidth]{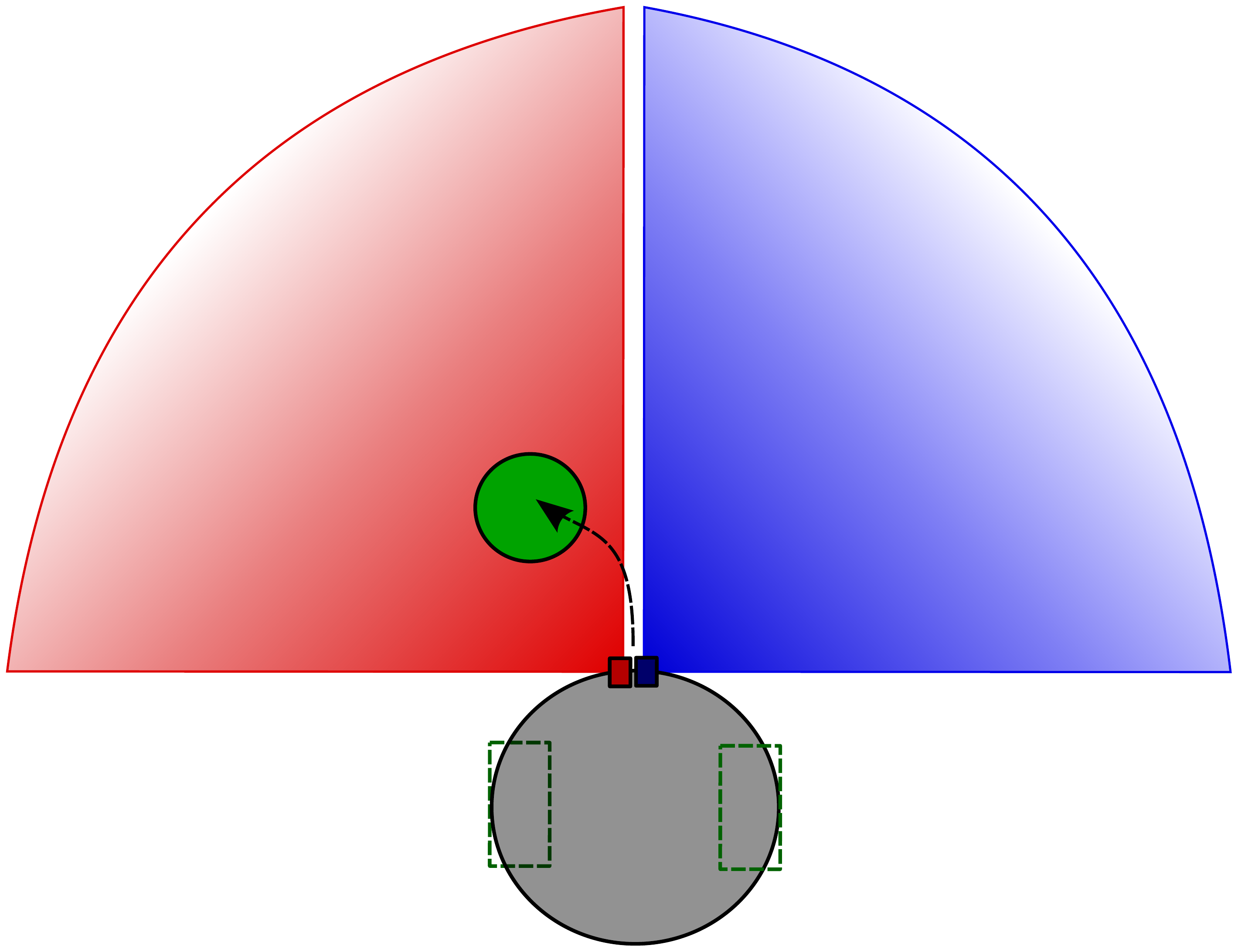}
        }
    \end{center}
    \caption{Demonstration of how wiring the left sensor to right motor, and right sensor to left motor can cause food attraction taxis.  In \subref{fig:robot_taxis1} the left sensor is active, so the robot turns forward and left, in \subref{fig:robot_taxis2} the right sensor is active so the robot turns forward and right.  Finally in \subref{fig:robot_taxis3} the robot turns and collects the food item.}
   \label{fig:robot_taxis}
\end{figure}

Figure \ref{fig:robot_taxis} shows how strong connections between opposite sensors and motors can cause food attraction taxis behaviour.

\section{Exploration}
\label{section:exploration}
To ensure that the robot explores the environment the robot needs to have some kind of exploratory behaviour.  This was implemented by, every 70ms, randomly choosing either the right or left motor.  Then for the remaining 70ms this motor's neurons were stimulated with a current taken from a Poisson distribution with a mean of 2.35mA.  

One possible way in which this exploration could be implemented directly in the network, would be to have the motor neurons connected such that they are oscillating at 14Hz (see section \ref{section:meth_phasic_activity} for details of how oscillating behaviour can occur).  If a winner-takes-all mechanism was then implemented between the left and right motors, this would have the effect that one population of neurons would be active whilst the motor neurons are phasically active.  When the motor neurons become phasically inactive then it would be easier for the winner-takes-all mechanism to switch the active group.  This network would display the same properties as the implemented exploration motor stimulation.

\begin{figure}[htbp]
	\begin{center}
	  	\includegraphics[width=0.6\textwidth]{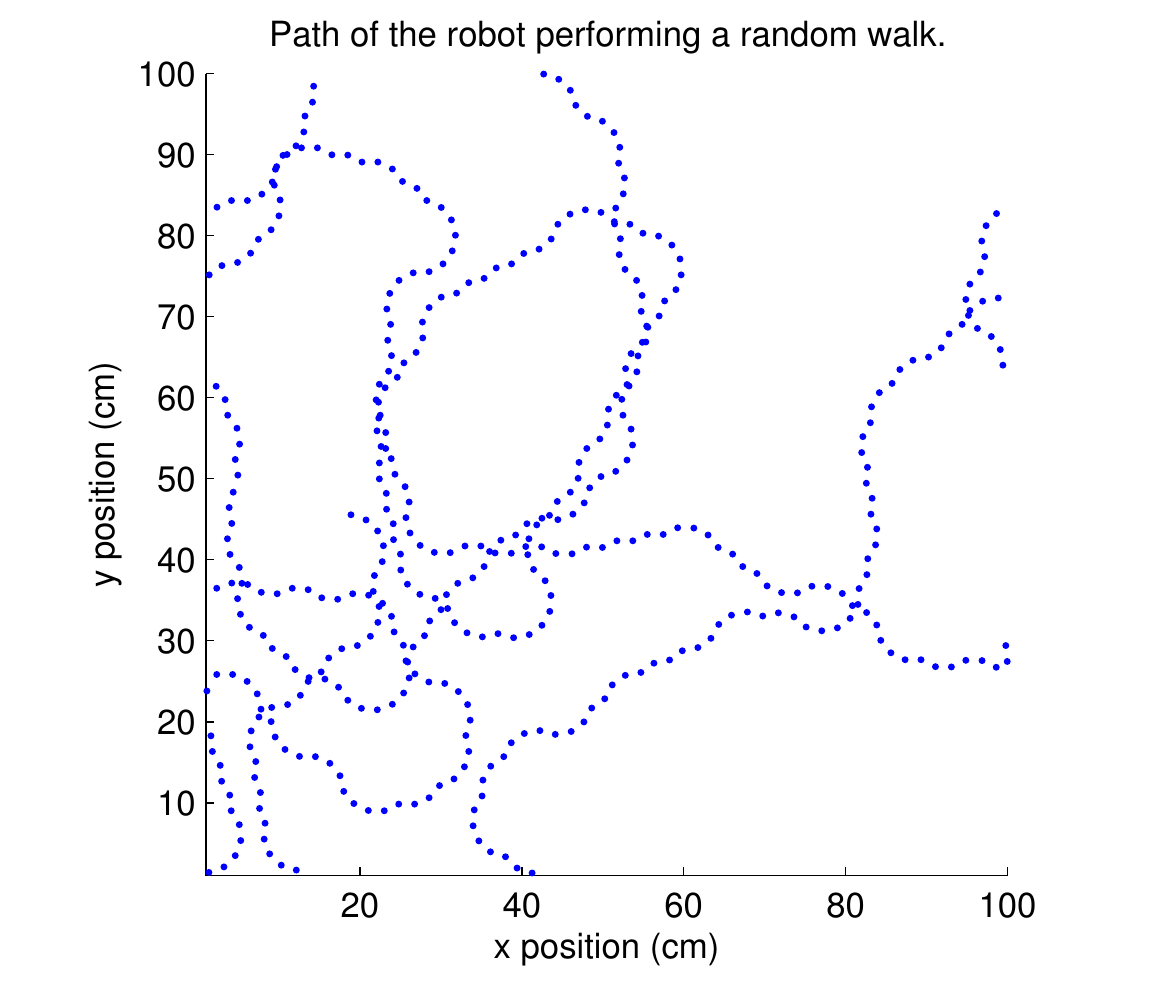}	
	\end{center}
	\caption{The path taken by the robot for 30 seconds in an environment with no food.  The random motor stimulation causes the environment to be explored effectively.}
	\label{fig:random_walk}
\end{figure}

Once the synaptic weights between the sensors and the motors have increased enough then the current to the motors from the sensors will override the exploration current.  The effect of this is that before the robot learns anything it will perform a random walk. After learning the robot will only perform a random walk whenever it can't sense anything.  Also, if the environment changes, causing the synaptic weights between sensors and motors to drop, then the robot will go back to performing a random walk. The path of the robot in an environment with no food, after having run for 30 seconds is shown in figure \ref{fig:random_walk}.  As can be seen the robot is able to effectively explore the environment.

\chapter{Results and Discussion}
\label{ch:results}

\section{Orbiting Behaviour}
\label{section:orbiting}
\begin{figure}[htbp]
	\begin{center}
	  	\includegraphics[width=0.6\textwidth]{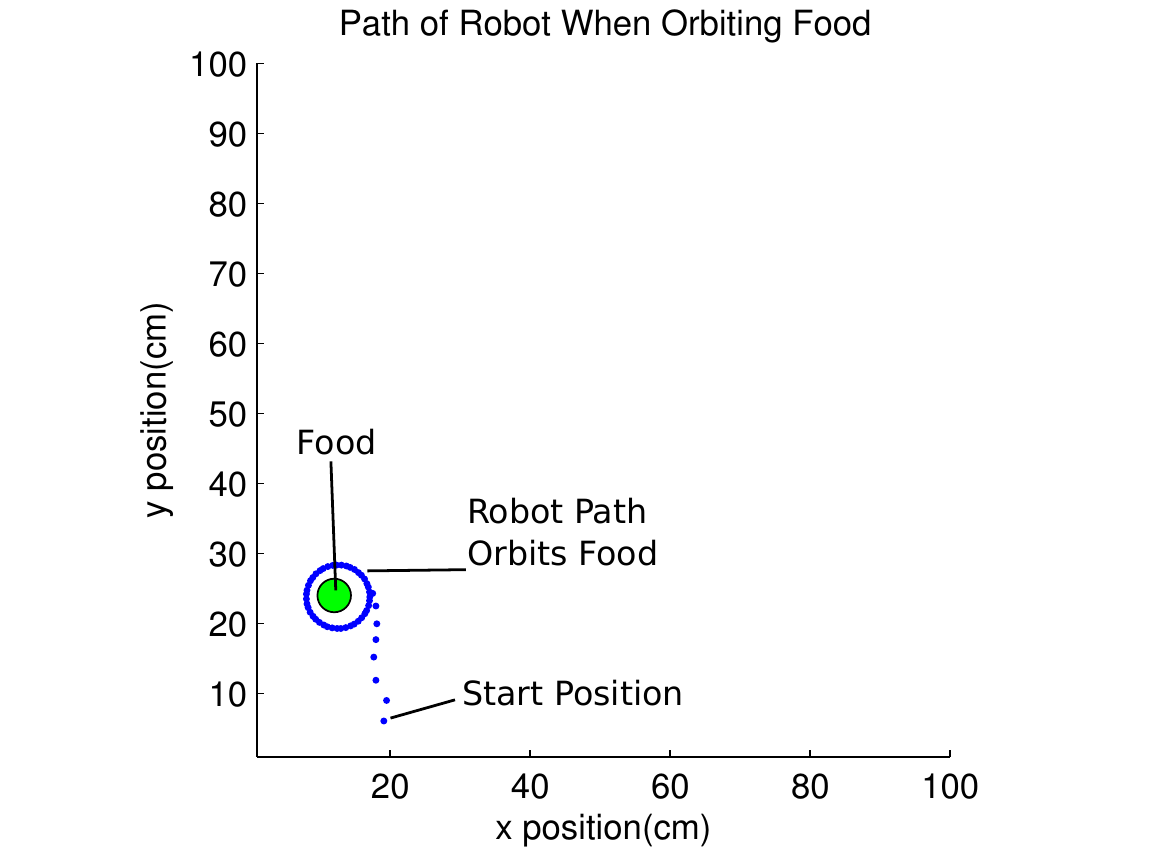}	
	\end{center}
	\caption{The path of the robot is shown in blue, with the robot's position being plotted every 70ms.  Once the robot can sense the food with its left sensor then it turns left, it continues to sense the food with its left sensor and so continues turning left.  However it cannot turn fast enough to reach the food and ends up orbiting it.}
	\label{fig:orbiting}
\end{figure}

An interesting property of the robot and environment is that there is no such thing as a fixed set of optimal synaptic weights. In almost all scenarios in the environment the best set of synaptic weights would be to have all left sensor to right motor and right sensor to left motor synapses set to their maximum weight and all other sensor to motor synapses set to zero.  This will cause the robot to turn the quickest towards food.  However, because the robot has a non-zero turning circle, in some situations the robot can orbit a food item indefinitely.  Figure \ref{fig:orbiting} shows the path of a robot using these synaptic weights when it approaches a food item at 90 degrees.  At each time step the robot senses food with its left sensor and so turns to the left, however it never reaches the food.  The optimal robot needs to determine when it is orbiting food, and modify its behaviour accordingly.

\subsection{Experimental Set-up}
\begin{figure}[htbp]
	\begin{center}
	  	\includegraphics[width=0.6\textwidth]{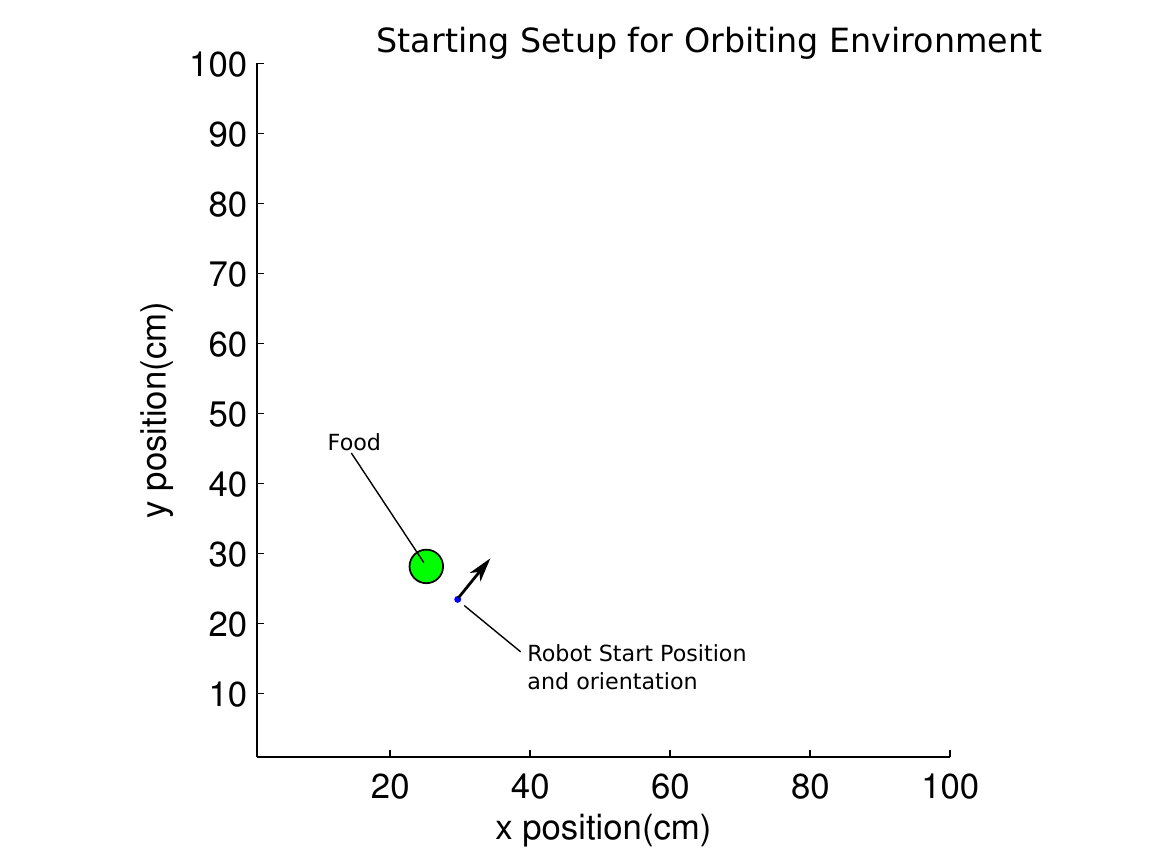}	
	\end{center}
	\caption{The starting position and set-up of the environment to force the robot to start orbiting.}
	\label{fig:orbiting_start_setup}
\end{figure}

To demonstrate the ability of the robot to learn to modify its behaviour when it gets stuck in one of these orbits the robot was placed in an environment with one food item.  The synaptic weights between opposite sensors/motors were set to their maximum allowable value, and all other sensor to motor synapses were set to zero. This forces the robot to initially perform food attraction behaviour.  The robot was positioned in such a way that it would start out orbiting the food item as shown in figure \ref{fig:orbiting_start_setup}.  

In the first scenario none of the robot's synapses are plastic, it does not learn. The robot was simulated for 40 seconds, this was repeated 20 times.  In the second scenario the robot's synapses were modified using DA-modulated STDP, as such the robot was allowed to learn.  The robot was again simulated for 40 seconds, for 20 trials.

\FloatBarrier
\subsection{Results and Discussion}
\begin{figure}[htbp]
	\begin{center}
	  	\includegraphics[width=0.99\textwidth]{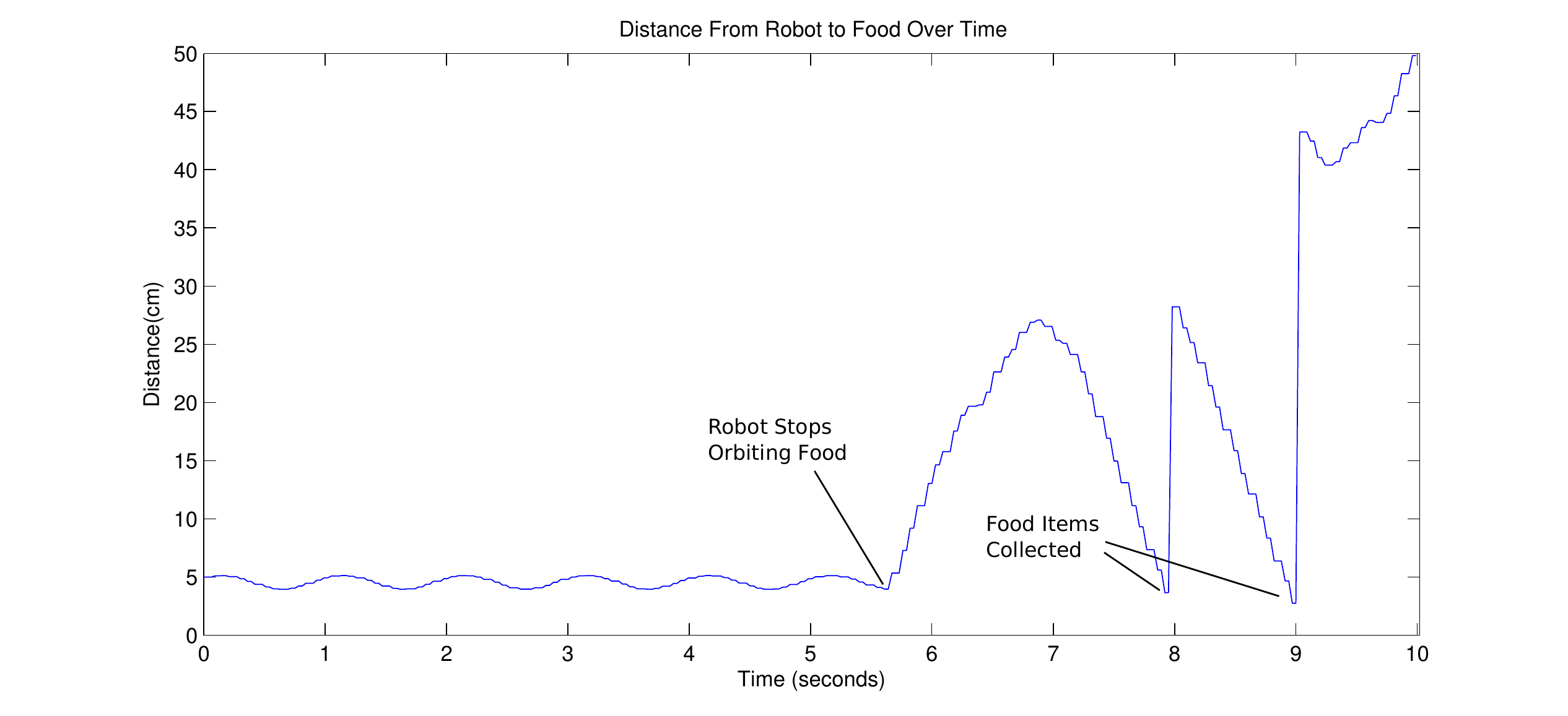}	
	\end{center}
	\caption{The distance between the robot and the food item is plotted over time, the synaptic weights are initially set to force food attraction behaviour and plasticity is turned on.  The robot starts out orbiting the food, but after 5.8 seconds is able to unlearn the orbiting behaviour and stops orbiting the food.}
	\label{fig:dist_orbit_escape}
\end{figure}

In none of the trials with learning disabled did the robot manage to break out of its orbiting behaviour.  With learning enabled the robot managed to learn to stop orbiting the food, and was able to collect the food item, in all 20 trials.  On average it took the robot 5.1 seconds to stop orbiting the food.  The distance to the food for the first 10 seconds of one such trial is plotted in figure \ref{fig:dist_orbit_escape}.  This figure shows the robot was able to modify its behaviour to stop orbiting after less than 6 seconds, the food item is then collected at 8 seconds.  Once the food is collected it is replaced at a random location, and the robot collects the food one more time.

\begin{figure}[htbp]
	\begin{center}
	  	\includegraphics[width=0.85\textwidth]{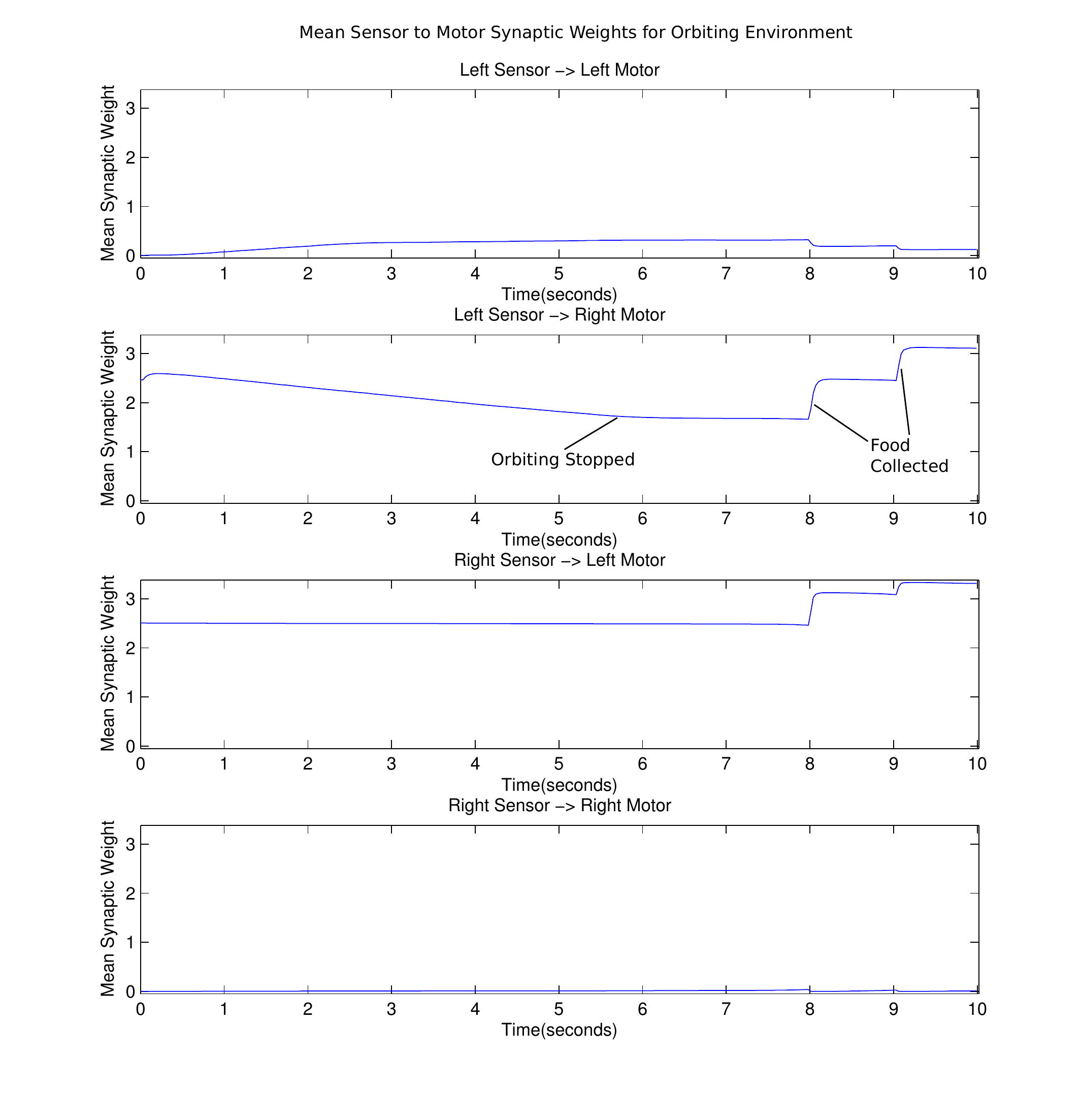}	
	\end{center}
	\caption{The mean synaptic weights between the left and right sensor/motor groups is shown for a robot in an environment with one food item.  The robot is initially orbiting the food, at 14 seconds this orbiting is unlearned.  At 35,37 and 39 seconds the robot collects a food item.}
	\label{fig:weights_orbit_escape}
\end{figure}

To see how this relearning is achieved the mean synaptic weights between both sensors and both motors over time are plotted in figure \ref{fig:weights_orbit_escape}.  There are several interesting properties that are shown.  Firstly the synaptic weights between the left sensor and the right motor are slowly decaying whilst the robot is orbiting the food.  This is because the baseline level of dopamine in the system is a small negative value, therefore highly active synapses with strong weights will have their synaptic weight reduced due to their high eligibility trace.  This has the effect of reducing the strength of any behaviour that the robot is currently performing, in this case turning left when the left food sensor is activated.  The negative dopamine can be thought of as analogous to hunger, if whatever the robot is currently doing isn't giving it food then it has an incentive to change its behaviour.

Notice that the right sensor to left motor synaptic weights remain strong.  This is due to the fact that the right sensor neurons are not being stimulated and as such the eligibility trace remains small. Whilst the robot is orbiting the synaptic weights between the left sensor and the left motor are also increased.  This is because the left motor neurons are still being randomly stimulated by the exploration method described in section \ref{section:exploration}.  This causes some random firings in the left motor population that are uncorrelated with the left sensor neuron firings.  The parameters of STDP used favour long term depression over long term potentiation and as such the eligibility trace for these synapses will be negative.  Given a negative eligibility trace, a negative dopamine value will increase the synaptic weights.

\begin{figure}[htbp]
	\begin{center}
	  	\includegraphics[width=0.99\textwidth]{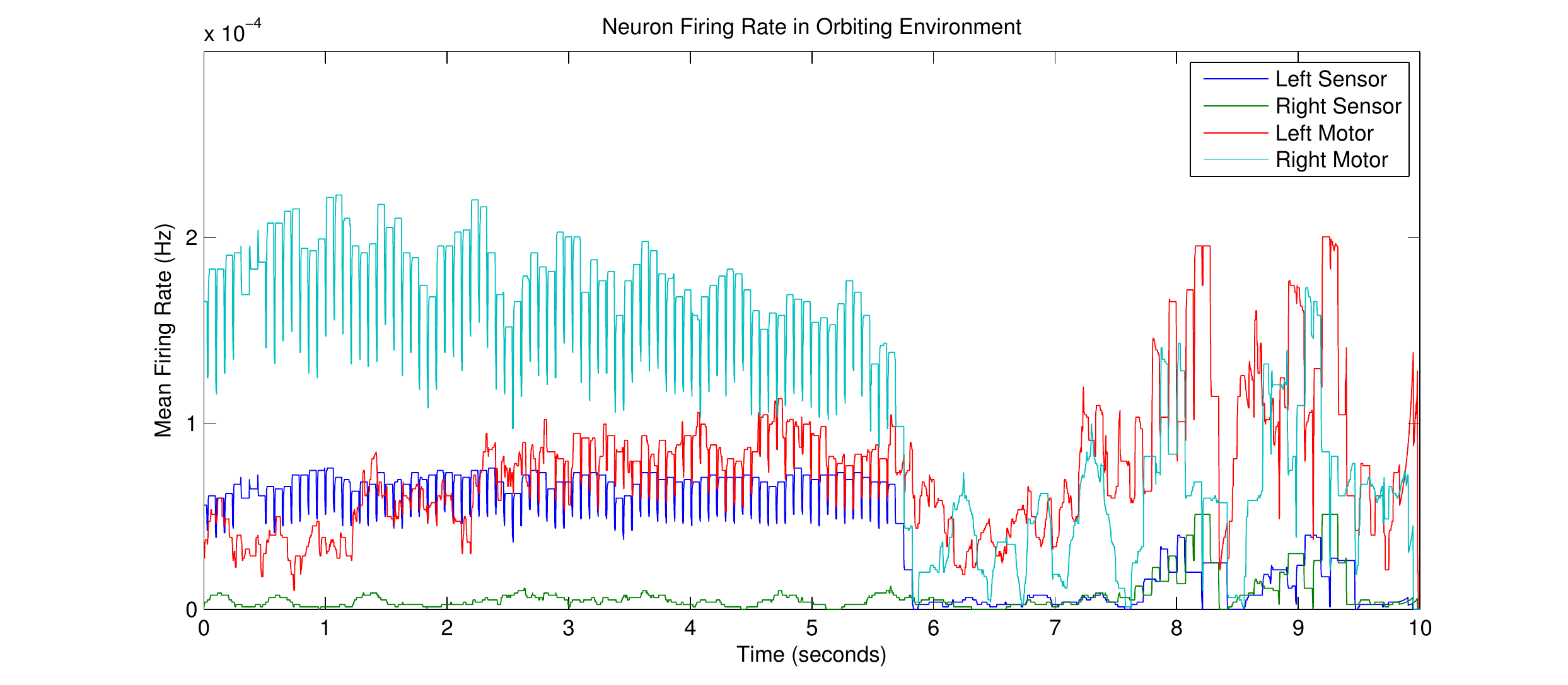}	
	\end{center}
	\caption{The mean firing rate of neurons in the sensor and motor groups over a single trial in the orbiting environment.  After 5.8 seconds the left motor firing has increased enough, and the right motor firing has decreased enough, that the robot stops orbiting the food.}
	\label{fig:firingrate_orbit_escape}
\end{figure}

Figure \ref{fig:firingrate_orbit_escape} shows the neural firing rate of both the sensor neuron groups and the motor neuron groups for the same trial.  This shows that as the synaptic weights between the left sensor and right motor are decreased, the firing rate of the right motor also decreases.  Equivalently, as the synaptic weights between the left sensor and the left motor are increased, so is the firing rate of the left motor neurons.  After 5.8 seconds the combination of increased synaptic weights to the left motor and the random exploration stimulus of the motors means that the firing rate in the left motor is greater than the firing rate in the right motor and the robot turns away from the food item.  The robot then performs a random walk until it can again sense the food.  Once the food is collected notice that the left sensor to right motor synaptic weights are increased (figure \ref{fig:weights_orbit_escape}), and the left sensor to left motor synaptic weights are decreased as the robot starts to relearn food attraction behaviour.

\begin{figure}[htbp]
	\begin{center}
	  	\includegraphics[width=0.85\textwidth]{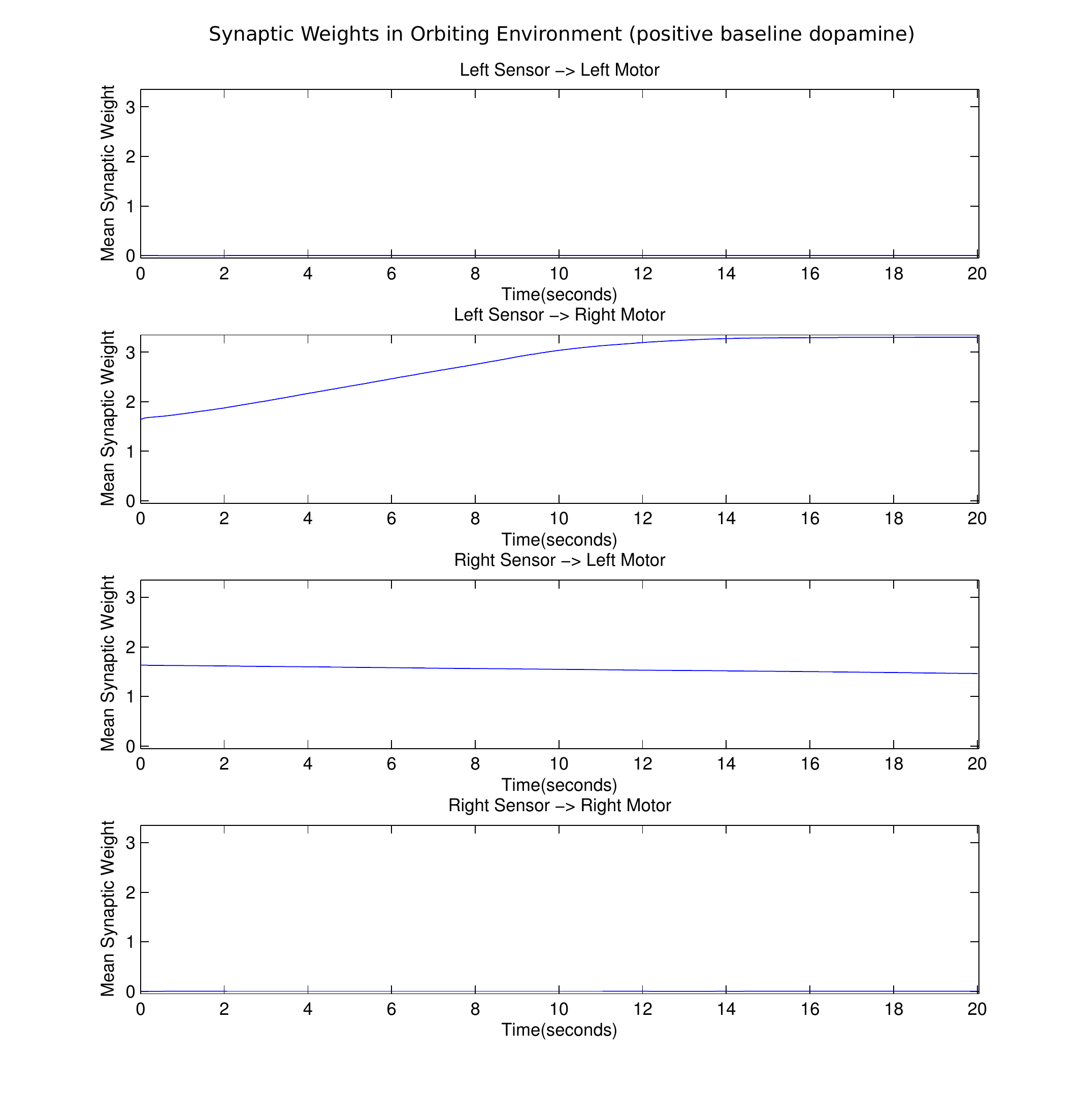}	
	\end{center}
	\caption{The synaptic weights between sensors and motors when the baseline level of dopamine is positive, and set to 0.0004$\mu$M in the orbiting environment.  This positive dopamine acts to reinforce the current behaviour, and the left sensor to right motor synaptic weights increase.}
	\label{fig:weights_orbit_pos_dop}
\end{figure}

\FloatBarrier
To confirm that the ability to relearn, when the current behaviour isn't working, relies on the baseline level of dopamine being negative, the experiment was re-run with the baseline level of dopamine set to 0.0004$\mu$M.  In none of these 20 trials did the robot manage to escape from orbiting the food.  Figure \ref{fig:weights_orbit_pos_dop} shows the synaptic weights for one of these trials.  As this figure shows, a positive level of background dopamine has the effect of reinforcing any behaviour that the robot is currently performing, as shown by the increase in weights between the left sensor and the right motor.

\section{Food Attraction Learning}
\label{section:food_attraction_learning}
In this section we test the ability of the robot to learn food-attraction behaviour by modifying it's synaptic weights.

\subsection{Experimental Set-up}
A 100cmx100cm environment consisting of 20 randomly placed food items was constructed.  When a food item is collected it is moved to a random position in the environment.  The robot was placed in the environment and simulated for 1000 seconds with plasticity taking place between the sensor neurons and the motor neurons.  This process was repeated for 50 trials, each time resetting the sensor to motor synaptic weights to zero.

For comparison a robot with no connections between sensor neurons and motor neurons and no plasticity was run for 50 trials of 1000 seconds each in the environment.  The effect of this is that the robot performs a random walk, and only collects food by randomly driving into it.

\subsection{Results and Discussion}
\label{section:food_attraction_learning_results}
\begin{figure}[htbp]
     \begin{center}
     	\includegraphics[width=0.8\textwidth]{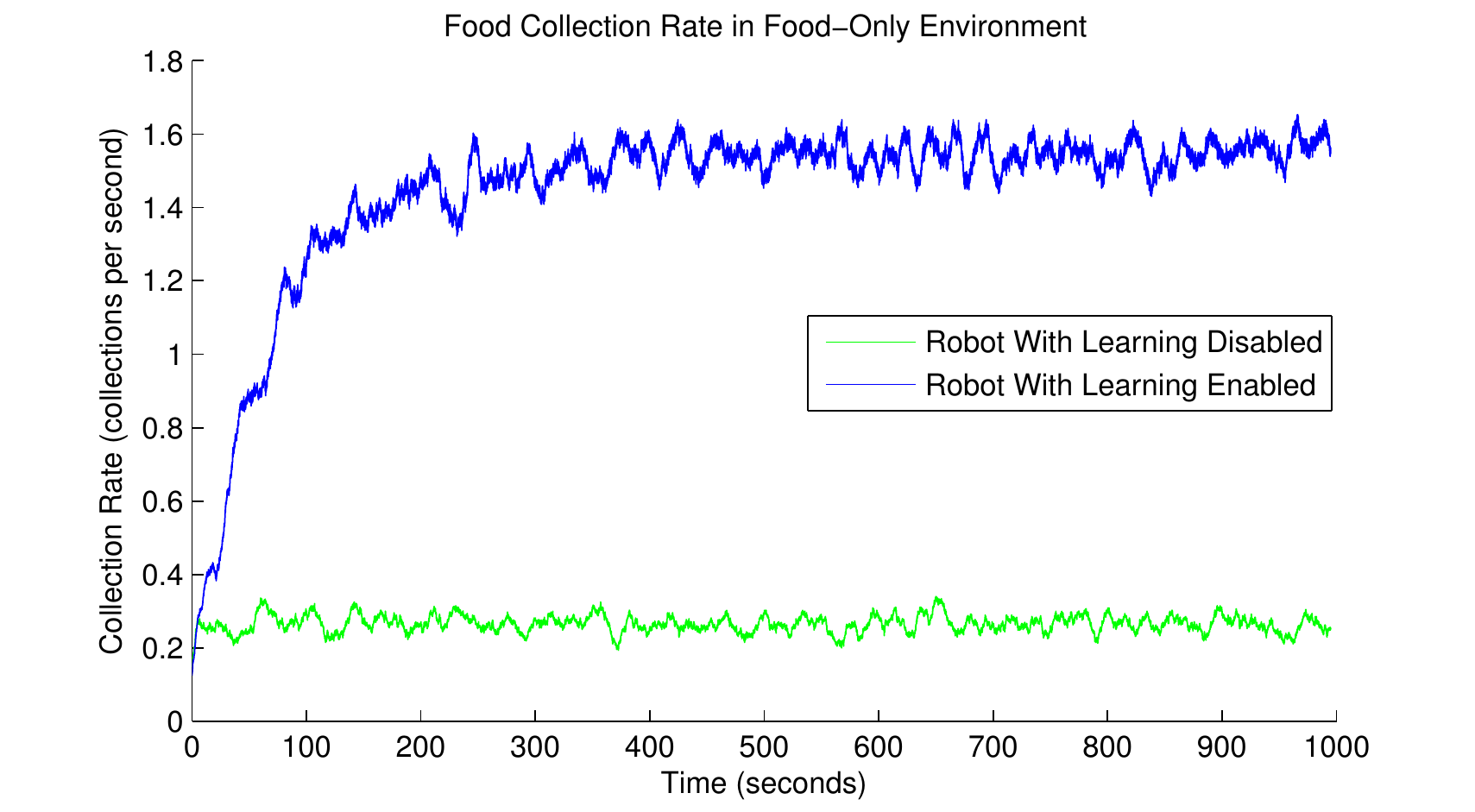}
    \end{center}
    \caption{The food collection rate over time for a robot in a 100cmx100cm environment with 20 randomly placed food items.}
   \label{fig:foodonly_scores}
\end{figure}

The mean food collection rate over time for the robot with learning enabled and the random walk robot is shown in figure \ref{fig:foodonly_scores}.  The robot with learning enabled was able to rapidly learn to turn towards and collect food items, and this high collection rate remained stable for the rest of the trial.  On average the robot took about 200 seconds for the full food attraction behaviour to be learnt and stabilized.

Table \ref{tab:foodonly_scores} shows the mean food collected over the 50 trials and whether the robot was able to correctly learn food attraction behaviour.  The robot was deemed to have learnt attraction behaviour if at the end of the trial the synapses for food attraction (left sensor to right motor, and right sensor to left motor) were on average more than 0.5mV, as well as being on average more than 10\% higher than the synapses for food avoidance.  These values being high enough that exploration behaviour was overridden, and different enough that food attraction was preferred.  The robot was able to learn this behaviour in all of the trials with learning enabled.
\begin{table}[htbp]
\centering
\begin{tabular}{|l|l|l|l|}
\hline
Robot Type & Mean Food Collected & Std. Deviation Food Collected & \% Correct \\ \hline
Learning Disabled & 269 & 16.1 & 0 \\ \hline
Learning Enabled & 1418 & 77.1 & 100 \\ \hline
\end{tabular}
\caption{This table shows the total amount of food collected for the robots, averaged over the 50 trials as well as what percentage of robots that were able to correctly learn food attraction behaviour.}
\label{tab:foodonly_scores}
\end{table}

\begin{figure}[htbp]
	\begin{center}
	  	\includegraphics[width=0.85\textwidth]{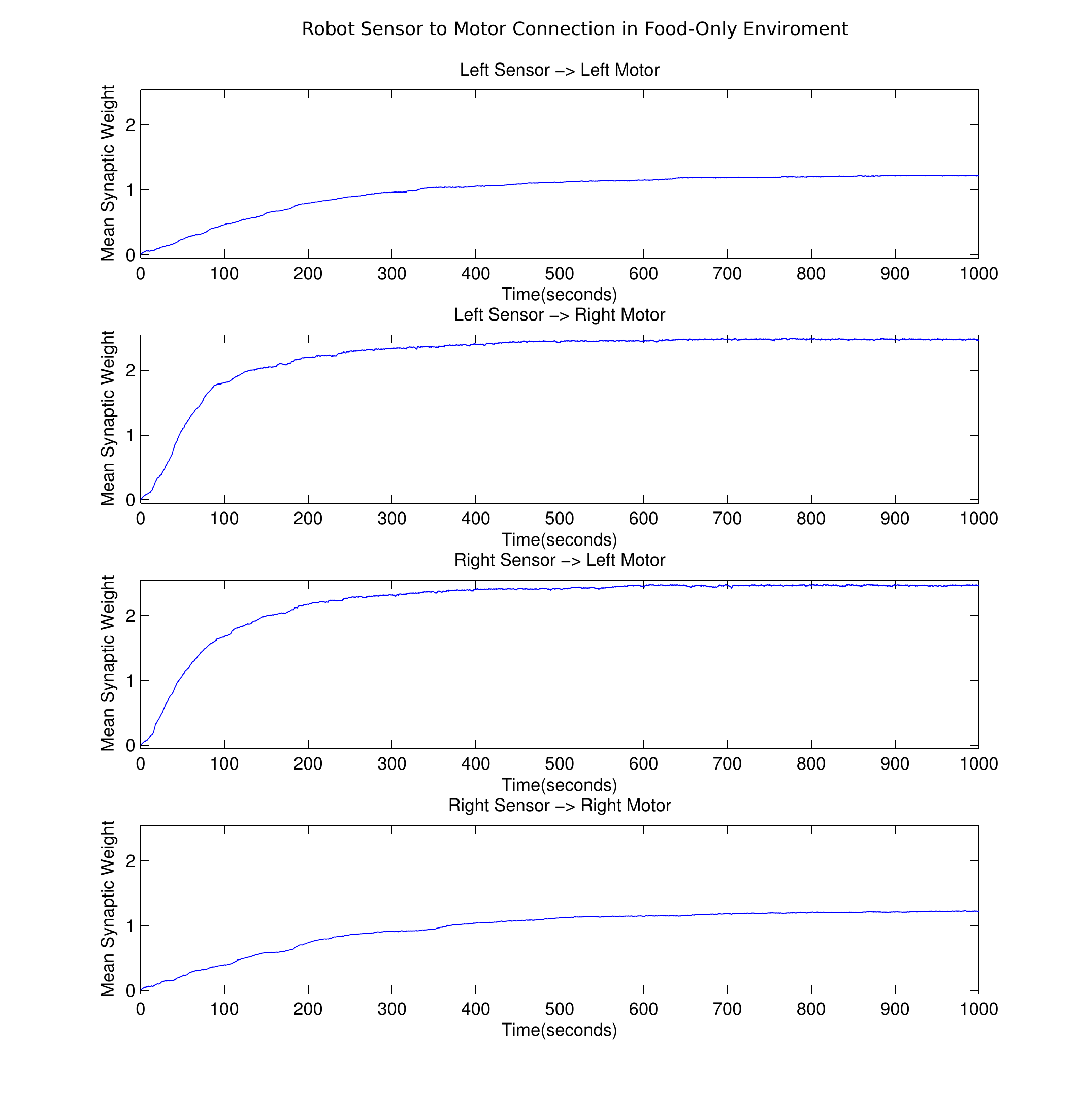}	
	\end{center}
	\caption{These graphs show the mean synaptic weights between the sensor and motor neurons averaged over 50 trials in the food-only environment. The left to right, and right to left connection weights are rapidly increased to their maximum value, causing food-attraction behaviour.}
	\label{fig:foodonly_weights}
\end{figure}

Figure \ref{fig:foodonly_weights} shows the mean weights between the sensors and motors, averaged over the 50 trials.  The synaptic weights between opposite sensors/motors are rapidly increased and reach the maximum weight after about 100 seconds on average.  This causes the robot to perform food attraction behaviour, as reflected by the increase in food collection in figure \ref{fig:foodonly_scores}.

\begin{figure}[htbp]
	\begin{center}
	  	\includegraphics[width=0.99\textwidth]{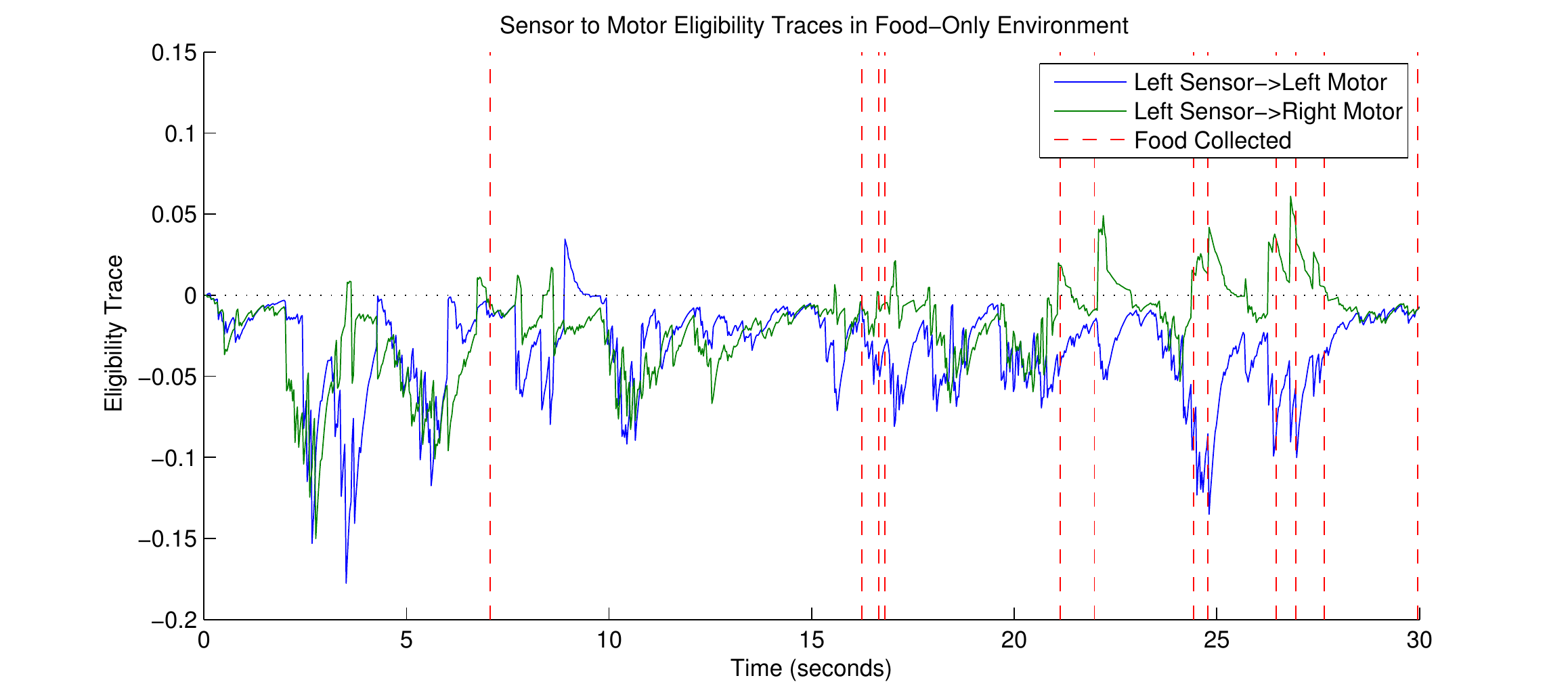}	
	\end{center}
	\caption{The mean synaptic eligibility trace for the left sensor to left motor neurons, and the left sensor to right sensor neurons.  The vertical lines display when food is collected.  Initially the weights are zero and the robot performs a random walk. The probability that the eligibility trace will be higher for the left sensor to left motor connections when food is achieved acts to increase these connections.  This acts as a feedback loop as the robot is more likely to turn towards the food and so the left sensor to left motor eligibility trace is even more likely to be positive.}
	\label{fig:foodonly_elig1}
\end{figure}

To demonstrate how the robot is able to learn to perform food attraction behaviour, the mean eligibility trace for synapses between the left sensor and both motors, for the first 30 seconds in one of the trials, has been plotted in figure \ref{fig:foodonly_elig1}.  At the start the synaptic weights are set to zero and the robot performs a random walk, notice that the eligibility traces fluctuate up and down, but neither is consistently higher.  When the robot collects a food item it is more likely that it was recently turning towards the food item, and as such the left sensor and right motor neurons will have been active and correlated (left sensor stimulated, followed by right motor neurons being active).  This means that the eligibility trace is likely to be higher for the left sensor to right motor when a food item is collected.  This can start to be seen at 16 seconds, but is most noticeable after 21 seconds.  The synaptic weights are only significantly modified when the dopamine level is high (when food is collected), so the left sensor to right motor connections will become strengthened.  This in turn means that the robot is more likely to turn towards food items and as such the left sensor to right motor connection strength is even more likely to be increased.  This action reinforcement is why the food attraction synaptic connections quickly become saturated to their maximum value.

\begin{figure}[htbp]
     \begin{center}
        \subfigure[Eligibility Trace for Single Trial]{
            \label{fig:foodonly_elig_stable}
            \includegraphics[width=0.85\textwidth]{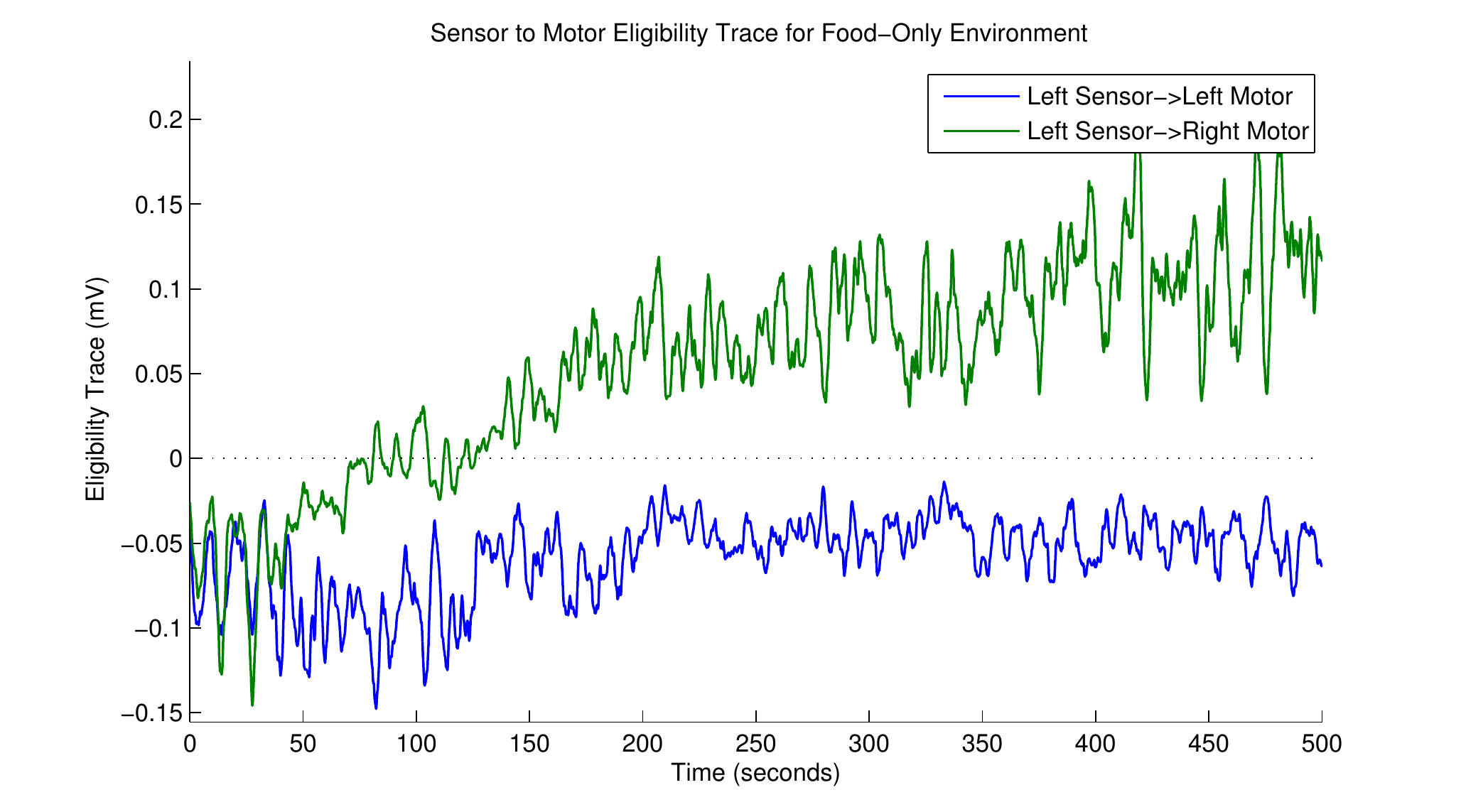}
        }
        \subfigure[Synaptic Weights for Single Trial]{
           \label{fig:foodonly_weights_stable}
           \includegraphics[width=0.85\textwidth]{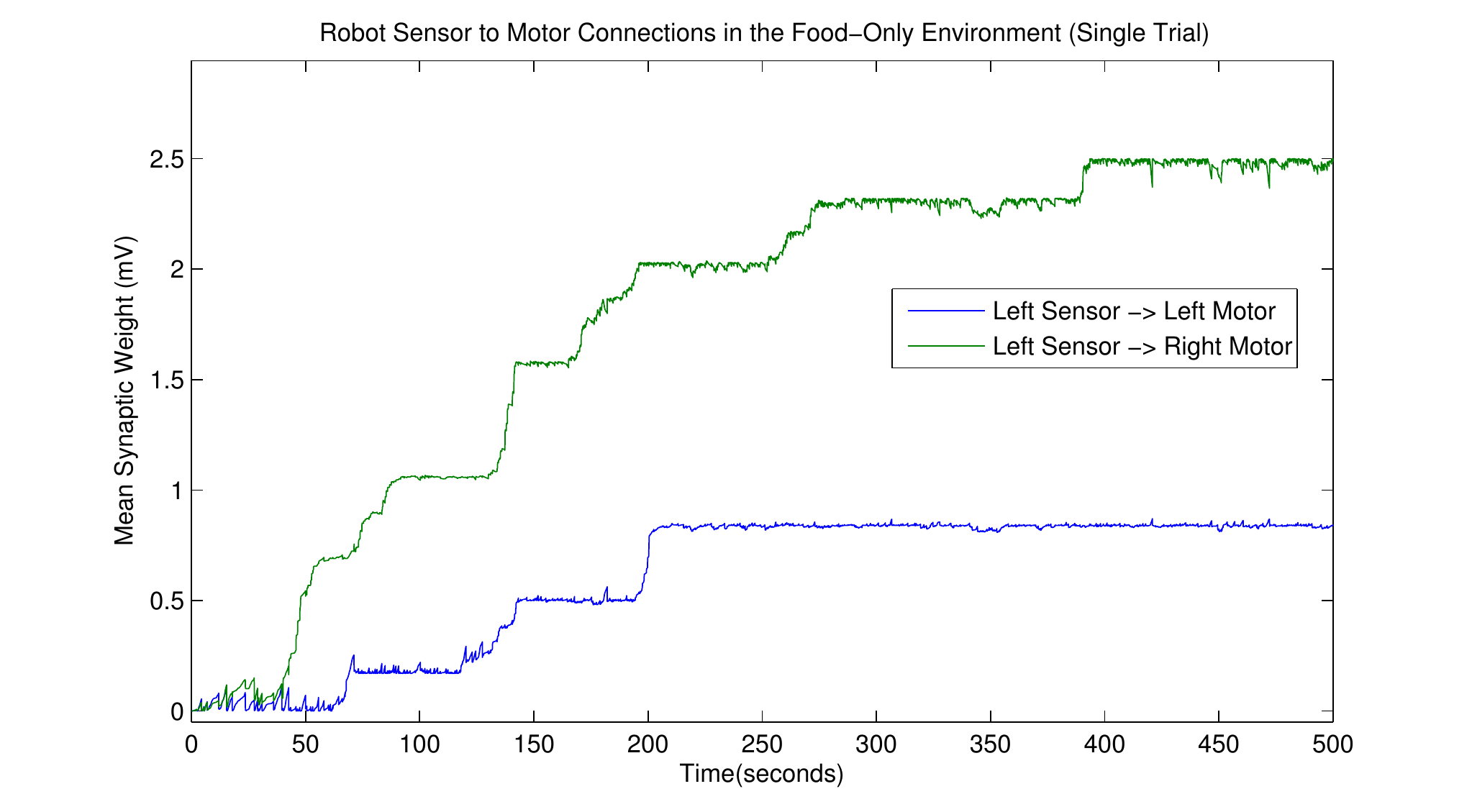}
        }
    \end{center}
    \caption{These graphs show the eligibility trace and synaptic weights between the left sensor neurons and both groups of motor neurons for the first 500 seconds in a single trial in the food-only environment.  Initially the left sensor to left motor eligibility traces are highly negative, this drives up their synaptic weights.  After 200 milliseconds the left sensor to left motor eligibility traces have risen enough that their synaptic weights are stable and do not increase.}
   \label{fig:foodonly_stable_example}
\end{figure}

\FloatBarrier
Notice that, from figure \ref{fig:foodonly_weights}, the left sensor to left motor and right sensor to right motor synaptic weights also increase, but they level off at a lower value.  The reason for this is the negative baseline level of dopamine.  Figure \ref{fig:foodonly_stable_example} shows the eligibility trace for the left sensor to motor synapses for a single trial, as well as the synaptic weights for these synapses.  Initially the synaptic weights between the left sensor and the left motor are small, as such the firing of the left sensor neurons and the left motor neurons are uncorrelated, causing a negative eligibility trace.  The negative baseline level of dopamine acts to increase the synaptic weights of synapses with negative eligibility trace.  As the synaptic weights between the left sensor and left motor increase, the neurons start to fire in a more correlated manner (stimulating the left motor causes some firings in the left motor).  This in turn causes the eligibility trace to rise.  Eventually, at around 200 seconds, the eligibility trace rises to a point where the positive dopamine spikes caused by food act to cancel out the negative baseline level of dopamine.  At this point the left sensor to left motor synaptic weights stabilize.  

The left sensor to right motor synapses are also subject to this process.  However, the main cause of potentiation in these synapses comes from the large spikes in dopamine when food is collected.  These large spikes in dopamine, which occur when the left sensor to right motor eligibility traces are highest, override any effects the small negative baseline dopamine has.

\begin{figure}[htbp]
	\begin{center}
	  	\includegraphics[width=0.85\textwidth]{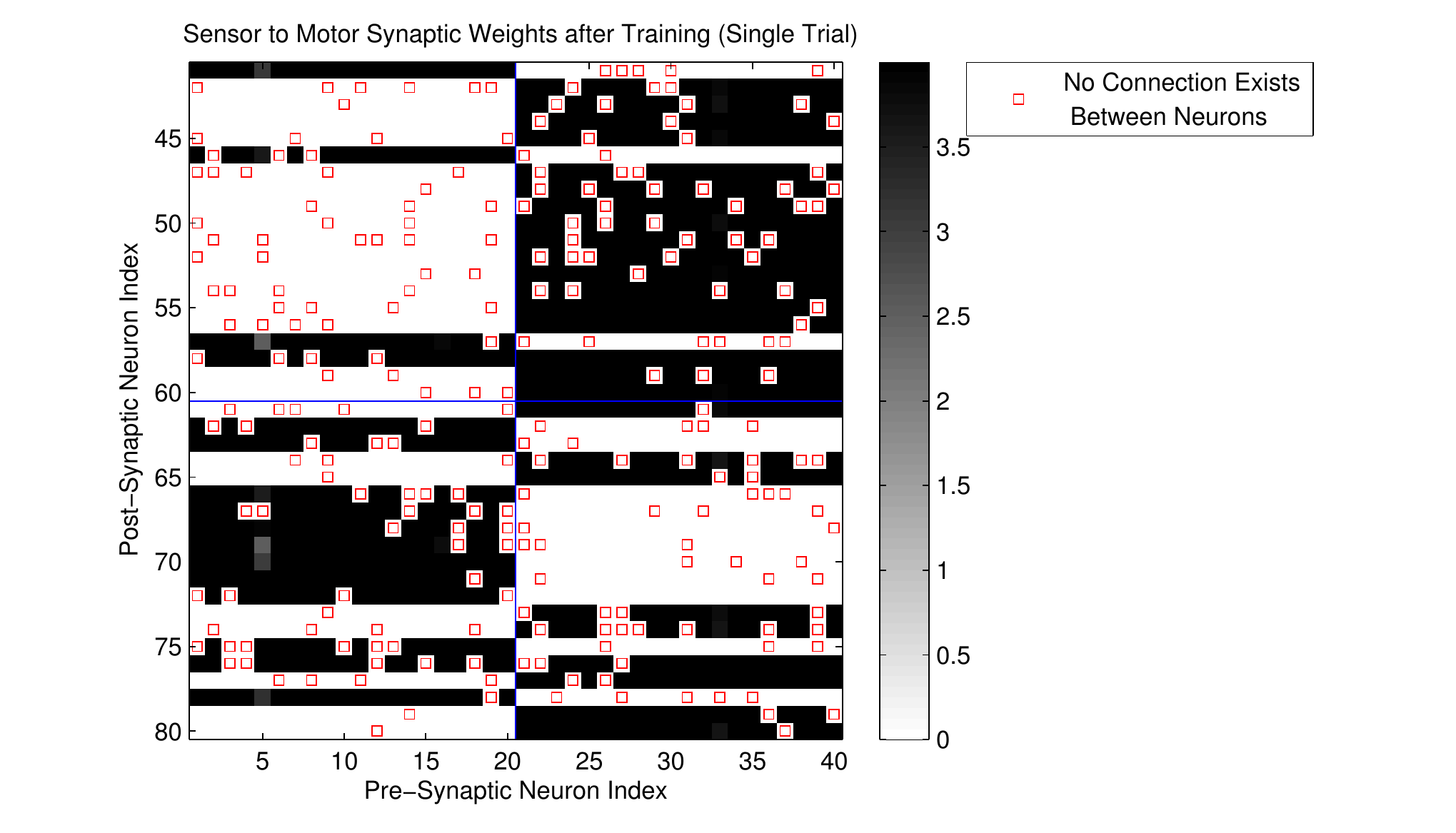}	
	\end{center}
	\caption{The synaptic weights between sensors and motors, after training in the food-only environment, is plotted.  Motor neuron indices are on the left axis and range from 41-60 (left motor) and 61-80 (right motor).  Sensor neuron indices are on the bottom axis and range from 1-21 (left sensor) and 21-40 (right sensor).  Most neurons are either fully potentiated (black) or fully depressed (white).  Motor neurons have differentiated themselves to either respond to left sensor neurons or right sensor neurons, as evident by the horizontal banding.}
	\label{fig:foodonly_weights_matrix}
\end{figure}

The final synaptic weights for all sensor neuron to motor neuron synapses for a single trial is shown in figure \ref{fig:foodonly_weights_matrix}.  There are a couple of interesting properties to note about these weights.  Firstly, a lot of synapses are either fully potentiated (and have their weight set to 4mV) or are fully depressed (and have a weight of 0mV).  This is due to the feedback mechanism inherent in STDP, if a synaptic weight starts to become potentiated then it has a greater chance of being further potentiated in future.

The second interesting property is that individual motor neurons have differentiated themselves to either only respond to the left sensor, or only respond to the right sensor.  This occurs because all the sensor neurons are stimulated at once, so if a synapse from a left sensor neuron to a left motor neuron becomes potentiated then it is likely that the left motor neuron will fire after all the left sensor neurons are stimulated, and as such all synapses from the left sensor to that left motor neuron will become potentiated.  Once the robot has learnt attraction behaviour, then a common sequence of neural firings is:
\begin{verbatim}
Left Sensor->Right Motor->Right Sensor->Left Motor->Left Sensor->Right Motor...
\end{verbatim}
As can be seen, once a motor neuron is reacting strongly to the left sensor it is more likely to fire just before the right sensor is stimulated. This acts to depress any connections from the right motor, and keep the neuron differentiated.

\section{Food-Poison Plasticity Learning}
\label{section:foodpoison}

Once the robot has learnt a behaviour, it should be able to modify its behaviour if the environment changes.  This is demonstrated by replacing food items with poison once the robot has learnt food attraction behaviour.

\subsection{Experimental Set-up}
The robot was placed in a 100cmx100cm environment consisting of 20 food items for 1000 seconds, giving the robot enough time to learn food attraction behaviour.  The food items were then replaced with poison, which produces a negative dopamine response, and the robot was run for a further 1000 seconds.  Each trial was run 50 times.  As a comparison a robot with learning disabled was also run in this environmental set-up for 50 trials.

\subsection{Results and Discussion}
\begin{figure}[htbp]
	\begin{center}
	  	\includegraphics[width=0.99\textwidth]{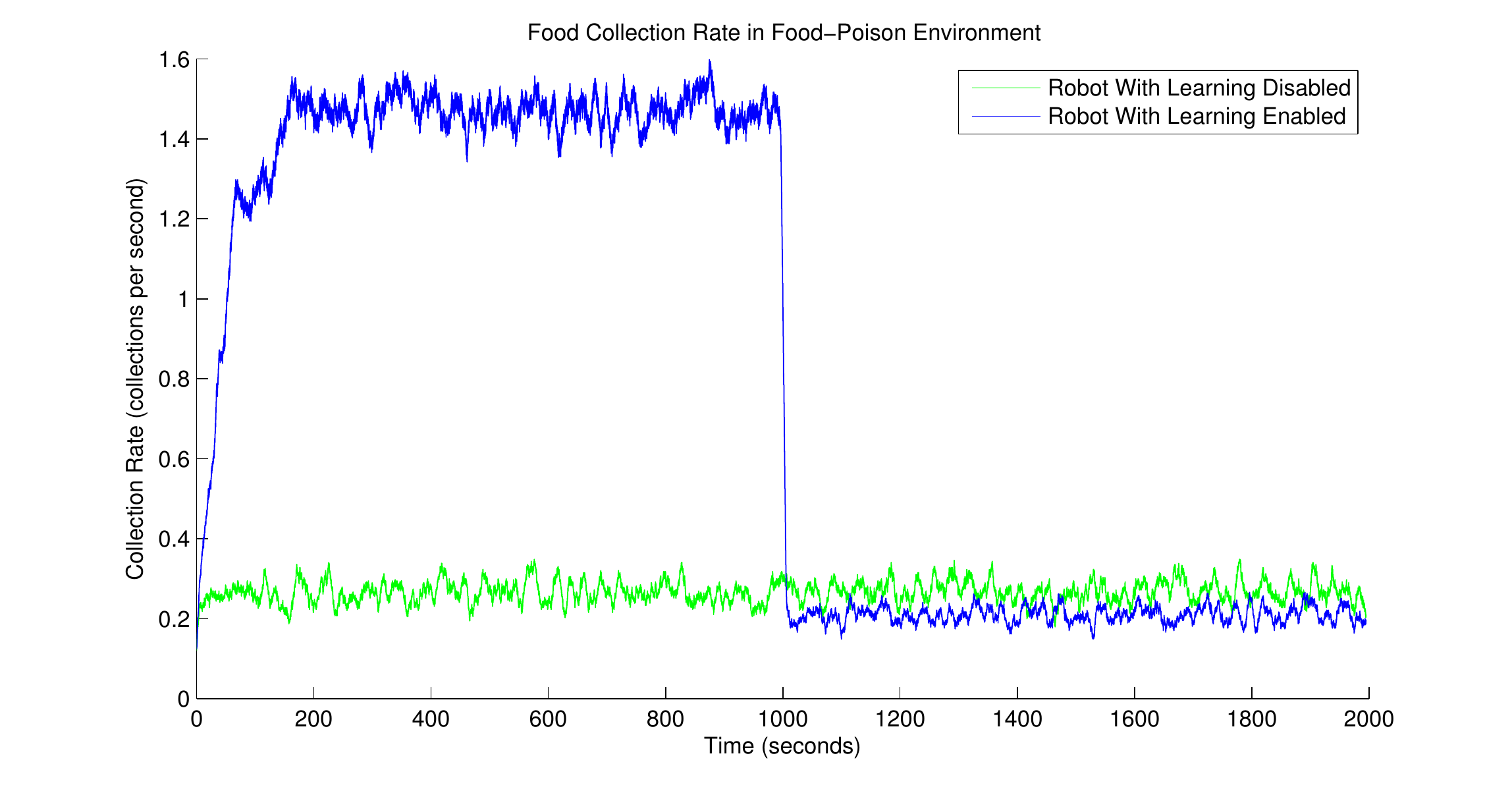}	
	\end{center}
	\caption{The collection rate over time for robots in a 100cmx100cm environment with 20 food items.  After 1000 seconds these food items are replaced with poison.  The learning robot quickly unlearns food attraction behaviour once poison is introduced.}
	\label{fig:foodpoison_scores}
\end{figure}

The collection rate over time for the dynamic food-poison environment is shown in figure \ref{fig:foodpoison_scores}.  The robot is able to unlearn the food attraction behaviour almost immediately. 

\begin{figure}[htbp]
	\begin{center}
	  	\includegraphics[width=0.95\textwidth]{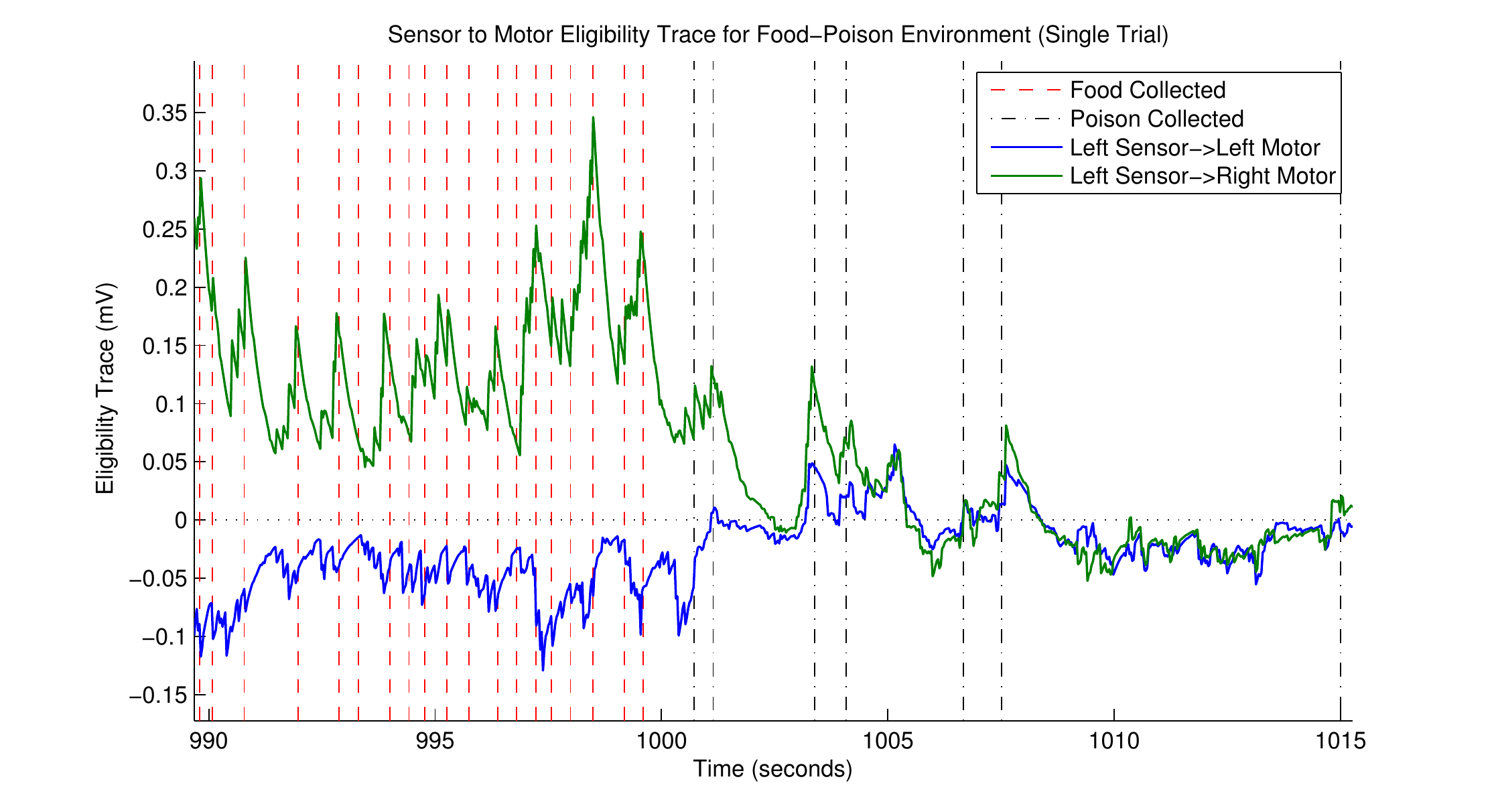}	
	\end{center}
	\caption{The mean eligibility trace for synapses from the left sensor to the left and right motors is shown.  Food items are replaced with poison at 1000ms. Before the poison is introduced the eligibility trace for the left sensor to right motor is very high.  Poison is collected at exactly the point at which the left sensor to right motor eligibility traces are highest.  After poison has been collected, and food attraction unlearnt, the eligibility trace fluctuates around zero.}
	\label{fig:foodpoison_elig}
\end{figure}

\begin{figure}[htbp]
     \begin{center}
        \subfigure[Synaptic Weights in the Food-Poison Environment]{
            \label{fig:foodpoison_weights}
            \includegraphics[width=0.9\textwidth]{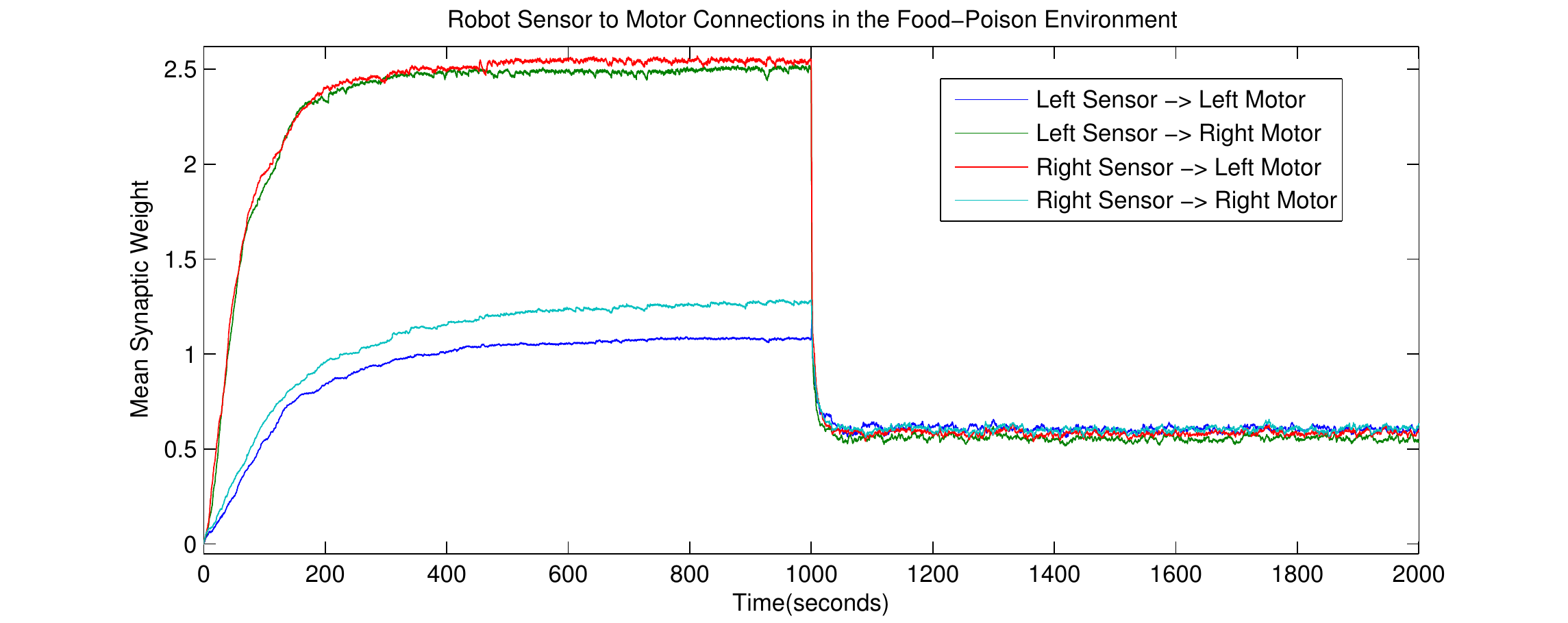}
        }
        \subfigure[Synaptic Weights in the Food-Poison Environment (Close-Up)]{
           \label{fig:foodpoison_weights_zoom}
           \includegraphics[width=0.9\textwidth]{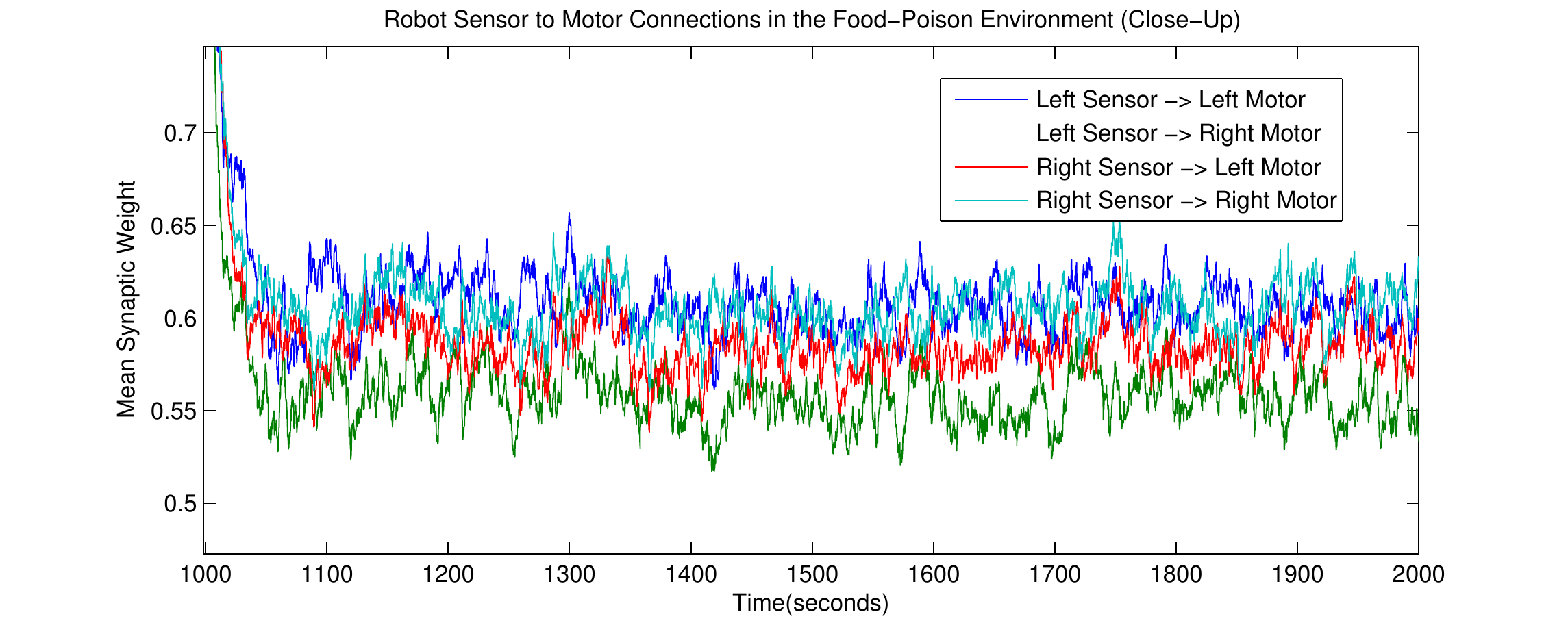}
        }
    \end{center}
    \caption{The mean synaptic weights between sensors and motors, averaged over the 50 trials in the food-poison environment is shown.  \subref{fig:foodpoison_weights} shows that after poison is introduced the synaptic weights are rapidly reduced.  \subref{fig:foodpoison_weights_zoom} shows a close-up of the last 1000 seconds, this shows that on average the food avoidance synaptic weights are very slightly higher than those for food attraction.}
   \label{fig:foodpoison_weights_all}
\end{figure}

To see how such a rapid relearning takes place it is useful to examine the eligibility trace at the point that poison is introduced, the mean eligibility trace for connections from the left sensor to the left and right motors, over the period where food is switched with poison, is shown in figure \ref{fig:foodpoison_elig}.  Due to the fact that the robot has learnt food attraction behaviour there are strong connections between the left sensor and the right motor.  This means that the eligibility trace between these two sets of neurons is very high, as a stimulation of the left sensor always causes a response in activity in the right motor.  This high eligibility trace means that when poison is introduced, and a negative spike in dopamine is received, there is a rapid decrease in the weights between the left sensor and right motor.  This is shown in figure \ref{fig:foodpoison_weights} where the mean synaptic weights between sensors and motors are plotted, averaged over all 50 trials.  There is also some reduction in synaptic weights for the left sensor to left motor and right sensor to right motor synapses.  This is because a few of these synaptic connections have become potentiated to their maximum value and as such have a high eligibility trace, the drop in weights from these saturated connections is greater than the increase in synaptic weight from uncorrelated connections.

The other thing to note about the food collection rate in this environment is that after poison is introduced the rate of food collection is only marginally less than for the robot performing a random walk, this is reflected in the synaptic weights for avoidance only being marginally higher, as can be seen in figure \ref{fig:foodpoison_weights_zoom}.  This is due to the fact that the robot is not rewarded for avoiding poison (in terms of an increase in dopamine), therefore the synapses corresponding to poison avoidance will not be potentiated.  As such all the sensor to motor synapses end up with their weights at very low values.  In this situation the exploration stimulus will override any current provided by the sensor to motor connections.  One possible way to get the robot to learn food avoidance behaviour would be to have a positive baseline level of dopamine. Though, as has been shown in section \ref{section:orbiting}, this would prevent the robot escaping orbiting behaviour, so more research would be needed to find a solution that works for both problems.

\section{Dopamine Response to Secondary Stimulus.}
In this section the ability of the robot to learn to elicit a positive dopamine response to a new stimulus that is repeatedly paired with a dopamine inducing stimulus is tested.

\subsection{Experimental Set-up}
\begin{figure}[htbp]
	\begin{center}
	  	\includegraphics[width=0.9\textwidth]{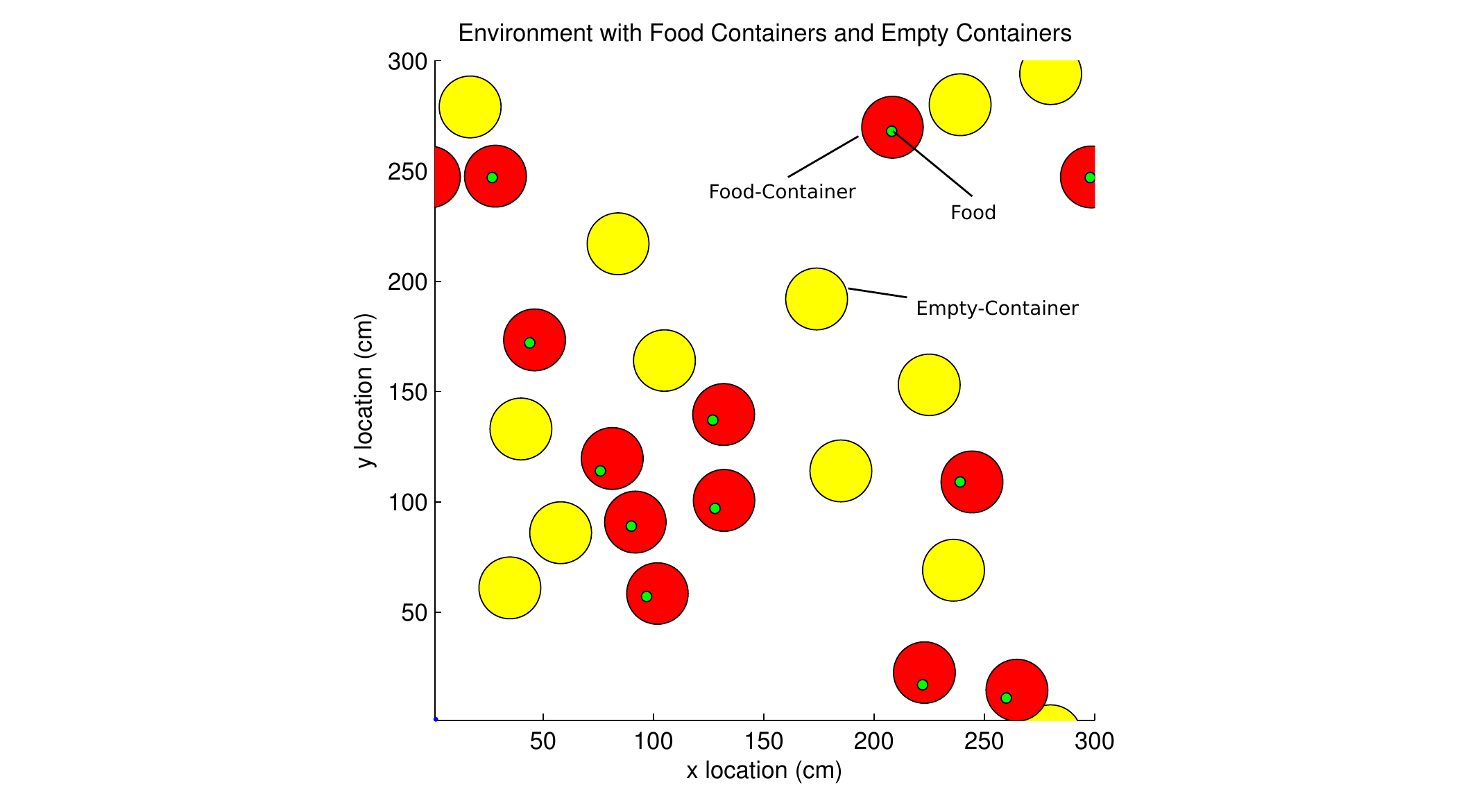}	
	\end{center}
	\caption{The environmental set-up to demonstrate secondary stimulus dopamine learning, consisting of food items, food-containers, and empty-containers.}
	\label{fig:environment_empty}
\end{figure}

\begin{figure}[htbp]
	\begin{center}
	  	\includegraphics[width=0.8\textwidth]{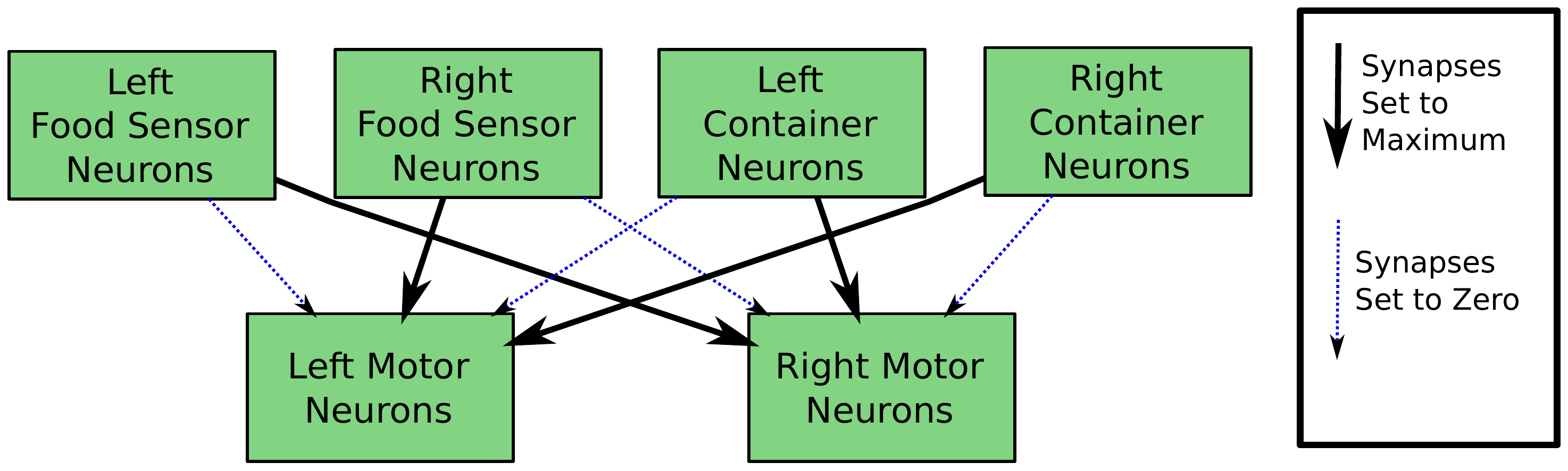}	
	\end{center}
	\caption{The starting weights for the robot to force attraction behaviour.}
	\label{fig:taxis_synapses}
\end{figure}

To demonstrate this reward predicting dopamine response, an environment consisting of food items in containers was used, see section \ref{section:network_architecture} for detail in how the architecture of the robot was changed for this environment.  The environment is 300cmx300cm and consists of 12 containers which contain food, and 12 containers which contain no food, this environment is visualized in figure \ref{fig:environment_empty}.  The robot has a different sensor for detecting if it is entering a food containing container, or an empty container.  However, the left and right container sensors cannot distinguish between the two container types.  At the start of the experiment the sensor to motor connections were set for food and container attraction behaviour.  Figure \ref{fig:taxis_synapses} shows how the synaptic weights were set to force attraction behaviour.  The result of this attraction behaviour is that the robot will visit both empty and food containing containers equally often, when inside a food container the robot will turn towards and consume the food item.

The synaptic weights between the two container touch-sensor neurons and the dopaminergic neurons were initially set to zero and plasticity was turned on for these connections.  The robot was placed in the environment and run for a 3000 second trial.  This trial was repeated 50 times.  

\FloatBarrier
\subsection{Results and Discussion}
\begin{figure}[htbp]
	\begin{center}
	  	\includegraphics[width=0.85\textwidth]{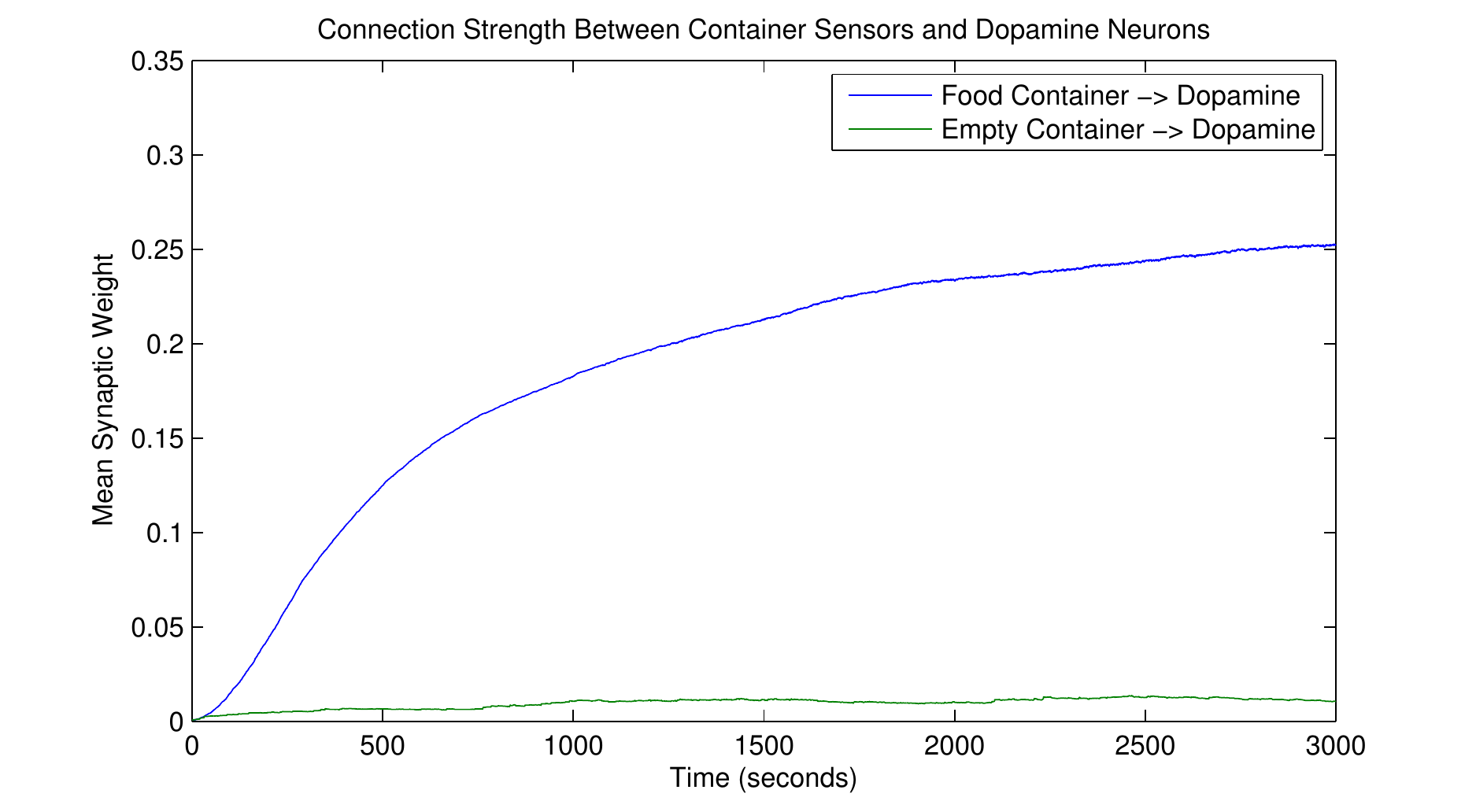}	
	\end{center}
	\caption{The mean synaptic weight over time between food-container touch-sensor neurons and dopaminergic neurons, as well as between empty-container touch-sensor neurons and dopaminergic neurons over 50 trials.}
	\label{fig:empty_container_weights}
\end{figure}

In all 50 trials the food-container touch-sensor to dopaminergic neuron synapses were potentiated by more than four times that of the empty-container touch-sensor to dopaminergic neuron synapses by the end of the trial.  Figure \ref{fig:empty_container_weights} shows the mean synaptic weight over time for the food container to dopaminergic neuron synapses and for the empty container to dopaminergic neuron synapses.  As can be seen, the synapses between the reward predicting stimulus and the dopaminergic neurons are potentiated until levelling off at a maximum value.  The non-reward predicting stimulus, by contrast, has very little potentiation in its synaptic connections to the dopaminergic neurons.  These remaining stable at just above zero.

\begin{figure}[htbp]
	\begin{center}
	  	\includegraphics[width=0.8\textwidth]{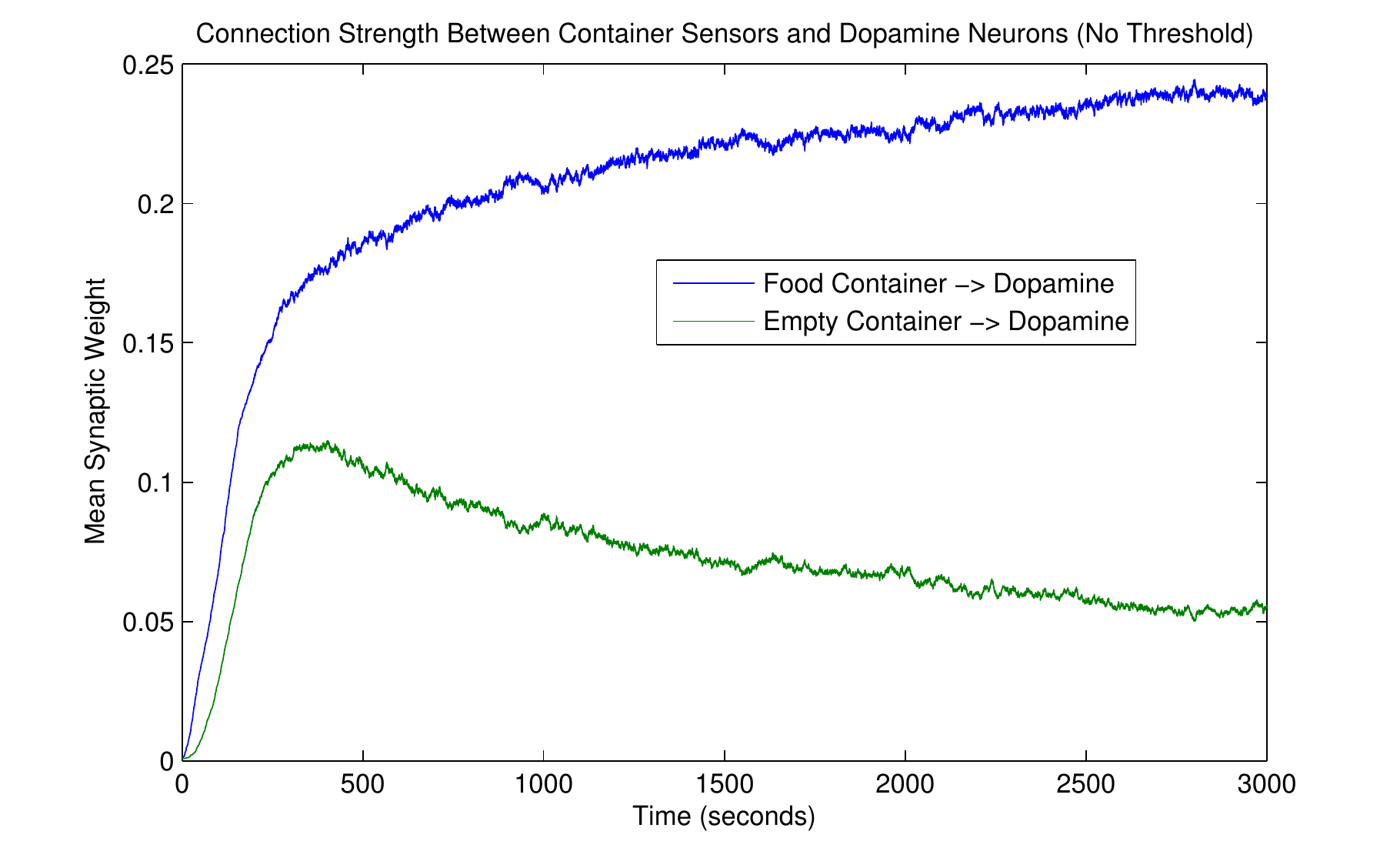}	
	\end{center}
	\caption{The synaptic weights between the empty-container sensor and dopaminergic neurons, and between the food-container sensor and dopaminergic neurons is shown. In this experiment there was no dopamine threshold, such that a single dopaminergic neuron spiking could raise the level of dopamine.}
	\label{fig:empty_container_weights_nothresh}
\end{figure}

To see the effect that dopamine thresholding (described in section \ref{section:moving_dopamine}) has on the dynamics of the system, the experiment was re-run without requiring five dopaminergic neurons to fire in order to raise the level of dopamine.  The effect of this is that background firing in the dopaminergic neurons can raise the level of dopamine.  With dopamine thresholding switched off the food-container touch-sensor to dopaminergic neuron synaptic weights at the end of the trial were higher than the empty-container touch-sensor to dopaminergic neuron synaptic weights in only 80\% of the trials, as compared with 100\% with thresholding enabled.  The synaptic weights for the experiment without thresholding are plotted in figure \ref{fig:empty_container_weights_nothresh}.

\begin{figure}[htbp]
	\begin{center}
	  	\includegraphics[width=0.9\textwidth]{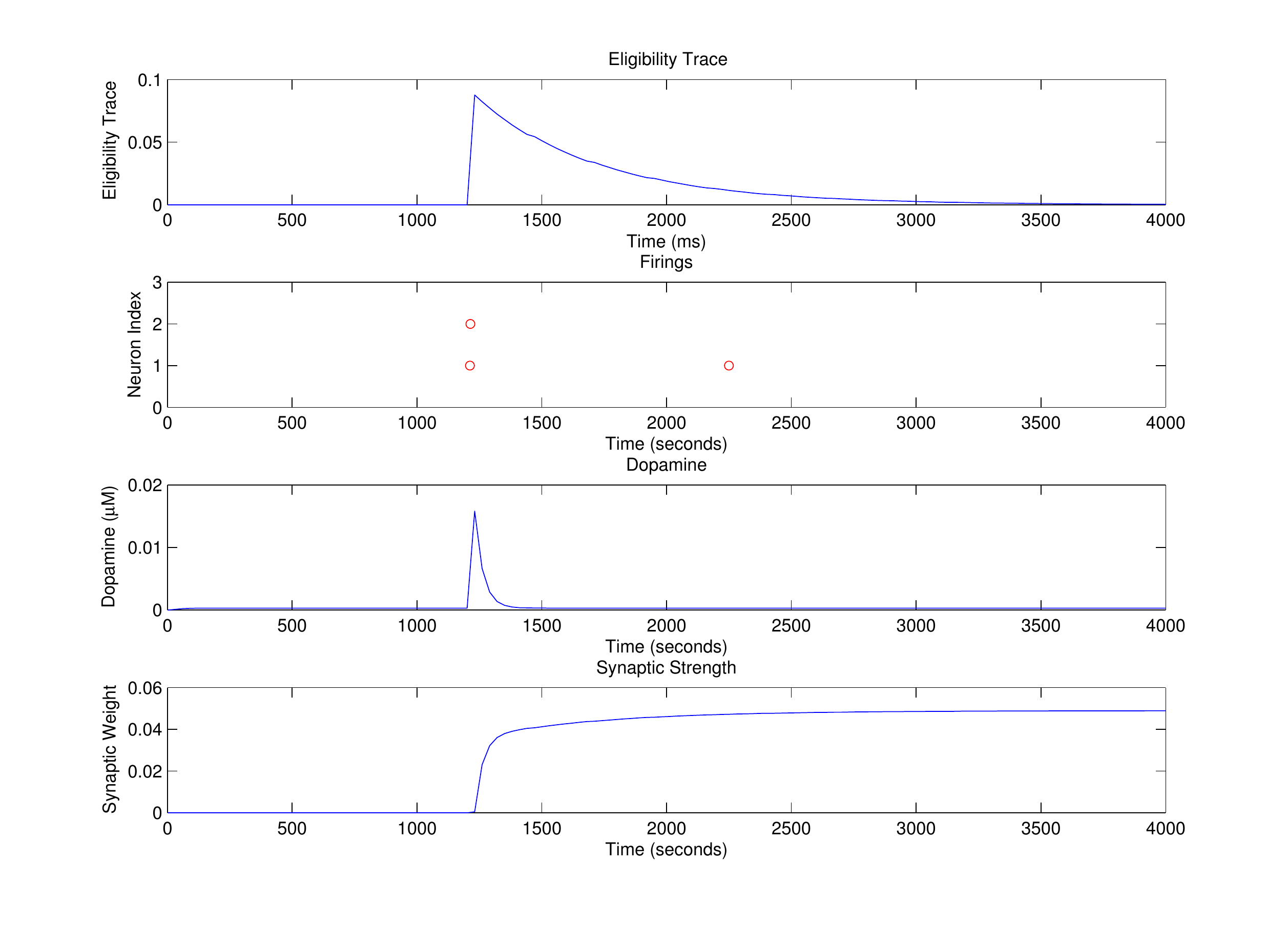}	
	\end{center}
	\caption{An example of how empty-container to dopaminergic neuron synapses can become potentiated.  Neuron 1 is a empty-container sensor neuron, and neuron 2 is a dopaminergic neuron.  At 1300ms they coincidentally fire, the corresponding spike in dopamine causes the synaptic weight to increase.}
	\label{fig:dop_synapse_example1}
\end{figure}

Note that there is in essence two processes that are happening.  For the first 400 seconds both sets of synapses are increased.  To see why both sets of synaptic weights are increased, and why the reward predicting set is increased more, it is useful to plot the eligibility trace for a single synapse.  In figure \ref{fig:dop_synapse_example1} the eligibility trace, firing rate, and synaptic strength is plotted for a single synapse from an empty-container neuron to a single dopaminergic neuron.  At 1300ms the container neuron happens to fire just before the dopaminergic neuron purely by chance, due to the background firing rate.  This causes the eligibility trace to jump up, as the second neuron is a dopaminergic neuron its firing immediately causes a spike in the level of dopamine (which would not happen with dopamine thresholding).  The higher dopamine level combined with the high eligibility trace cause the synaptic weight to increase.  This could be described as a double feedback loop, the increase in synaptic weight means that the two neurons are more likely to fire in sequence in future, meaning that their eligibility trace is more likely to increase.  While an increase in synaptic strength means that the dopaminergic neuron is more likely to fire, and hence increase the level of dopamine, at exactly the point that will raise the strength of the synapse the most.

If we consider what happens when a reward predicting stimulus (food-container touch-sensor) neuron coincidentally fires before a dopaminergic neuron, then not only do we have the same increase in dopamine and synaptic weight as described above. It is also likely that there is a very large spike in dopamine within 1000ms (due to a food item being collected), which will have the effect of raising the synaptic weight even more since the eligibility trace will still be positive.  This explains why the rate of increase of synaptic weights from the reward predicting stimulus to dopaminergic neurons is faster than for the non-reward predicting stimulus.

The second thing to note about figure \ref{fig:empty_container_weights_nothresh} is that after about 400 seconds the synaptic weights of the empty-container touch-sensor to dopaminergic synapses start to decrease.  This is due to the homeostatic mechanism described in section \ref{section:meth_stability}, the combined synaptic weights for all synapses leading to dopaminergic neurons is limited.  Once this limit is reached then all synapses leading to dopaminergic neurons have their weight reduced.  The food container to dopaminergic synapses will then increase their weight faster than the empty container to dopaminergic synapses, so the next time the weights are reduced the empty container synapses will be reduced by more than they increased in this time interval.  

Without dopamine thresholding and synaptic weight dampening there would be nothing to stop the empty-container touch-sensor to dopaminergic neuron synapses potentiating to their maximum value.  By using both of these techniques the robot is more robust and able to correctly learn to produce a spike in dopamine for a reward-predicting stimulus, but not for a non reward-predicting stimulus.

\section{Secondary Behaviour Learning}
\label{section:secondary_behaviour_learning}
In this section we explore the ability of the robot to learn a secondary behaviour.  This being a behaviour that, in and of itself, does not give a direct reward, but when succeeded by another behaviour can lead to a reward.  The behaviour to be learnt is that of food-container attraction, driving towards food-containers is an advantageous behaviour, as entering a food-container often leads to food being collected if the robot subsequently performs food-attraction behaviour.

\subsection{Experimental Set-up}
An environment consisting of 17 food-containers, each containing a single item of food, was created.  The food-containers are positioned randomly and are moved to a random location whenever their associated food item is collected.  Five different robots were run in this environment for 50 trials of 2000 seconds each.  All robots, except the benchmark robot, have their initial sensor to motor synaptic weights set for food attraction behaviour, no preference for container attraction/avoidance is set.  The result is that initially all the robots wander randomly whilst outside containers.  Whilst inside containers the robots will turn towards and collect food items. 

Note that unlike in the food-only environment, this environment has a fixed optimal robot.  This is because the size of the food-containers means that the robot cannot get stuck orbiting them.  Also, whilst in the food-container, if the robot starts orbiting a food item then its orbit will take it out of the container.  As such orbiting isn't a problem in this environment and we can set up a fixed optimal robot as a benchmark.  The dynamics of the five robots is:
\begin{itemize}
\item Learning Disabled - Plasticity is switched off for all synapses. The result is that the robot wanders randomly whilst outside containers.  Whilst inside containers the robot turns and collects food items.
\item No Container Dopamine - Connections between container touch-sensors and dopaminergic neurons are set to zero, and plasticity is switched off for these synapses.  As such no dopamine response is received when the container touch-sensor is stimulated.  Plasticity is enabled for container range-sensor to motor synapses.
\item Container Dopamine Learning Enabled - Connections between container touch-sensors and dopaminergic neurons are initially set to zero.  Plasticity is enabled for these connections, allowing them to become potentiated and for the robot to learn a dopamine response to entering a food-container.  Container range-sensor to motor synapses are plastic.
\item Fixed High Container Dopamine - Connections between container touch-sensors and dopaminergic neurons are set to their maximum value and plasticity is switched off for these synapses.  This forces a spike in dopamine whenever the robot enters a food-container.  Container range-sensor to motor synapses are plastic.
\item Benchmark - This robot has plasticity switched off.  Food attraction synapses are set to their maximal value as well as food-container attraction synapses.  All other sensor to motor synapses are set to zero.
\end{itemize}

\subsection{Results and Discussion}
\begin{figure}[htbp]
	\begin{center}
	  	\includegraphics[width=0.99\textwidth]{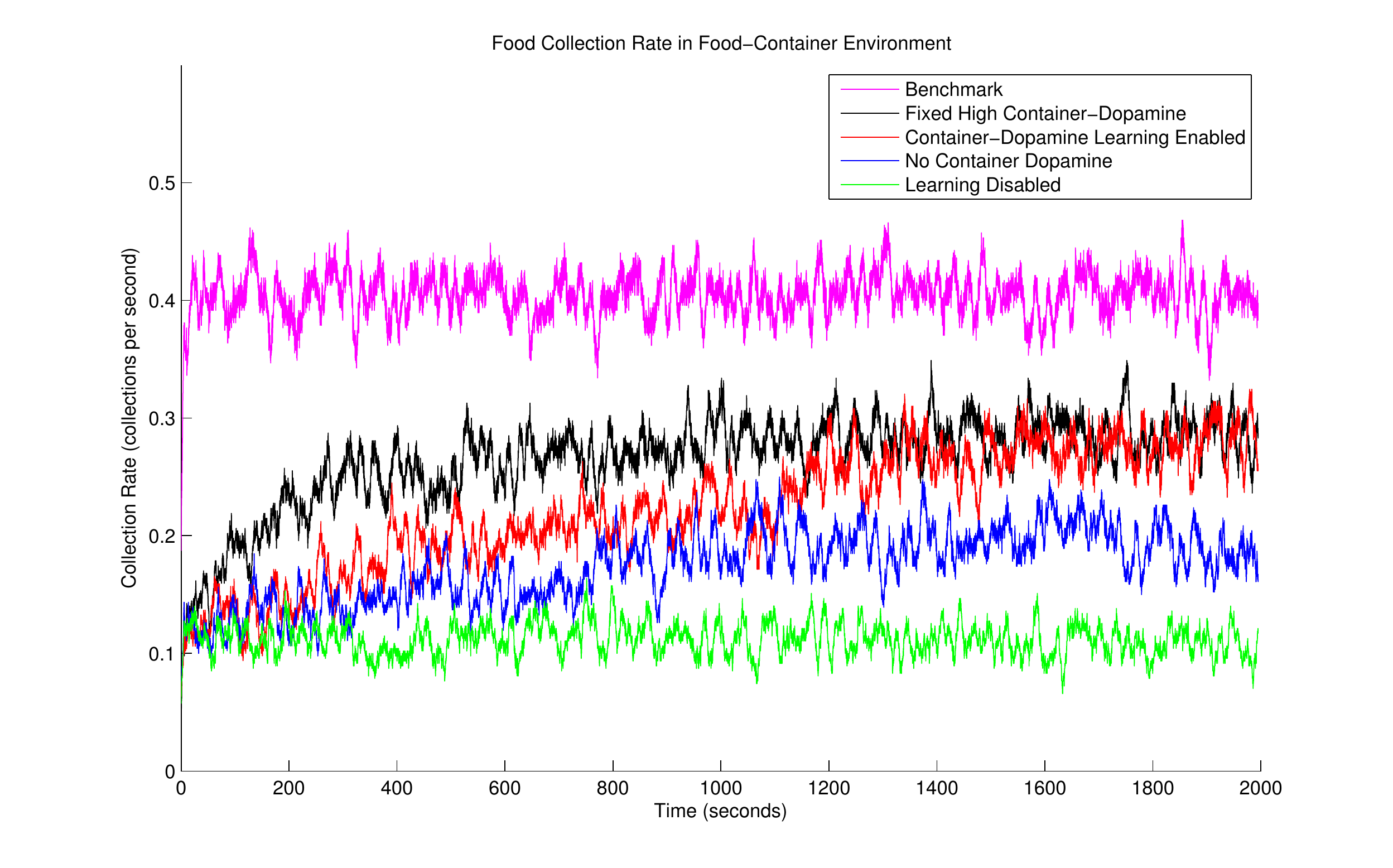}	
	\end{center}
	\caption{The averaged score over time for five different robots run in a 300cmx300cm environment with 20 food-items contained in containers.}
	\label{fig:foodcontainer_scores_contdop}
\end{figure}

In figure \ref{fig:foodcontainer_scores_contdop} the average score over time for each of the five robots is plotted.  Correspondingly, the total food collected as well as the percentage of robots that had learnt food-container attraction behaviour by the end of the trial is shown in table \ref{tab:foodcontainer_scores_slow}.  A robot was deemed to have learnt container-attraction behaviour if, at the end of the trial, the left container range-sensor to right motor neuron synapses where stronger than 0.5mV on average, as well as being more than 10\% stronger on average than the left container range-sensor to left motor neuron synapses.  The equivalent also had to be true for the right container range-sensor synapses for the robot to be deemed correct.

\begin{table}[htbp]
\centering
\begin{tabular}{|p{4cm}|l|l|l|}
\hline
Robot Type & Mean Food Collected & Std. Deviation Food Collected & \% Correct \\ \hline
Learning Disabled & 224 & 32.8 & 0 \\ \hline
No Container-Dopamine & 348.5 & 87.5 & 50 \\ \hline
Container-Dopamine Learning Enabled & 446.7 & 103.4 & 96 \\ \hline
Fixed High Container-Dopamine & 529.7 & 152.5 & 94 \\ \hline
Benchmark & 808.4 & 21.5 & 100 \\ \hline
\end{tabular}
\caption{This table shows the total amount of food collected for the three robots, averaged over the 50 trials.}
\label{tab:foodcontainer_scores_slow}
\end{table}

There are several interesting properties shown in these figures.  Firstly, note that all the robots managed to perform better than the robot with learning disabled that performed a random walk whilst outside the containers.  This means that even the robot which did not receive any dopamine when the container touch-sensors were stimulated (\emph{No Container Dopamine}) was able to learn some food-container attraction behaviour, from table \ref{tab:foodcontainer_scores_slow} we can see that it managed to learn container attraction in 50\% of cases.  This is due to the fact that there is on average 800ms between entering a container and food being collected.  This is small enough that DA-modulated STDP can still have a small effect on potentiating the synapses that cause container attraction behaviour.

The next thing to notice is that both the robot with fixed strong weights between food-container sensors and dopaminergic neurons (\emph{Fixed High Container Dopamine}), and the robot that has to learn to potentiate these synapses (\emph{Container Dopamine Learning Enabled}), perform better than the robot without any dopamine response to the food-container touch-sensor.  They are both able to learn food-container attraction behaviour in about 95\% of cases.  This shows that by being able to learn to elicit a dopamine response to a reward-predicting stimulus helps in being able to learn a sequence of behaviours needed for the agent to get a reward.

\begin{figure}[htbp]
	\begin{center}
	  	\includegraphics[width=0.8\textwidth]{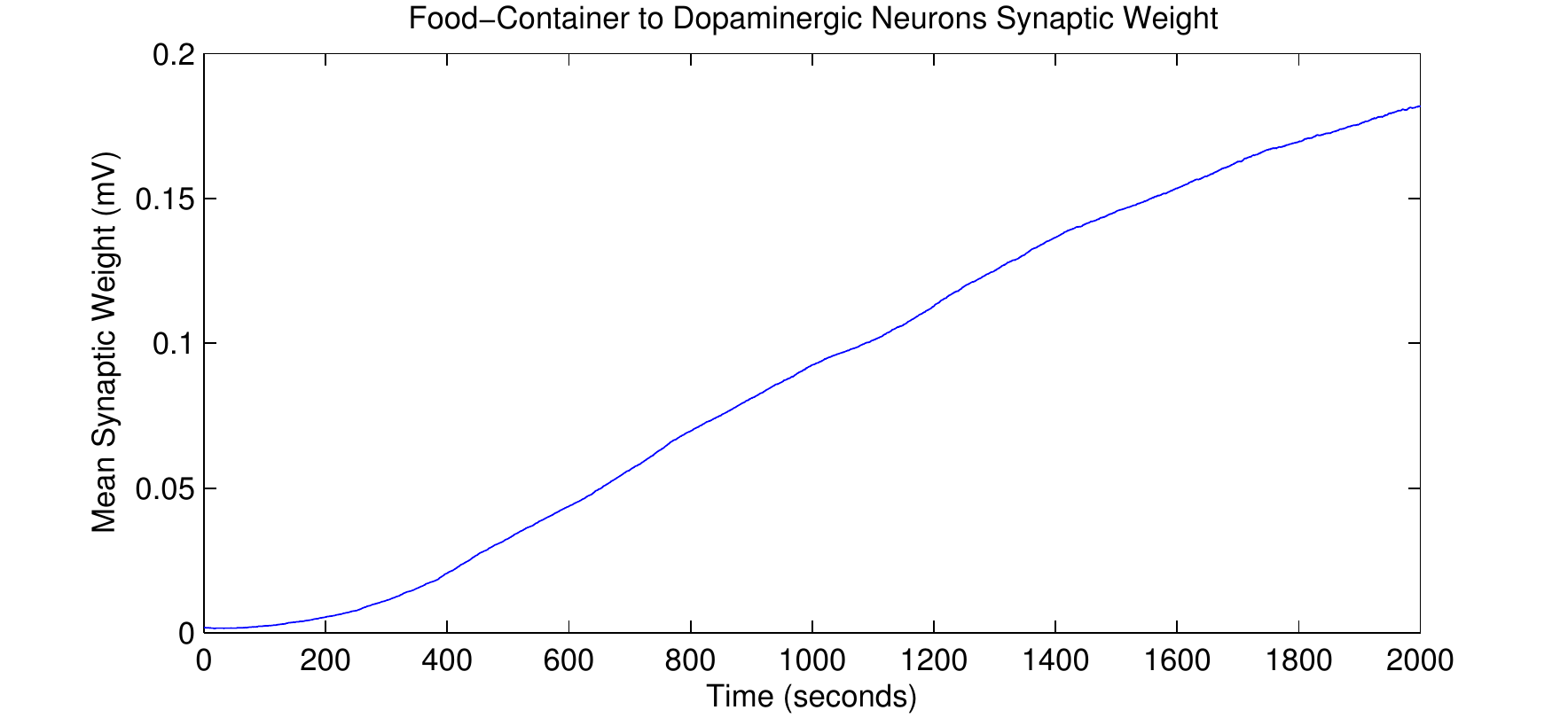}	
	\end{center}
	\caption{The average synaptic weight between the food-container touch-sensor neurons and the dopaminergic neurons, over the 50 trials, is shown for the \emph{Container Dopamine Learning Enabled} robot.}
	\label{fig:foodcontainer_dopweights_contdop}
\end{figure}

As expected, the robot which starts with its food-container touch-sensor to dopaminergic neurons already fully potentiated is able to learn food-container attraction behaviour faster than the robot which has to learn to potentiate these neurons before stimulating the food-container touch-sensor will elicit a dopamine response.  Figure \ref{fig:foodcontainer_dopweights_contdop} shows the mean synaptic weight between the food-container touch-sensor and dopaminergic neurons for the robot with container to dopamine plasticity enabled.  As can be seen the robot is able to correctly potentiate these synaptic weights.  As these synapses potentiate the robot is able to learn food-container attraction behaviour, this is shown by the fact that food collection rate for this robot slowly rises until it is equal to the robot that had its container touch-sensor to dopaminergic synapses strong from the start.

\begin{figure}[htbp]
	\begin{center}
	  	\includegraphics[width=0.8\textwidth]{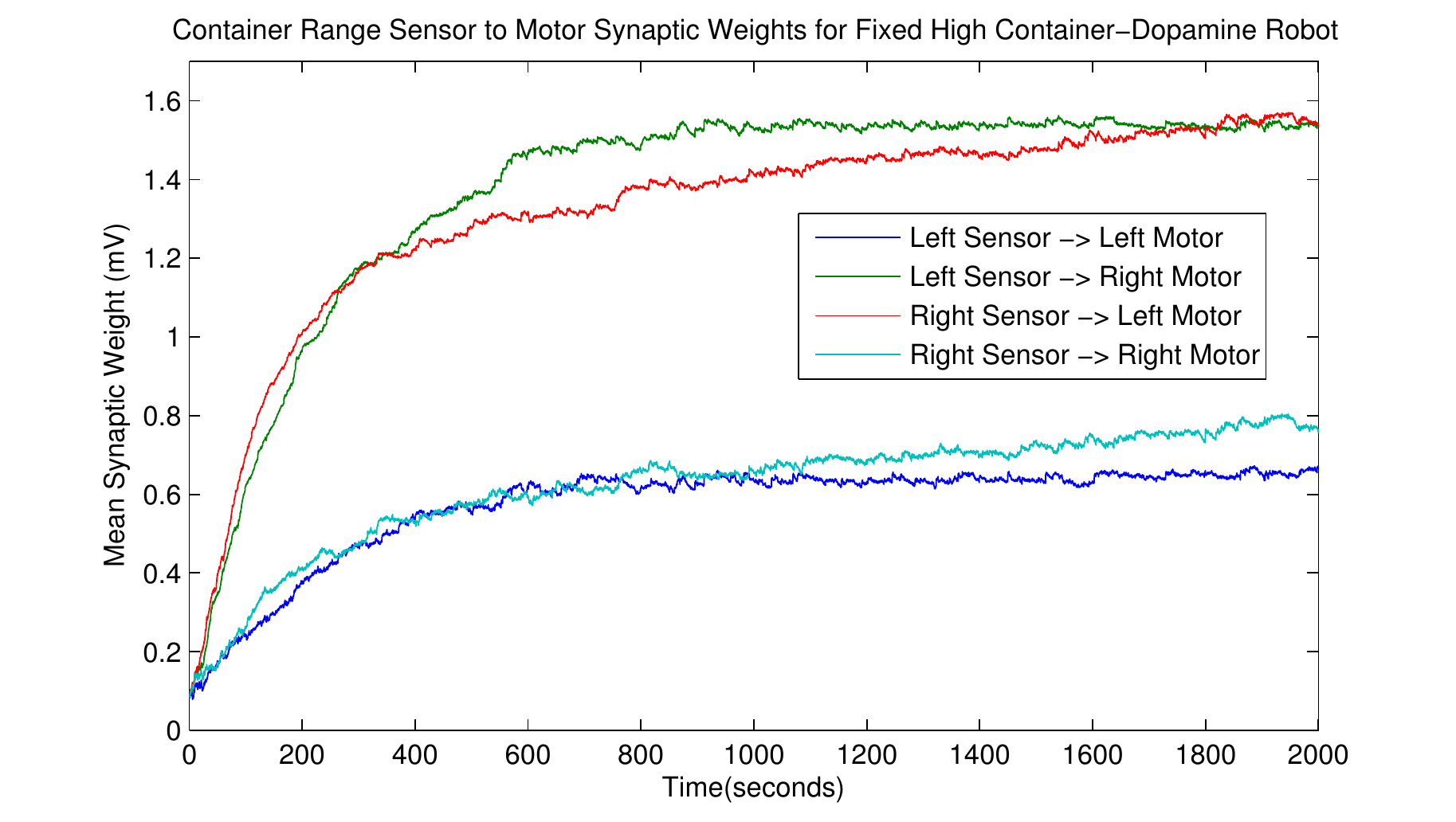}	
	\end{center}
	\caption{Mean synaptic weights between container sensor neurons and motor neurons for the \emph{Fixed High Container-Dopamine} robot in the food-container environment.}
	\label{fig:foodcontainer_highdop_weights}
\end{figure}

The last thing to note about the food collection rate in figure \ref{fig:foodcontainer_scores_contdop} is that none of the robots with learning enabled are able to match the food collection rate of a hard-coded optimal robot.  To see why this is the case, the synaptic weights for container range-sensor to motor synapses have been plotted in figure \ref{fig:foodcontainer_highdop_weights} for the robot with a fixed high container touch-sensor dopamine response (\emph{Fixed High Container-Dopamine}).  The mean synaptic weights for container attraction plateau at a value less than the maximum possible. With 85\% connection probability and a maximum single synapse weight of 4mV, the maximum mean synaptic weight between a group of sensor neurons and a group of motor neurons is 3.4mV.  The reason that they level off at a lower value is due to the motor neuron differentiation that was discussed in section \ref{section:food_attraction_learning_results} (figure \ref{fig:foodonly_weights_matrix}).  Some motor neurons have differentiated to only respond to container-avoidance behaviour and as such will not become potentiated.

In addition the synapses for container avoidance have also had their synaptic weights increased, though to a lesser extent, this is due to the negative baseline level of dopamine.  This mechanism was discussed in more depth in section \ref{section:food_attraction_learning_results}.  These two mechanisms are needed for the robot to be able to modify its behaviour if the environment changes.  As a direct consequence of this ability to cope with a dynamic environment, the robot does not perform as well as a robot with a fixed, optimal, set of weights for this environment.  The greater difference in synaptic weights allows the benchmark robot to respond to very weak sensory input, from the edge of the sensors range, and as such perform taxis from further away.

\begin{figure}[htbp]
	\begin{center}
	  	\includegraphics[width=0.95\textwidth]{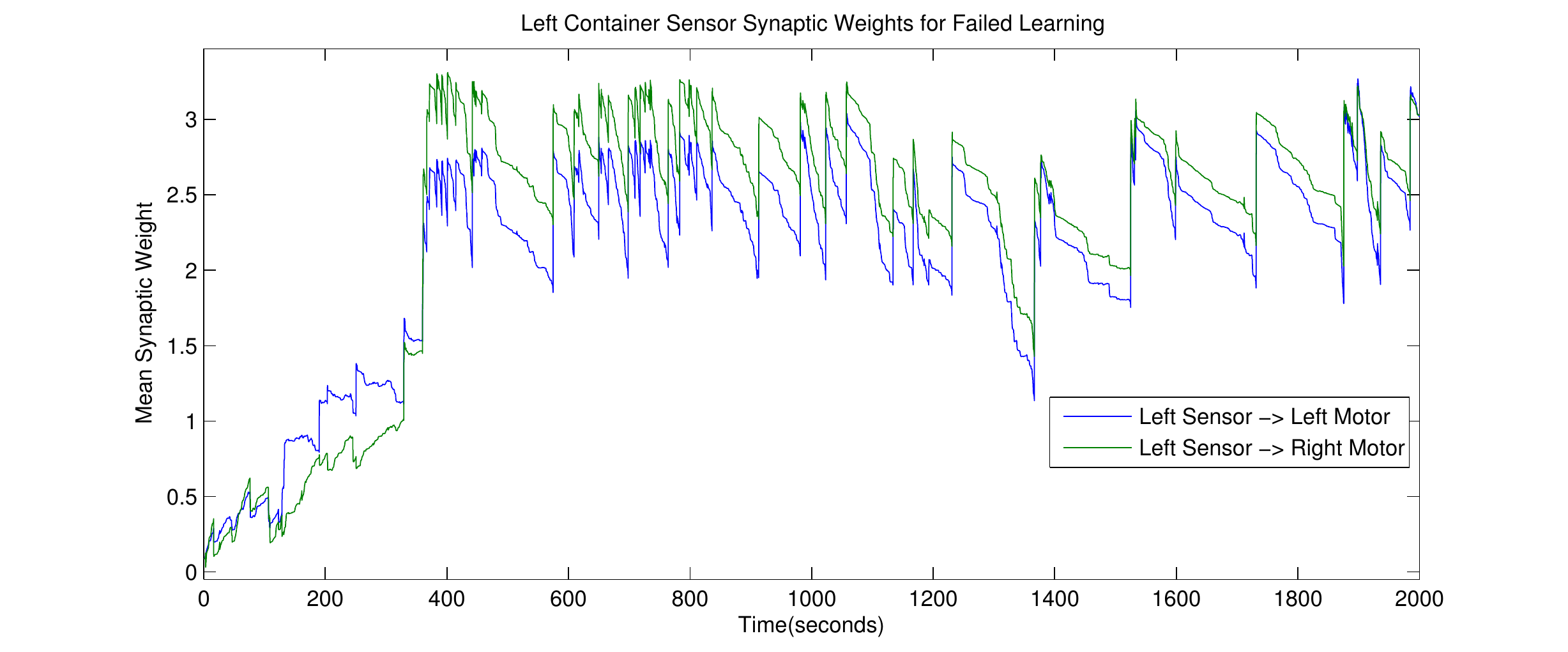}	
	\end{center}
	\caption{The synaptic weights between the left container sensor and the left and right motor neurons is shown for an example where the robot was unable to learn container attraction behaviour.  The weights have become synchronized and rise and fall together.}
	\label{fig:learn_failure_weights}
\end{figure}

\FloatBarrier
It is interesting to look at what happens in the case where the robot does not learn the correct behaviour. Figure \ref{fig:learn_failure_weights} shows the mean synaptic weights over time from the left container sensor to both motors, for a single trial of the \emph{Fixed High Container Dopamine} robot, in which the robot failed to learn correctly.  What can be seen from this figure is that if the synaptic weights from sensor neurons to both motors are close in value and both strong then the synaptic weights are increased and decreased in a synchronous fashion.  The reason for this is that if the left sensor is connected strongly to both motors, then stimulating the left sensor will have the same effect on all motor neurons, they will fire in response.  This means that the eligibility traces between the sensor and the two motors will all increase together and as such the synaptic weights will increase or decrease together.  In this way, once the synapses for avoidance and attraction have both become potentiated, it is hard for the robot to learn to only increase the synaptic weights needed for attraction behaviour.  

The cause for both sets of synaptic weights to become high in the first place is that it is still possible for the robot to collect food items even if it isn't performing container attraction behaviour.  For example, if the robot was just to constantly drive straight then purely by chance it will run into a food container and as such have the synaptic weights between all sensors and all motors increased.  Due to random exploration it may happen that this will occur a few times in a row, at which point the robot will get 'stuck' performing this sub-optimal behaviour.

\section{Dual Behaviour Learning}
Finally, we demonstrate the ability of the robot to learn two different, sequential, behaviours at the same time.  These behaviours being food-attraction and container-attraction.

\subsection{Experimental Set-up}
The same environmental set-up as in section \ref{section:secondary_behaviour_learning} was used; 20 food containers randomly positioned in a 300cmx300cm environment.  Three different robots were run in this environment for 50 trials, with 5000 seconds for each trial.  The three robots were:
\begin{itemize}
\item Random Walk - This robot had all sensor to motor connections set to zero and plasticity switched off, food is only collected by randomly driving into it.
\item Benchmark - This robot has all synaptic weights for food-attraction and container-attraction behaviour set to their maximum value.  All other sensor to motor synaptic weights were set to zero, and plasticity was switched off.
\item Learning Enabled - Initially all sensor to motor synapses are set to zero, plasticity is enabled for these synapses.  In addition food-container touch-sensor to dopaminergic neuron synapses were initially set to zero, and plasticity was enabled for these synapses.
\end{itemize}
\subsection{Results and Discussion}
\begin{figure}[htbp]
	\begin{center}
	  	\includegraphics[width=0.99\textwidth]{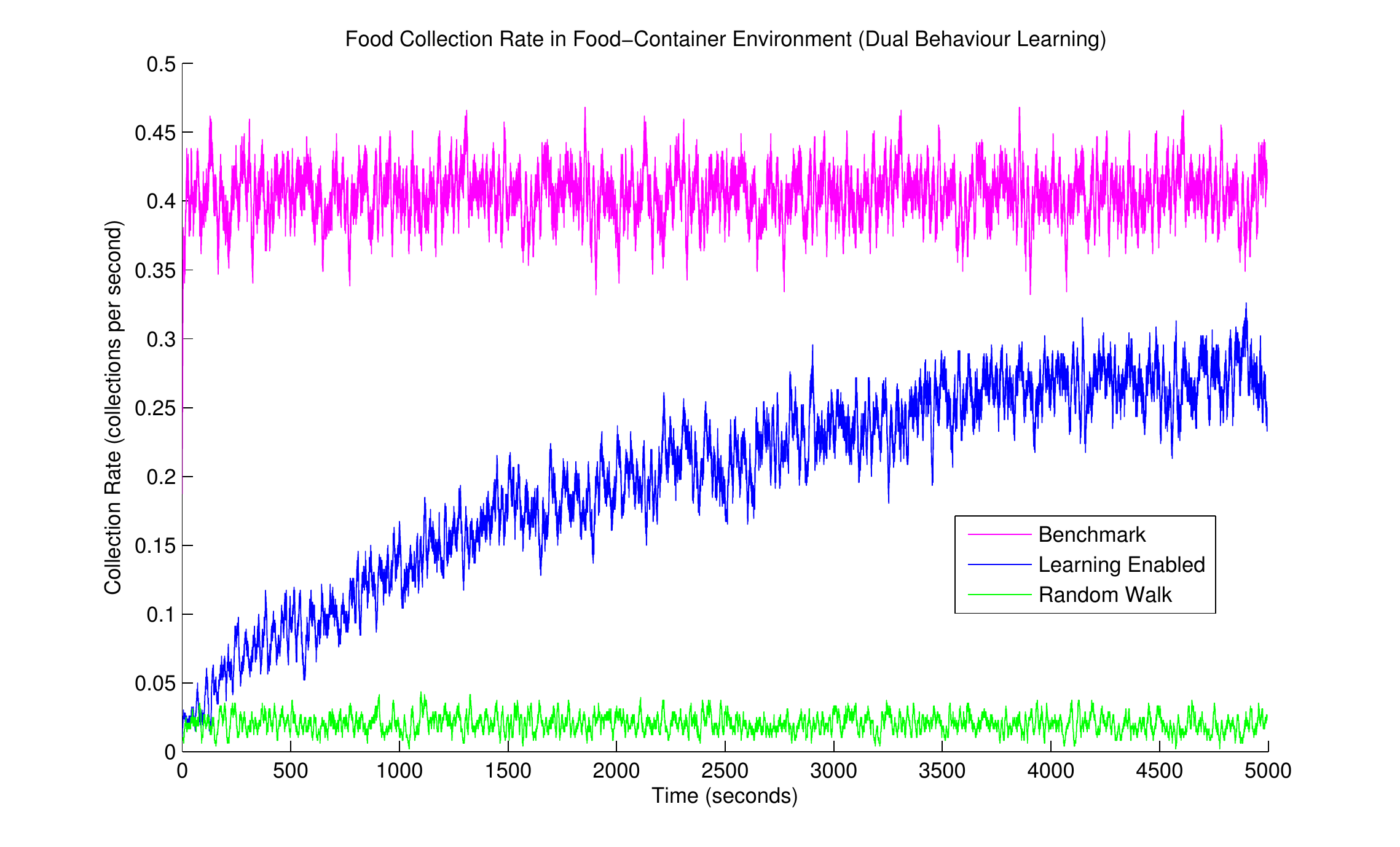}	
	\end{center}
	\caption{The collection rate over time, averaged over all trials, for each of the robots is shown.  The robot with learning enabled increases its collection rate over time.}
	\label{fig:foodcontainer_scores_dual}
\end{figure}

The average food collection rate over time for the three robots is shown in figure \ref{fig:foodcontainer_scores_dual}.  The collection rate for the learning robot, after 5000 seconds, is comparable to the robots with food-taxis behaviour hard-coded in the previous section (see figure \ref{fig:foodcontainer_scores_contdop}).  The learning robot was able to learn both food-attraction and container-attraction behaviour (the same definition of correctness as in sections \ref{section:food_attraction_learning_results} and \ref{section:secondary_behaviour_learning} was used) in 82\% of the trials.  This is slightly worse than the 94\% figure achieved when food-attraction behaviour was hard-coded.  This is to be expected, as there will be longer periods between food collection and as such there is more chance that random fluctuations in synaptic weight cause two sensor to motor groups to become synchronized, as described in section \ref{section:secondary_behaviour_learning}.

\begin{figure}[htbp]
     \begin{center}
        \subfigure[Food Sensor to Motor Synaptic Weights]{
            \label{fig:foodcontainer_foodweights_dual}
            \includegraphics[width=0.9\textwidth]{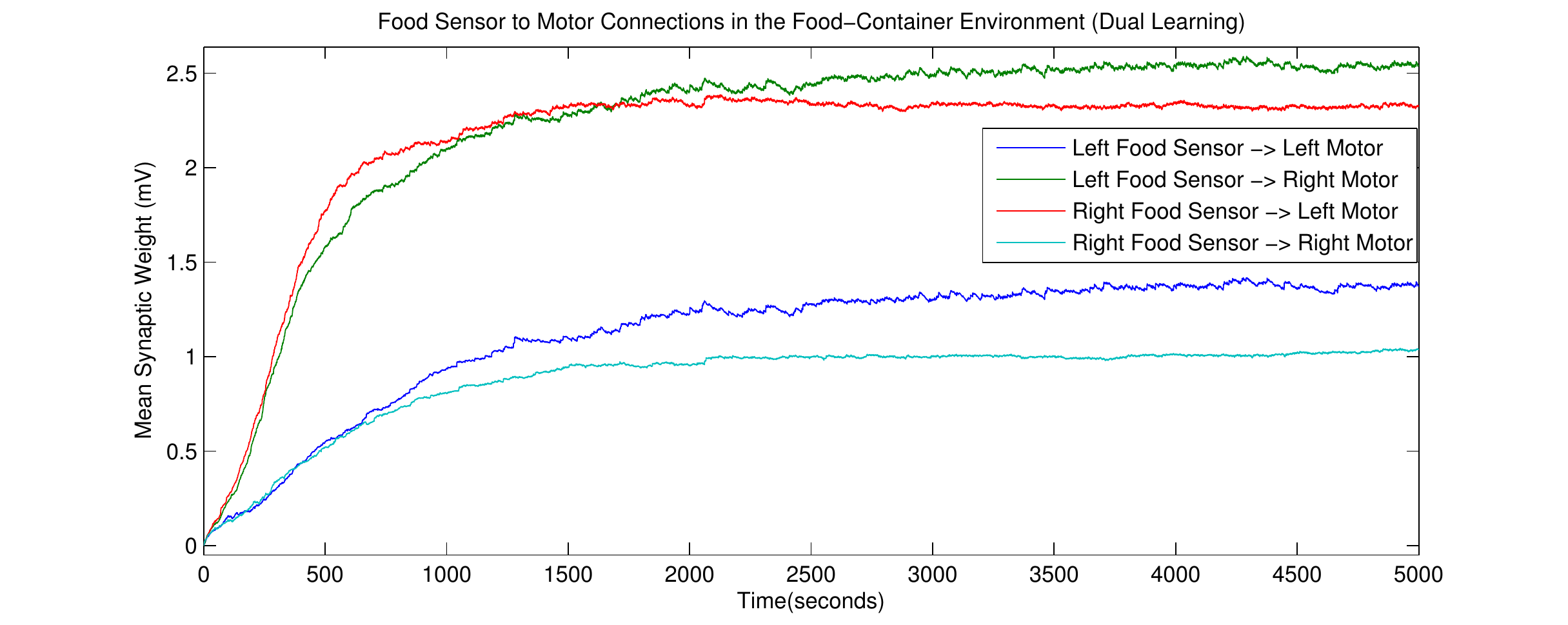}
        }
        \subfigure[Container Sensor to Motor Synaptic Weights]{
           \label{fig:foodcontainer_containerweights_dual}
           \includegraphics[width=0.9\textwidth]{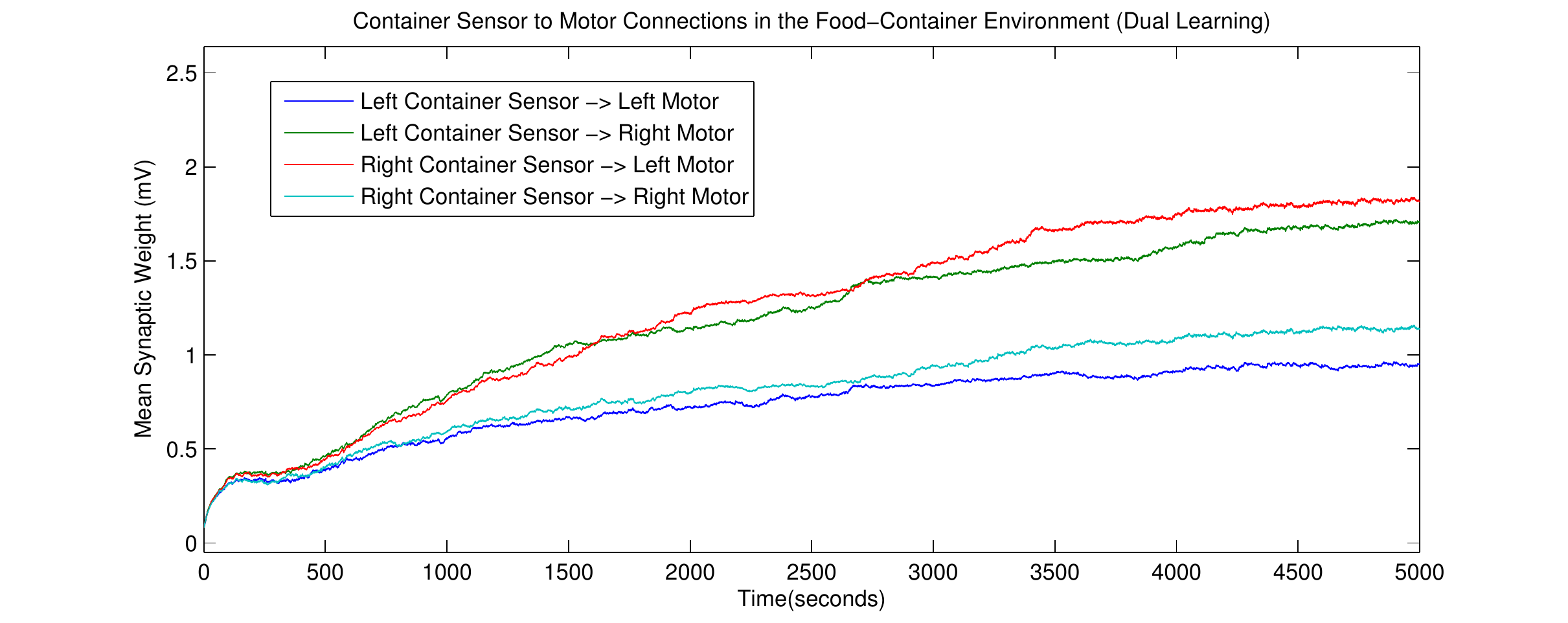}
        }
        \subfigure[Food-Container Touch-Sensor to Dopaminergic Neurons Synaptic Weights]{
           \label{fig:foodcontainer_touchweights_dual}
           \includegraphics[width=0.9\textwidth]{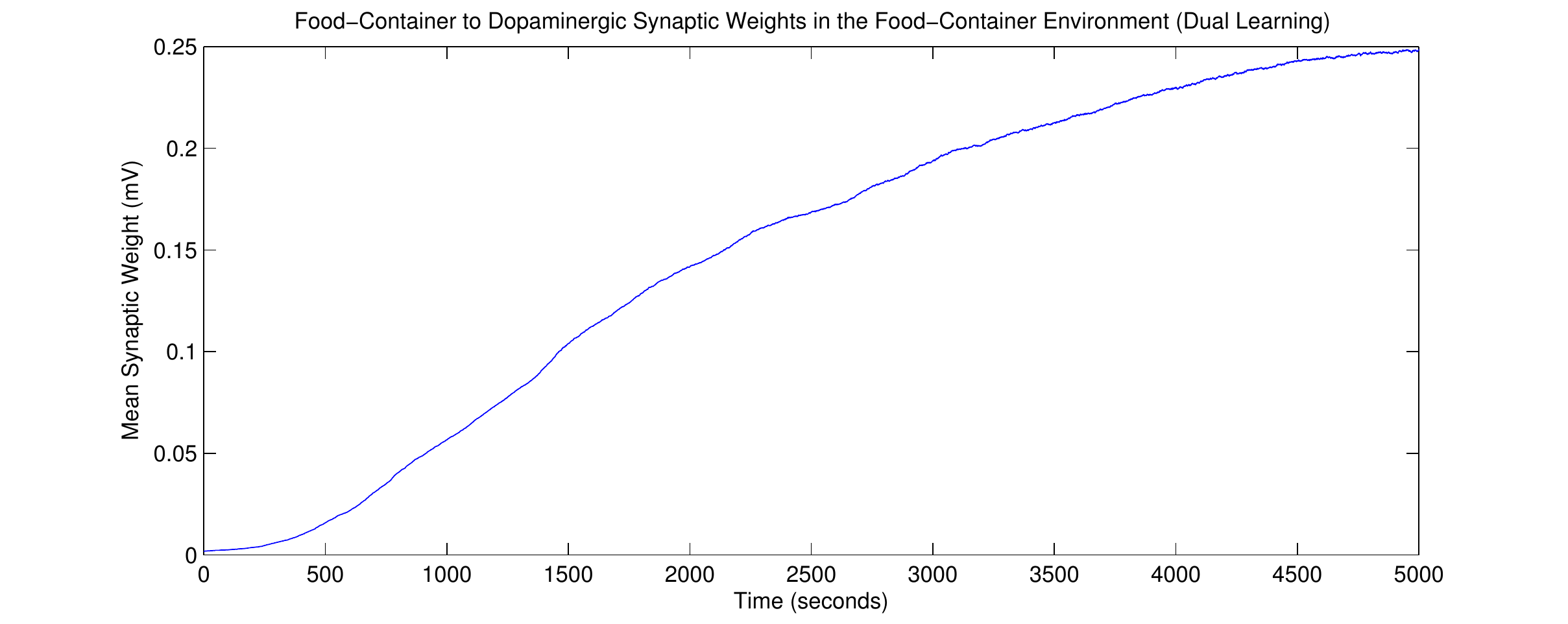}
        }
    \end{center}
    \caption{Synaptic weights for the dual learning robot in the food-container environment.  \subref{fig:foodcontainer_foodweights_dual} shows the mean synaptic weights between the food range-sensors and the motors.  The synapses for food-attraction behaviour are quickly potentiated.  \subref{fig:foodcontainer_containerweights_dual} shows the mean synaptic weights between the food-container range-sensors and the motors. The synapses for container-attraction behaviour are potentiated, but at a slower rate.  \subref{fig:foodcontainer_touchweights_dual} shows the mean synaptic weight between the food-container touch-sensor and the dopaminergic neurons.  Once food-attraction behaviour is learnt, these synapses are slows potentiated.}
   \label{fig:foodcontainer_dual}
\end{figure}

In figure \ref{fig:foodcontainer_dual} the mean synaptic weights for the learning robot, averaged over the 50 trials, is shown.  The first behaviour to be learnt is food-attraction, as shown by the quick potentiation of the corresponding synapses.  After this, the container-dopamine response and container-attraction behaviour are learnt in parallel.  With the container-attraction behaviour being reinforced by the corresponding increase in release of dopamine.
\chapter{Conclusions}
\label{ch:conclusion}

\section{Reinforcement Learning}
\label{section:conclusion_reinforcement_learning}

Comparing the properties of our implementation with classical reinforcement learning algorithms (an outline of which is given in section \ref{section:reinforcement_learning_in_machine_learning}) we can see that several properties of our model have correlates with the TD-Error reinforcement learning algorithms.  The most obvious of which is the eligibility trace that is stored within the synapses.  This directly correlates with the eligibility trace as it is defined in standard reinforcement learning algorithms such as Sarsa($\lambda$).  The role of this eligibility trace in both situations is to allow the agent to learn much faster, and over a greater time frame.

If the case of our novel exploration strategy whereby we randomly stimulate the motors with a small current, this can be considered as analogous to off-policy reinforcement learning algorithms.  Whereby the probability of choosing an action is related to the expected reward.  In our implementation, as the strength of a behaviour increases, and as such the confidence that it will produce the best reward increases, the probability of performing exploratory behaviour decreases.

In reinforcement learning algorithms the value function is propagated through the state space.  For example, initially only the goal state will have a positive value associated with it.  After repeated exploration of the environment then the states close to the goal state will have positive value function.  This is similar to the way in which our network is able to learn a dopaminergic neural response for stimulus that often precede food collection.

The key difference between our implementation and reinforcement learning algorithms is that in standard reinforcement learning algorithms the value function is updated based on the difference between expected and received reward.  For example if the value function predicts a reward of 0.8 and the agent subsequently receives a reward of 0.4, then the value function is decremented.  The value function can be thought of as analogous to the strength of connections to dopaminergic neurons.  The model used in this paper is not able to produce this negative response when a trained reward-predicting stimulus is subsequently stimulated but not paired with a reward.  For example, if a robot is trained in the food-container environment and then we remove the food items from the food-containers.  The network will still elicit a dopamine response when the food-container touch-sensor is stimulated and as such food-container attraction behaviour will not become unlearnt.  This is the main limitation of our implementation and possible solutions are discussed in section \ref{ch:future_work}.

Our implementation provides a method of reinforcement learning in continuous parameter spaces.  Standard reinforcement learning algorithms are designed for state and actions spaces that have been discretized.  Being able to operate with continuous spaces directly allows an agent to generalize much more easily when a previously unseen environmental state is encountered.

\section{Biological Plausibility and Conclusions}

In this paper we have shown that by allowing the baseline level of dopamine to be negative the robot is able to relearn when its current behaviour is not producing a reward.  In the brain the level of dopamine cannot become negative, though it may be that there is another type of pain neurotransmitter that acts in the reverse way to dopamine and as such has the effect of reducing highly active synapses, though whether or not one exists is currently unknown.

One consequence of our model was that over time motor neurons tended to differentiate themselves to only respond to a single sensor.  This is an interesting property as it provides a mechanism as to how a group of neurons in the brain can self organize into functionally different subgroups.

In the brain it has been shown that dopaminergic neurons have two different states of activity, these being background firing and bursting activity when stimulated, such that background firing does not result in a significant level of dopamine increase~\cite{Lisman_1997}.  In our model we have shown that by separating dopaminergic neuron activity into these two distinct groups the robot is able to learn a dopamine response for a reward-predicting stimulus much more reliably.  This provides one possible explanation for why dopaminergic neurons have these two different firing states.

\chapter{Future Work}
\label{ch:future_work}

\begin{figure}[htbp]
	\begin{center}
	  	\includegraphics[width=0.4\textwidth]{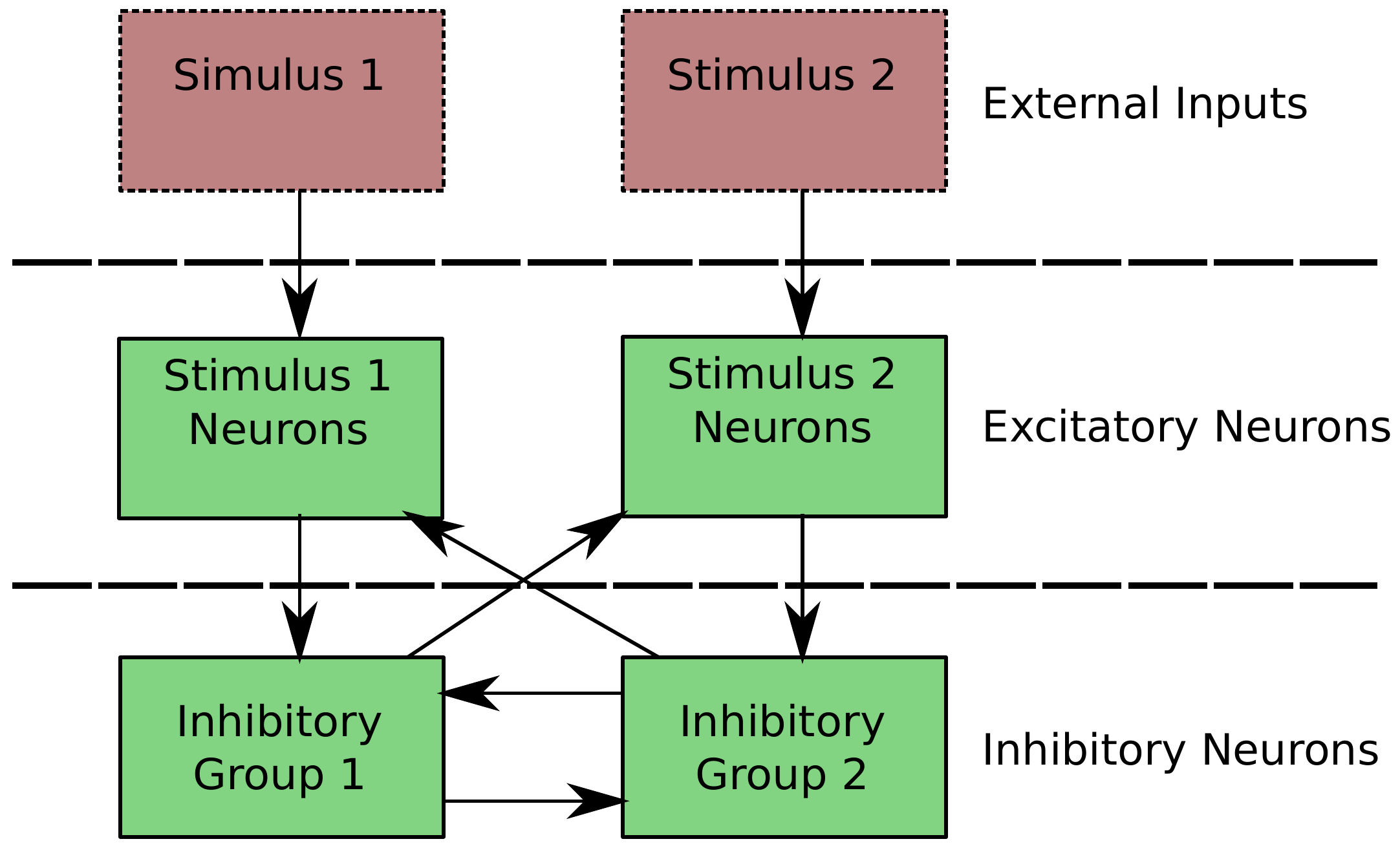}	
	\end{center}
	\caption{An example of how a winner-takes-all mechanism can be implemented in a SNN.  Each set of stimulus neurons acts to inhibit the other until the network reaches a stable state where only one group is active.}
	\label{fig:wta_neural_network}
\end{figure}

Several mechanisms that our robot used were explicitly programmed, separate to the neural network itself.  It would be advantageous to provide an implementation of these within the spiking neural network.  Such that they can be subject to plasticity and modified in response to environmental changes along with the rest of the network.  These mechanisms, and a possible SNN implementation are given below:
\begin{itemize}
\item Phasic Activity - The sensor neurons were stimulated at a rate of 14Hz, rather than being given a constant input current.  EEG recordings show that the brain also exhibits neural oscillatory behaviour.  A group of interconnected inhibitory neurons when provided constant input current will generate oscillatory firing.  This group of neurons could be used a control group and connected to the sensory neurons, forcing them to fire in an oscillatory manner.
\item Winner-Takes-All - We explicitly implemented a winner takes all mechanism between sensors and between motors.  One way in which this could possibly be implemented directly in the SNN is shown in figure \ref{fig:wta_neural_network}.
\item Exploration - We explicitly stimulated the motors to provide exploratory behaviour.  However, if the phasic activity and winner-takes-all mechanisms were implemented directly in the network as described above then no explicit exploration behaviour would be needed.  The combination of background firing and winner-takes-all would ensure that only one set of motor neurons was active at any one time, whilst the phasic activity would allow the active motor to optionally switch during the inactive phase of the neurons.
\end{itemize}

The main limitation in the approach taken in this paper is that the robot cannot unlearn an already learnt dopamine response.  To be able to do this a fuller SNN implementation of the TD-error algorithm would need to be implemented, whereby the dopamine response would correspond to the difference between expected reward and perceived reward. To be able to do this requires the SNN to have a short-term memory.  Chorley \& Seth~\cite{Chorley_2011} have shown a network architecture where the level of dopamine corresponds to the difference between perceived and expected reward through the use of a model of short-term memory.  Future work would incorporate this network architecture into the robot to allow sequential behaviours to be unlearnt.

Another obvious extension to the work in this paper would be to implement the control architecture in a real, rather than simulated, robot.  In this case the robot would need to be able to deal with collision avoidance, which could possibly be solved through a similar reinforcement learning mechanism as described in this paper.

\bibliographystyle{cj}
\bibliography{bibliography}
\end{document}